\documentclass{article}

\usepackage{microtype}

\usepackage{amsmath,amsfonts,amssymb}
\usepackage{mathtools}
\usepackage{amsthm}

\usepackage{graphicx}
\usepackage{booktabs}
\usepackage{subcaption}
\usepackage{multirow}
\usepackage{multicol}
\usepackage{tabularx}
\usepackage{makecell}
\usepackage{array}
\usepackage{colortbl}

\usepackage{textcomp}
\usepackage{bm}
\usepackage{dsfont}
\usepackage{xspace}
\usepackage{rotating}

\usepackage{url}
\usepackage{verbatim}
\usepackage{cite}
\usepackage{siunitx}
\usepackage{listings}
\usepackage{breqn}
\usepackage{titlesec}
\usepackage{placeins}
\usepackage{dblfloatfix}
\usepackage{xcolor}

\usepackage[most]{tcolorbox}

\usepackage{hyperref}

\usepackage[accepted]{icml2026}

\usepackage[capitalize,noabbrev]{cleveref}

\theoremstyle{plain}

\theoremstyle{definition}

\theoremstyle{remark}

\definecolor{BestColor}{RGB}{255,235,200}
\definecolor{BestGray}{gray}{0.90}

\crefname{figure}{Fig.}{Figs.}
\crefname{table}{Table}{Tables}
\crefname{equation}{Eq.}{Eqs.}

\usepackage[disable,textsize=tiny]{todonotes}

\icmltitlerunning{EvoGM: Learning to Merge LLMs via Evolutionary Generative Optimization}

\newcommand{\algEvoGM}{%
\begin{algorithm}[htb]
    \caption{EvoGM: Evolutionary Generative Merging}
    \label{alg:evogm}
  \begin{algorithmic}
    \STATE {\bfseries Input:} Base model $\boldsymbol{\theta}_{pre}$, initial experts $\{\boldsymbol{\theta}_i\}_{i=1}^N$, validation set $\mathcal{D}_{val}$, population size $P$, number of rounds $R$, number of iterations per round $T$
    \STATE {\bfseries Output:} Optimal merging coefficients $\boldsymbol{\lambda}^*$
    
    \STATE Compute initial task vectors: $\boldsymbol{\tau}_i \leftarrow \boldsymbol{\theta}_i - \boldsymbol{\theta}_{pre}$ 
    
    \FOR{round $r=1$ {\bfseries to} $R$}
        \STATE $\mathcal{P} \leftarrow$ \text{InitPop}($P$); $\mathcal{F} \leftarrow f(\boldsymbol{\theta}(\mathcal{P}), \mathcal{D}_{val})$
        \STATE $\mathcal{H} \leftarrow \{(\boldsymbol{\lambda}^{(p)}, \mathcal{F}^{(p)})\}_{p=1}^P$
        \STATE Reset Dual-Generators $G_{- \to +}, G_{+ \to -}$
        
        \FOR{iteration $t=1$ {\bfseries to} $T$}
            \STATE $\mathcal{H}^+, \mathcal{H}^- \leftarrow$ \text{PairWinnersLosers}($\mathcal{H}$)
            \STATE Train Dual-Generators $G_{- \to +}, G_{+ \to -}$
            \STATE $\boldsymbol{\lambda}_{new} \leftarrow G_{- \to +}(\mathcal{H}^+)$
            \STATE $\mathcal{F}_{new} \leftarrow f(\boldsymbol{\theta}(\boldsymbol{\lambda}_{new}), \mathcal{D}_{val})$
            \STATE $\mathcal{H} \leftarrow \mathcal{H} \cup \{(\boldsymbol{\lambda}_{new}, \mathcal{F}_{new})\}$; $\mathcal{P} \leftarrow \text{Select}(\mathcal{H}, P)$
        \ENDFOR
  
        \IF{$r < R$}
            \STATE Select top-$N$ coefficients $\{\boldsymbol{\lambda}^{(i)}\}_{i=1}^N$ from history $\mathcal{H}$
            \STATE Update experts: $\boldsymbol{\theta}_i \leftarrow \boldsymbol{\theta}(\boldsymbol{\lambda}^{(i)})$
            \STATE Re-compute task vectors: $\boldsymbol{\tau}_i \leftarrow \boldsymbol{\theta}_i - \boldsymbol{\theta}_{pre}$ 
        \ENDIF
    \ENDFOR
    \STATE {\bfseries return} best $\boldsymbol{\theta}(\boldsymbol{\lambda}^*)$ found in $\mathcal{H}$
  \end{algorithmic}
  \end{algorithm}
}

\definecolor{BestColor}{gray}{0.91}
\definecolor{BestGray}{gray}{0.90}
\newcommand{\best}[1]{\cellcolor{BestGray}\textbf{#1}}

\newcommand{\TableSingletask}{
    \begin{table*}[t]
        \centering
        \small
        \setlength{\tabcolsep}{3pt}
        \renewcommand{\arraystretch}{1.15}

        \caption{
        Single-task performance comparison of 10 fine-tuned Qwen2.5-1.5B models.
        Best results are highlighted with \textbf{bold} and light-gray background,
        and second-best results are \underline{underlined}, computed independently for each column (Val/Test).
        }\label{tab:singletask}

        \resizebox{\textwidth}{!}{%
        \begin{tabular}{
        >{\centering\arraybackslash}m{2.4cm}
        cccccccccccccccc
        }
        \toprule
      
        \multirow{2}{*}{\textbf{Method}}
        & \multicolumn{2}{c}{\textbf{MMLU}}
        & \multicolumn{2}{c}{\textbf{MMLU-Pro}}
        & \multicolumn{2}{c}{\textbf{HellaSwag}}
        & \multicolumn{2}{c}{\textbf{K-Cross}}
        & \multicolumn{2}{c}{\textbf{GSM8K}}
        & \multicolumn{2}{c}{\textbf{NLGraph}}
        & \multicolumn{2}{c}{\textbf{TruthQA}}
        & \multicolumn{2}{c}{\textbf{AbstainQA}} \\
        \cmidrule(lr){2-17}
      
        & Val & Test
        & Val & Test
        & Val & Test
        & Val & Test
        & Val & Test
        & Val & Test
        & Val & Test
        & Val & Test \\
        \midrule
      
        Base
        & 0.570 & 0.564
        & 0.271 & 0.207
        & 0.590 & 0.581
        & 0.410 & 0.297
        & 0.255 & 0.155
        & 0.285 & 0.284
        & 0.460 & 0.425
        & 0.185 & 0.101 \\
      
        MTL
        & 0.620 & 0.566
        & 0.286 & 0.207
        & 0.525 & 0.484
        & 0.395 & 0.327
        & 0.365 & 0.188
        & 0.320 & 0.319
        & 0.360 & 0.305
        & 0.005 & 0.003 \\
      
        Single Best
        & 0.615 & 0.586
        & \underline{0.386} & 0.232
        & 0.620 & 0.572
        & 0.410 & \underline{0.392}
        & 0.395 & 0.325
        & 0.355 & 0.376
        & 0.495 & 0.384
        & 0.210 & 0.119 \\
        \midrule
      
        Model Soup (22)
        & 0.610 & 0.574
        & 0.300 & 0.225
        & 0.590 & 0.578
        & 0.355 & 0.330
        & 0.300 & \underline{0.397}
        & 0.285 & 0.197
        & 0.465 & 0.283
        & 0.070 & 0.048 \\
      
        TA (23)
        & 0.620 & 0.570
        & 0.286 & \underline{0.234}
        & 0.605 & 0.588
        & 0.370 & 0.391
        & 0.350 & 0.260
        & 0.280 & 0.286
        & 0.465 & 0.389
        & 0.075 & 0.060 \\
      
        TIES (23)
        & 0.600 & 0.557
        & 0.329 & 0.221
        & 0.510 & 0.470
        & 0.345 & 0.342
        & 0.295 & 0.302
        & 0.390 & 0.316
        & 0.495 & 0.416
        & 0.075 & 0.028 \\
      
        DARE (23)
        & 0.605 & 0.576
        & 0.300 & 0.233
        & 0.615 & \underline{0.593}
        & 0.380 & 0.390
        & 0.320 & 0.243
        & 0.285 & 0.286
        & 0.455 & 0.395
        & 0.095 & 0.043 \\
      
        \midrule

        CMA (25)
        & 0.615 & 0.576
        & \underline{0.386} & 0.217
        & 0.600 & 0.592
        & \underline{0.415} & \underline{0.392}
        & 0.430 & 0.343
        & 0.430 & 0.445
        & \underline{0.530} & \underline{0.428}
        & 0.170 & 0.075 \\
      
        PSO-Merging (25)
        & \underline{0.635} & 0.595
        & \underline{0.386} & 0.232
        & \underline{0.650} & 0.587
        & \cellcolor{BestGray}\textbf{0.430} & \cellcolor{BestGray}\textbf{0.393}
        & \underline{0.490} & 0.354
        & \underline{0.455} & \underline{0.465}
        & 0.525 & 0.421
        & \cellcolor{BestGray}\textbf{0.260} & \cellcolor{BestGray}\textbf{0.147} \\
      
        Model Swarm (25)
        & \cellcolor{BestGray}\textbf{0.640} & \underline{0.606}
        & 0.371 & \cellcolor{BestGray}\textbf{0.236}
        & 0.635 & 0.587
        & 0.410 & \underline{0.392}
        & 0.410 & 0.332
        & 0.360 & 0.373
        & 0.495 & 0.392
        & \underline{0.230} & 0.095 \\
      
        \midrule
        Ours
        & \cellcolor{BestGray}\textbf{0.640} & \cellcolor{BestGray}\textbf{0.625}
        & \cellcolor{BestGray}\textbf{0.429} & 0.224
        & \cellcolor{BestGray}\textbf{0.660} & \cellcolor{BestGray}\textbf{0.594}
        & \cellcolor{BestGray}\textbf{0.430} & 0.379
        & \cellcolor{BestGray}\textbf{0.495} & \cellcolor{BestGray}\textbf{0.434}
        & \cellcolor{BestGray}\textbf{0.545} & \cellcolor{BestGray}\textbf{0.537}
        & \cellcolor{BestGray}\textbf{0.540} & \cellcolor{BestGray}\textbf{0.441}
        & \underline{0.230} & \underline{0.121} \\
      
        \bottomrule
        \end{tabular}%
        }
      \end{table*}      
}   

\newcommand{\TableMultitask}{
    \begin{table*}[t]
        \centering
        \scriptsize
        \setlength{\tabcolsep}{2.2pt}
        \renewcommand{\arraystretch}{1.10}

        \caption{
            Multi-task performance comparison of 10 fine-tuned Qwen2.5-1.5B models.
            Best results are highlighted with \textbf{bold} and light-gray background,
            and second-best results are \underline{underlined}, computed independently for each column (Val/Test).
        }\label{tab:multitask}
      
        \resizebox{\textwidth}{!}{%
        \begin{tabular}{
          >{\centering\arraybackslash}m{2.2cm}
          ccccccccccccccccc
        }
        \toprule
      
        \multirow{2}{*}{\textbf{Method}}
        & \multicolumn{2}{c}{\textbf{MMLU}}
        & \multicolumn{2}{c}{\textbf{MMLU-Pro}}
        & \multicolumn{2}{c}{\textbf{HellaSwag}}
        & \multicolumn{2}{c}{\textbf{K-Cross}}
        & \multicolumn{2}{c}{\textbf{GSM8K}}
        & \multicolumn{2}{c}{\textbf{NLGraph}}
        & \multicolumn{2}{c}{\textbf{TruthQA}}
        & \multicolumn{2}{c}{\textbf{AbstainQA}}
        & \multirow{2}{*}{\textbf{Average}} \\
        \cmidrule(lr){2-17}
      
        & Val & Test
        & Val & Test
        & Val & Test
        & Val & Test
        & Val & Test
        & Val & Test
        & Val & Test
        & Val & Test \\
        \midrule
      
        Base
        & 0.570 & 0.564
        & 0.271 & 0.207
        & 0.590 & 0.581
        & 0.410 & 0.297
        & 0.255 & 0.155
        & 0.285 & 0.284
        & 0.460 & 0.425
        & 0.185 & 0.101
        & 0.352 \\
      
        MTL
        & 0.620 & 0.566
        & 0.286 & 0.207
        & 0.525 & 0.484
        & 0.395 & 0.327
        & 0.365 & 0.188
        & 0.320 & 0.319
        & 0.360 & 0.305
        & 0.005 & 0.003
        & 0.330 \\
      
        Single Best
        & 0.615 & 0.586
        & 0.386 & 0.232
        & 0.620 & 0.572
        & 0.410 & 0.392
        & 0.395 & 0.325
        & 0.355 & 0.376
        & 0.495 & 0.384
        & 0.210 & 0.119
        & 0.405 \\
        \midrule
      
        Model Soup (22)
        & \underline{0.610} & \underline{0.574}
        & 0.300 & 0.225
        & 0.590 & 0.578
        & 0.355 & 0.330
        & 0.300 & \cellcolor{BestGray}\textbf{0.397}
        & 0.285 & 0.197
        & 0.465 & 0.283
        & 0.070 & 0.048
        & 0.350 \\
      
        Task Arithmetic (23)
        & \cellcolor{BestGray}\textbf{0.620} & 0.570
        & 0.286 & \cellcolor{BestGray}\textbf{0.234}
        & 0.605 & 0.588
        & 0.370 & 0.391
        & \underline{0.350} & 0.260
        & 0.280 & 0.286
        & 0.465 & 0.389
        & 0.075 & 0.060
        & 0.364 \\
      
        TIES (23)
        & 0.600 & 0.557
        & 0.329 & 0.221
        & 0.510 & 0.470
        & 0.345 & 0.342
        & 0.295 & \underline{0.302}
        & \cellcolor{BestGray}\textbf{0.390} & \underline{0.316}
        & 0.495 & 0.416
        & 0.075 & 0.028
        & 0.356 \\
      
        DARE (23)
        & 0.605 & \cellcolor{BestGray}\textbf{0.576}
        & 0.300 & \underline{0.233}
        & 0.615 & \underline{0.597}
        & 0.380 & 0.390
        & 0.320 & 0.243
        & 0.285 & 0.286
        & 0.455 & 0.395
        & 0.095 & 0.043
        & 0.364 \\
      
        \midrule

        CMA (25)
        & 0.600 & 0.569
        & \cellcolor{BestGray}\textbf{0.357} & 0.217
        & \cellcolor{BestGray}\textbf{0.645} & \cellcolor{BestGray}\textbf{0.609}
        & \underline{0.405} & \cellcolor{BestGray}\textbf{0.415}
        & 0.275 & 0.211
        & 0.275 & 0.275
        & \underline{0.500} & \cellcolor{BestGray}\textbf{0.421}
        & \underline{0.140} & \underline{0.078}
        & \underline{0.375} \\

        PSO-Merging (25)
        & 0.580 & 0.539
        & 0.314 & 0.197
        & 0.530 & 0.532
        & \cellcolor{BestGray}\textbf{0.415} & \underline{0.414}
        & 0.315 & 0.261
        & \underline{0.380} & \cellcolor{BestGray}\textbf{0.352}
        & \cellcolor{BestGray}\textbf{0.520} & \underline{0.418}
        & 0.130 & 0.059
        & 0.372 \\
      
        Model Swarm (25)
        & \underline{0.610} & 0.573
        & \underline{0.343} & 0.229
        & 0.585 & 0.545
        & 0.390 & 0.411
        & \cellcolor{BestGray}\textbf{0.375} & 0.246
        & 0.305 & 0.297
        & 0.440 & 0.410
        & 0.130 & 0.071
        & 0.372 \\
      
        \midrule
        Ours
        & \cellcolor{BestGray}\textbf{0.620} & \cellcolor{BestGray}\textbf{0.576}
        & 0.314 & 0.225
        & \underline{0.635} & 0.588
        & \underline{0.405} & 0.388
        & \underline{0.350} & 0.248
        & 0.295 & 0.290
        & 0.455 & 0.397
        & \cellcolor{BestGray}\textbf{0.165} & \cellcolor{BestGray}\textbf{0.130}
        & \cellcolor{BestGray}\textbf{0.380} \\
      
        \bottomrule
        \end{tabular}%
        }
      \end{table*}
}   

\newcommand{\TableParameterTuning}{
    \begin{table*}[!ht]
        \centering
        \small
        \setlength{\tabcolsep}{4pt}
        \renewcommand{\arraystretch}{1.15}

        \caption{
        Ablation study on key hyperparameters.
        The baseline configuration is marked with ``--''.
        In the $\Delta$ Test Acc column, the largest improvement is highlighted in \textbf{bold},
        while performance degradation is shown in \textit{italic}.
        }\label{tab:parameter-tuning}
        
        \begin{tabular}{
        >{\raggedright\arraybackslash}m{3.0cm}
        ccccccccc
        }
        \toprule
        \textbf{Parameter} & \textbf{Value} & Pop & Iter & Epochs & LR & Rounds & Dev Acc & Test Acc & $\Delta$ Test Acc \\
        \midrule
        
        \multirow{3}{*}{Population Size}
        & 10 & 10 & 5 & 400 & 0.001 & 3 & 0.4954 & 0.4100 & +0.0050 \\
        & 20 & 20 & 5 & 400 & 0.001 & 3 & 0.5104 & 0.4180 & -- \\
        & 30 & 30 & 5 & 400 & 0.001 & 3 & 0.5129 & 0.4240 & \textbf{+0.0305} \\
        \midrule
        
        \multirow{3}{*}{Max Iterations}
        & 3 & 20 & 3 & 400 & 0.001 & 3 & 0.5200 & 0.4195 & +0.0145 \\
        & 5 & 20 & 5 & 400 & 0.001 & 3 & 0.5104 & 0.4180 & -- \\
        & 8 & 20 & 8 & 400 & 0.001 & 3 & 0.5150 & 0.4245 & +0.0195 \\
        \midrule
        
        \multirow{3}{*}{Generator Epochs}
        & 200 & 20 & 5 & 200 & 0.001 & 3 & 0.5125 & 0.4245 & +0.0115 \\
        & 400 & 20 & 5 & 400 & 0.001 & 3 & 0.5104 & 0.4180 & -- \\
        & 600 & 20 & 5 & 600 & 0.001 & 3 & 0.5129 & 0.4240 & \textit{-0.0035} \\
        \midrule
        
        \multirow{3}{*}{Learning Rate}
        & 0.0001 & 20 & 5 & 400 & 0.0001 & 3 & 0.5175 & 0.4270 & +0.0220 \\
        & 0.0010 & 20 & 5 & 400 & 0.0010 & 3 & 0.5104 & 0.4180 & -- \\
        & 0.0050 & 20 & 5 & 400 & 0.0050 & 3 & 0.5175 & 0.4155 & +0.0105 \\
        \midrule
        
        \multirow{3}{*}{Evolution Rounds}
        & 1 & 20 & 5 & 400 & 0.001 & 1 & 0.5046 & 0.4015 & \textit{-0.0165} \\
        & 3 & 20 & 5 & 400 & 0.001 & 3 & 0.5104 & 0.4180 & -- \\
        & 5 & 20 & 5 & 400 & 0.001 & 5 & 0.5264 & 0.4195 & +0.0130 \\
        \bottomrule
        \end{tabular}
        
    \end{table*}
}   

\newcommand{\TableAblationStudy}{
\begin{table*}[!ht]
\centering
\scriptsize
\setlength{\tabcolsep}{2.2pt}
\renewcommand{\arraystretch}{1.10}

\caption{
Ablation study of EvoGM on 8 single-task benchmarks.
Best results are highlighted with \textbf{bold} and a light-gray background,
computed independently for each column (Val/Test).
}\label{tab:ablation}

\resizebox{\textwidth}{!}{%
\begin{tabular}{
>{\centering\arraybackslash}m{2.2cm}
cccccccccccccccc
}
\toprule
\multirow{2}{*}{\textbf{Method}}
& \multicolumn{2}{c}{\textbf{MMLU}}
& \multicolumn{2}{c}{\textbf{MMLU-Pro}}
& \multicolumn{2}{c}{\textbf{HellaSwag}}
& \multicolumn{2}{c}{\textbf{K-Cross}}
& \multicolumn{2}{c}{\textbf{GSM8K}}
& \multicolumn{2}{c}{\textbf{NLGraph}}
& \multicolumn{2}{c}{\textbf{TruthQA}}
& \multicolumn{2}{c}{\textbf{AbstainQA}} \\
\cmidrule(lr){2-17}

& Val & Test
& Val & Test
& Val & Test
& Val & Test
& Val & Test
& Val & Test
& Val & Test
& Val & Test \\
\midrule

Single-Generator
& 0.620 & 0.576
& 0.385 & 0.232
& 0.650 & 0.590
& 0.425 & 0.412
& \best{0.540} & 0.339
& 0.515 & 0.529
& 0.515 & 0.439
& 0.210 & 0.119 \\

w/o Rounds
& 0.630 & 0.614
& 0.400 & \best{0.247}
& 0.640 & 0.593
& \best{0.450} & \best{0.417}
& 0.510 & 0.337
& 0.515 & 0.521
& 0.510 & \best{0.441}
& 0.215 & \best{0.142} \\

w/o Cycle Loss
& 0.620 & 0.576
& 0.386 & 0.232
& 0.645 & 0.588
& 0.435 & 0.396
& 0.410 & 0.326
& 0.410 & 0.372
& 0.515 & 0.429
& 0.210 & 0.119 \\

\midrule
EvoGM 
& \best{0.640} & \best{0.625}
& \best{0.429} & 0.224
& \best{0.660} & \best{0.594}
& \best{0.450} & 0.379
& 0.495 & \best{0.434}
& \best{0.545} & \best{0.537}
& \best{0.540} & \best{0.441}
& \best{0.230} & 0.121 \\
\bottomrule
\end{tabular}%
}
\end{table*}
}

\newcommand{\TableNumberModel}{
\begin{table*}[!ht]
\centering
\small
\setlength{\tabcolsep}{3.6pt}
\renewcommand{\arraystretch}{1.15}

\caption{Performance comparison across different numbers of models and evaluation stages.}
\label{tab:ensemble_results}

\resizebox{\textwidth}{!}{%
\begin{tabular}{
>{\centering\arraybackslash}m{1.4cm}
ccccc
>{\centering\arraybackslash}m{1.4cm}
ccccc
}
\toprule

\textbf{Models}
& \multicolumn{5}{c}{\textbf{Initial Val}}
& \textbf{Models}
& \multicolumn{5}{c}{\textbf{Final Val}} \\
\cmidrule(lr){2-6} \cmidrule(lr){8-12}

& \textbf{MMLU}
& \textbf{MMLU-Pro}
& \textbf{HellaSwag}
& \textbf{K-Cross}
& \textbf{Avg}
&
& \textbf{MMLU}
& \textbf{MMLU-Pro}
& \textbf{HellaSwag}
& \textbf{K-Cross}
& \textbf{Avg} \\
\midrule

2  & 0.580 & 0.329 & 0.600 & 0.390 & 0.475 & 2  & 0.575 & 0.329 & 0.605 & 0.395 & 0.476 \\
3  & 0.620 & 0.300 & 0.625 & 0.360 & 0.476 & 3  & 0.615 & 0.314 & 0.640 & 0.365 & 0.484 \\
4  & 0.585 & 0.314 & 0.635 & 0.375 & 0.477 & 4  & 0.605 & 0.400 & 0.645 & 0.345 & 0.499 \\
5  & 0.615 & 0.314 & 0.615 & 0.355 & 0.475 & 5  & 0.605 & 0.414 & 0.650 & 0.350 & 0.505 \\
6  & 0.605 & 0.286 & 0.615 & 0.360 & 0.466 & 6  & 0.615 & 0.400 & 0.660 & 0.345 & 0.505 \\
7  & 0.605 & 0.286 & 0.595 & 0.370 & 0.464 & 7  & 0.595 & 0.400 & 0.655 & 0.395 & 0.511 \\
8  & 0.600 & 0.386 & 0.565 & 0.360 & 0.478 & 8  & 0.630 & 0.414 & 0.645 & 0.345 & 0.509 \\
9  & 0.600 & 0.386 & 0.565 & 0.360 & 0.478 & 9  & 0.615 & 0.386 & 0.655 & 0.375 & 0.508 \\
10 & 0.600 & 0.386 & 0.565 & 0.360 & 0.478 & 10 & 0.620 & 0.343 & 0.650 & 0.385 & 0.499 \\

\midrule

\textbf{Models}
& \multicolumn{5}{c}{\textbf{Initial Test}}
& \textbf{Models}
& \multicolumn{5}{c}{\textbf{Final Test}} \\
\cmidrule(lr){2-6} \cmidrule(lr){8-12}

& \textbf{MMLU}
& \textbf{MMLU-Pro}
& \textbf{HellaSwag}
& \textbf{K-Cross}
& \textbf{Avg}
&
& \textbf{MMLU}
& \textbf{MMLU-Pro}
& \textbf{HellaSwag}
& \textbf{K-Cross}
& \textbf{Avg} \\
\midrule

2  & 0.559 & 0.196 & 0.578 & 0.400 & 0.433 & 2  & 0.552 & 0.193 & 0.571 & 0.400 & 0.429 \\
3  & 0.576 & 0.255 & 0.563 & 0.385 & 0.445 & 3  & 0.578 & 0.254 & 0.561 & 0.385 & 0.445 \\
4  & 0.587 & 0.241 & 0.578 & 0.385 & 0.448 & 4  & 0.601 & 0.250 & 0.590 & 0.386 & 0.457 \\
5  & 0.579 & 0.251 & 0.585 & 0.391 & 0.452 & 5  & 0.600 & 0.253 & 0.601 & 0.380 & 0.459 \\
6  & 0.577 & 0.235 & 0.593 & 0.388 & 0.448 & 6  & 0.602 & 0.245 & 0.589 & 0.391 & 0.457 \\
7  & 0.574 & 0.234 & 0.592 & 0.394 & 0.449 & 7  & 0.589 & 0.234 & 0.599 & 0.393 & 0.454 \\
8  & 0.582 & 0.232 & 0.575 & 0.344 & 0.433 & 8  & 0.608 & 0.246 & 0.599 & 0.386 & 0.460 \\
9  & 0.582 & 0.232 & 0.575 & 0.344 & 0.433 & 9  & 0.595 & 0.244 & 0.595 & 0.398 & 0.458 \\
10 & 0.582 & 0.232 & 0.575 & 0.344 & 0.433 & 10 & 0.582 & 0.228 & 0.590 & 0.405 & 0.451 \\

\bottomrule
\end{tabular}%
}
\end{table*}
}

\newcommand{\TableSeenTask}{
    \begin{table}[!ht]
        \centering
        \scriptsize
        \setlength{\tabcolsep}{1.8pt}
        \renewcommand{\arraystretch}{1.05}

        \caption{Performance comparison on 8 GLUE benchmark tasks. Best results are highlighted with \textbf{bold} and light-gray background,
        and second-best results are \underline{underlined}. 
        }
        \label{tab:seen-task}
        
        \begin{tabular}{
        >{\centering\arraybackslash}m{1.5cm}
        ccccccccc
        }
        \toprule
        \textbf{Method}
        & \textbf{CoLA}
        & \textbf{MNLI}
        & \textbf{MRPC}
        & \textbf{QNLI}
        & \textbf{QQP}
        & \textbf{RTE}
        & \textbf{SST-2}
        & \textbf{STS-B}
        & \textbf{AVG} \\
        \midrule
        
        TA
        & 69.1 & 62.7 & 79.4 & \underline{89.8} & \underline{83.9} & 81.2 & \underline{91.7} & 73.2 & 78.9 \\
        
        DARE
        & 69.5 & 63.8 & 79.7 & \best{89.9} & \underline{83.9} & 81.2 & \underline{91.7} & 69.8 & 78.7 \\
        
        TIES
        & 69.2 & 59.4 & 77.7 & 89.3 & 83.4 & 80.5 & 91.3 & 68.4 & 77.4 \\
        
        DARE-TIES
        & 69.3 & 62.5 & 79.7 & \underline{89.8} & 83.8 & 81.6 & 91.3 & 71.1 & 78.6 \\
        
        DELLA
        & 69.3 & 64.4 & \underline{79.9} & \best{89.9} & 83.8 & \underline{82.0} & 91.1 & \underline{76.0} & 79.5 \\
        
        RankMean
        & 69.1 & 56.5 & 76.2 & 88.5 & 82.1 & 80.1 & 91.2 & 62.2 & 75.7 \\
        
        CMA
        & \underline{70.9} & \underline{82.9} & 75.7 & 89.4 & 73.9 & 80.9 & \best{92.2} & 69.7 & 79.4 \\
        
        AdaMerging
        & 69.9 & 77.2 & \underline{79.9} & \underline{89.8} & 81.7 & 79.1 & 91.4 & 66.1 & 79.4 \\
        
        Fisher
        & 69.3 & 54.0 & 76.7 & 84.6 & 83.6 & 77.6 & 88.1 & 74.4 & 76.0 \\
        
        RegMean
        & 69.1 & 26.6 & 75.3 & 79.3 & 77.2 & 61.7 & 86.0 & 48.1 & 65.4 \\
        
        PSO-Merging
        & 68.2 & \best{83.8} & \best{80.6} & 89.5 & 83.6 & 81.2 & 91.1 & 71.9 & \underline{81.2} \\
        
        \midrule
        Ours
        & \best{71.1} & 82.8 & 78.9 & 88.1
        & \best{84.8} & \best{82.2} & 90.0 & \best{80.9} & \best{82.4} \\
        
        \bottomrule
        \end{tabular}
        
    \end{table}
}

\newcommand{\FigureAblation}{
\begin{figure}[!htbp]
    \centering
    \begin{subfigure}{0.48\linewidth}
        \centering
        \includegraphics[width=\linewidth]{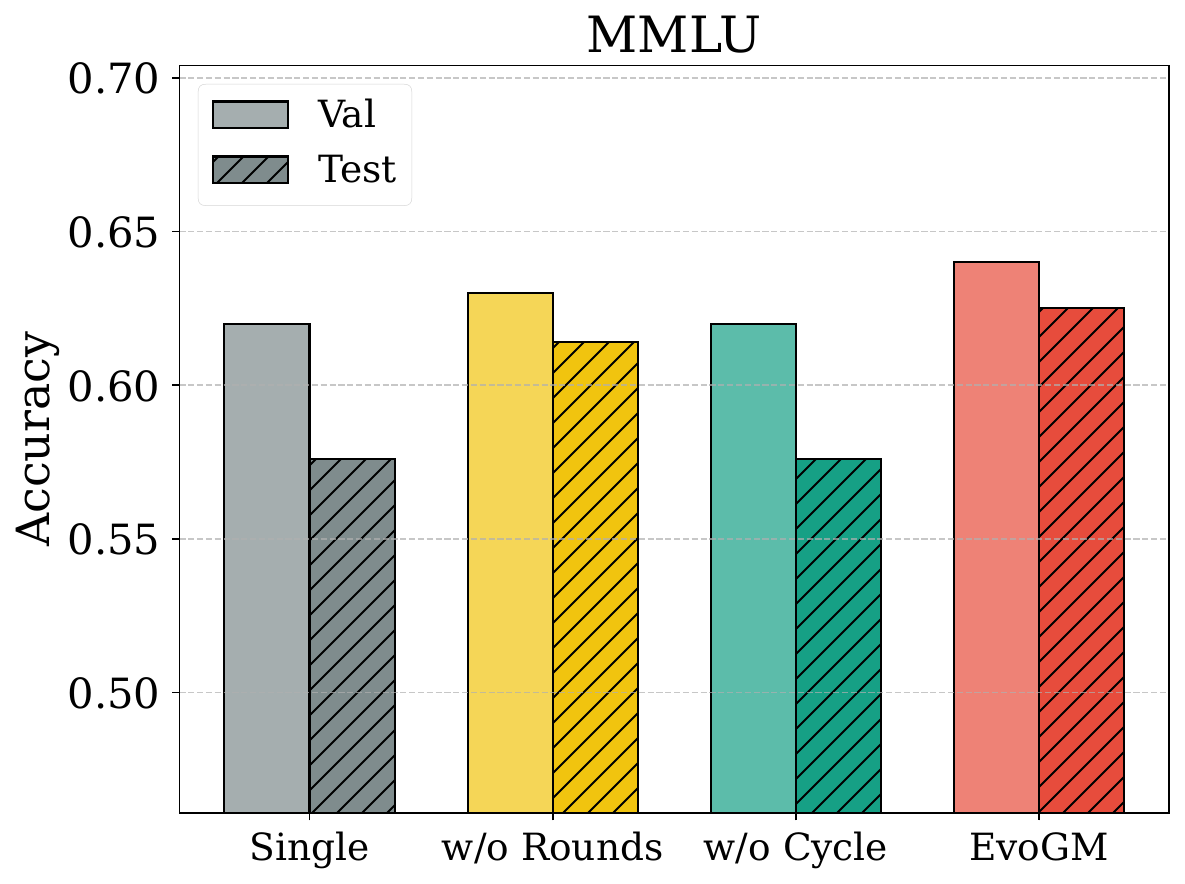}
        \caption{MMLU}
    \end{subfigure}
    \hfill
    \begin{subfigure}{0.48\linewidth}
        \centering
        \includegraphics[width=\linewidth]{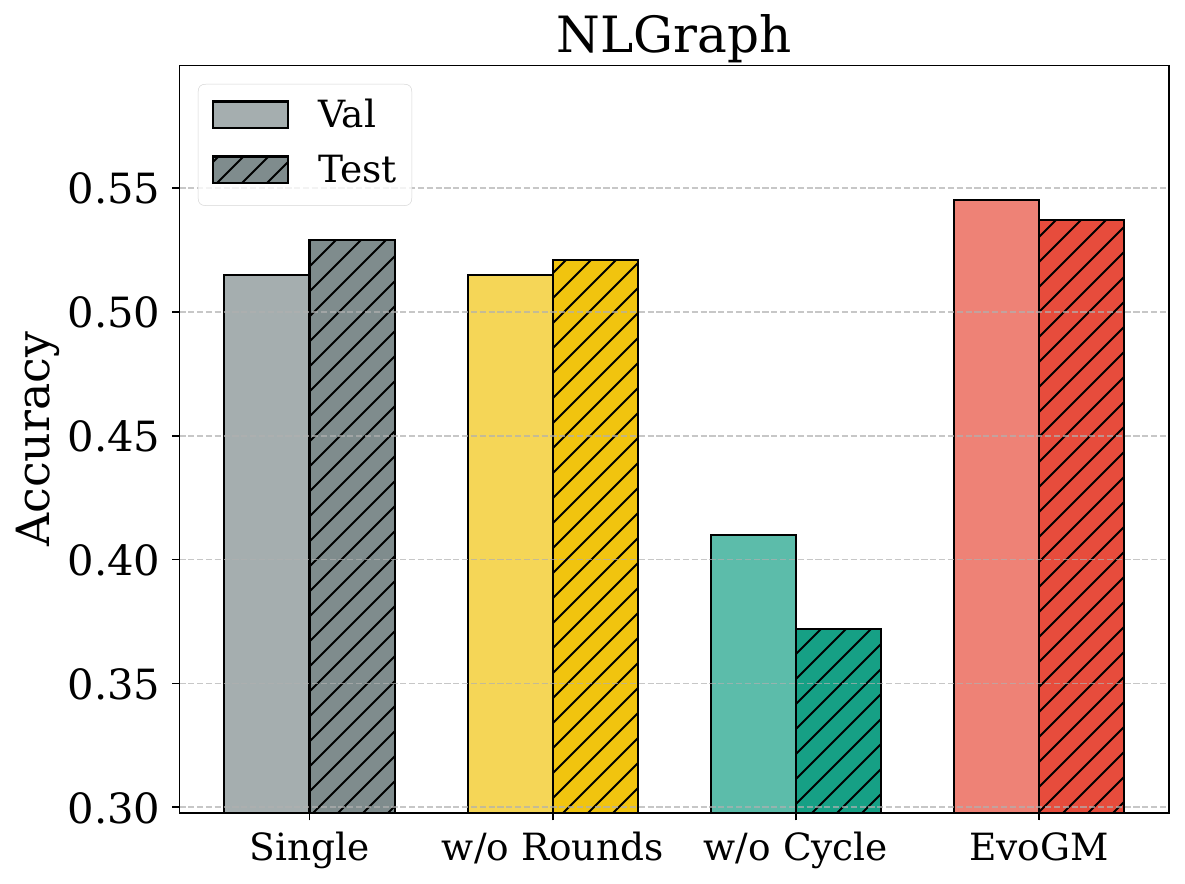}
        \caption{NLGraph}
    \end{subfigure}

    \vspace{0.5em}

    \begin{subfigure}{0.48\linewidth}
        \centering
        \includegraphics[width=\linewidth]{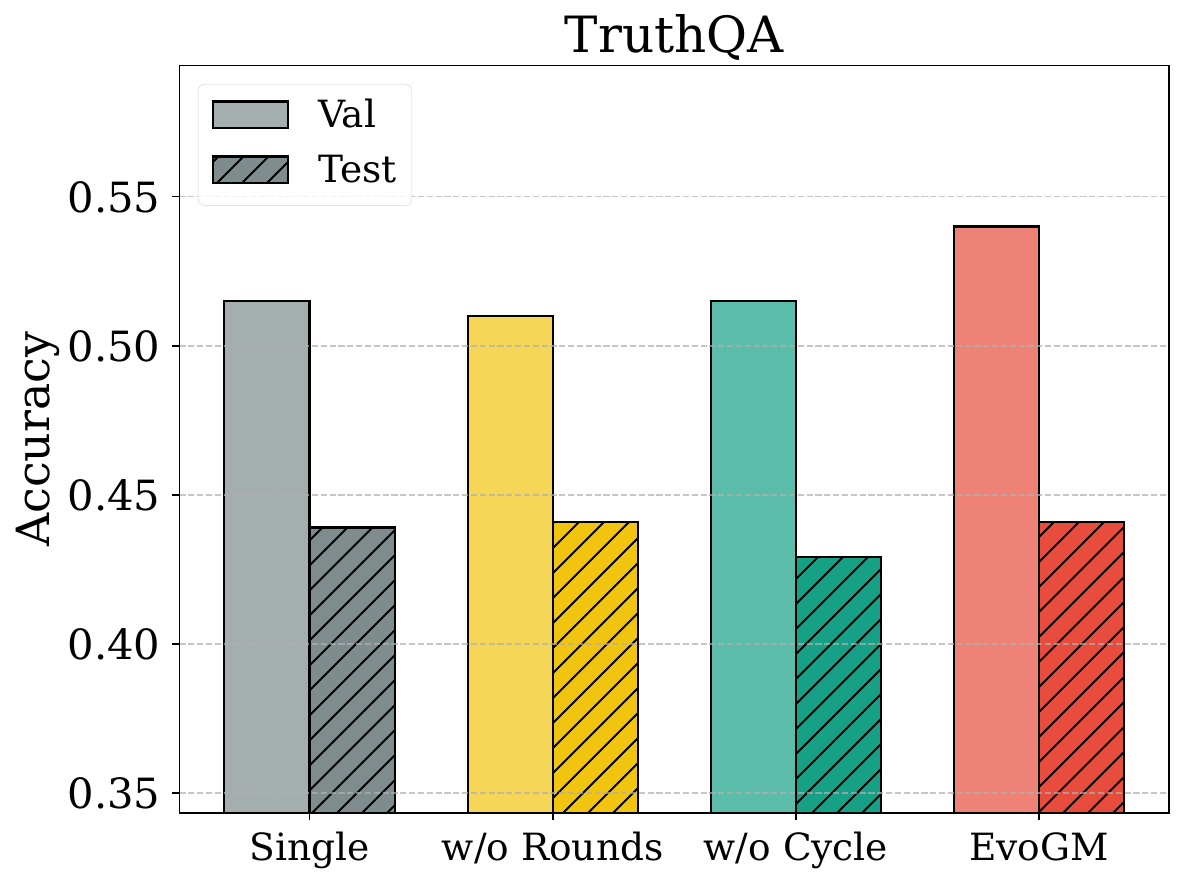}
        \caption{TruthQA}
    \end{subfigure}
    \hfill
    \begin{subfigure}{0.48\linewidth}
        \centering
        \includegraphics[width=\linewidth]{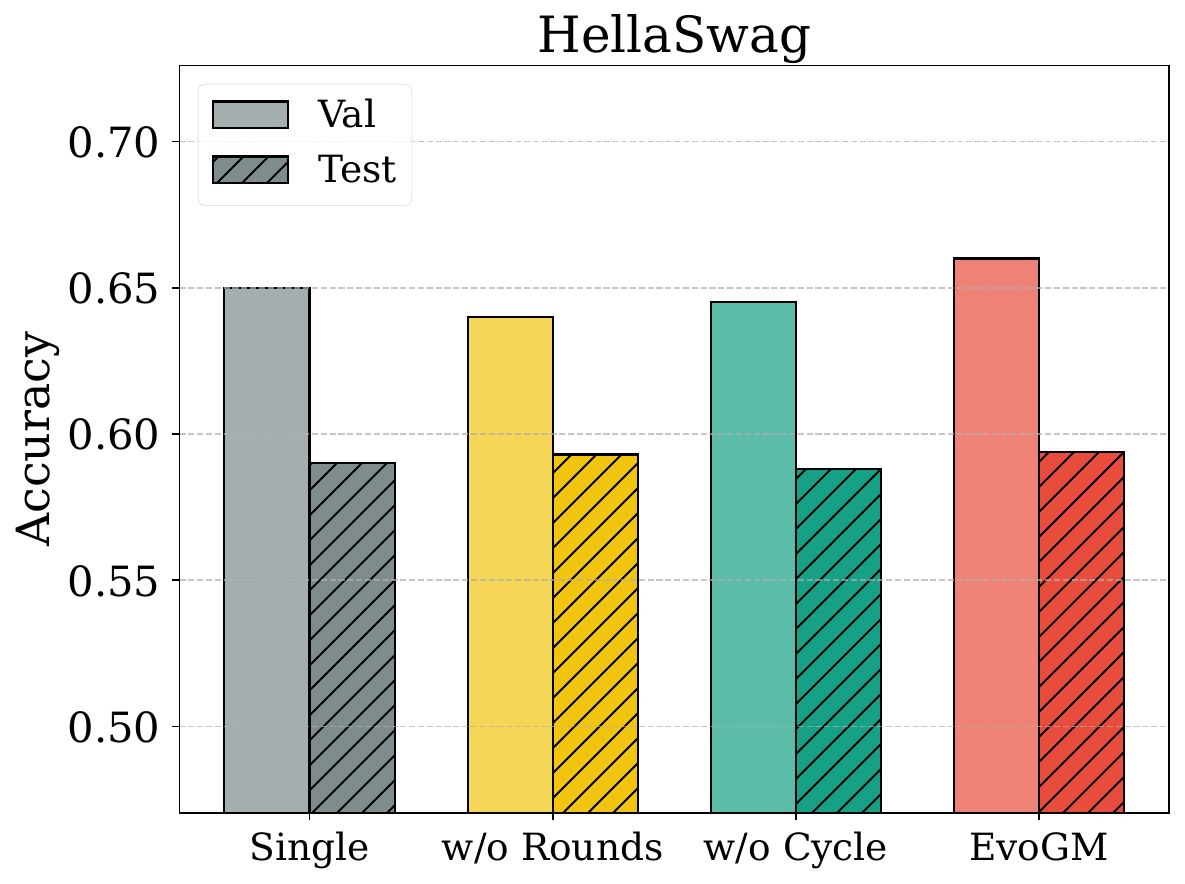}
        \caption{HellaSwag}
    \end{subfigure}

    \caption{Ablation study on different components. (1) Single-Generator represents the single generative model variant; (2) w/o Rounds represents the replacement of multi-round updates with five continuous iterations; and (3) w/o Cycle Loss represents the removal of the cycle-consistency constraint.}
    \label{fig:ablation}
\end{figure}
}

\newcommand{\FigureConvergence}{
\begin{figure}[htbp]
    \centering
    \begin{subfigure}{0.48\linewidth}
        \centering
        \includegraphics[width=\linewidth]{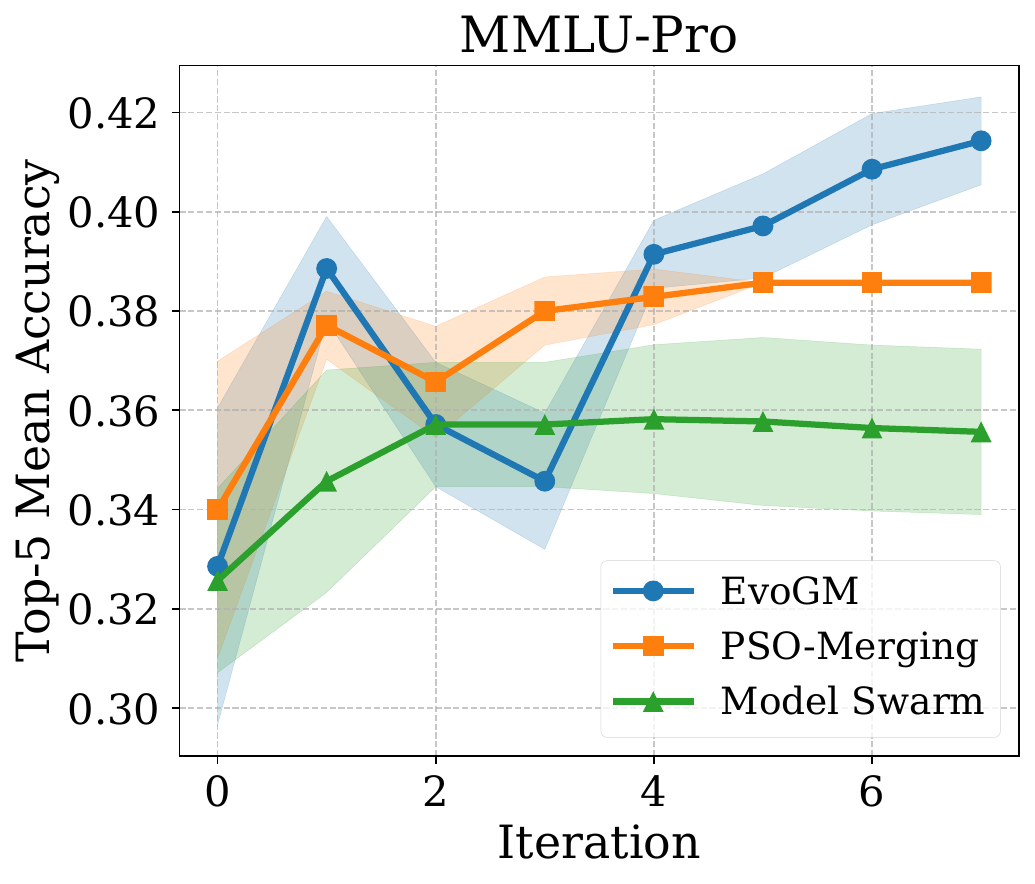}
        \caption{MMLU-Pro}
    \end{subfigure}
    \hfill
    \begin{subfigure}{0.48\linewidth}
        \centering
        \includegraphics[width=\linewidth]{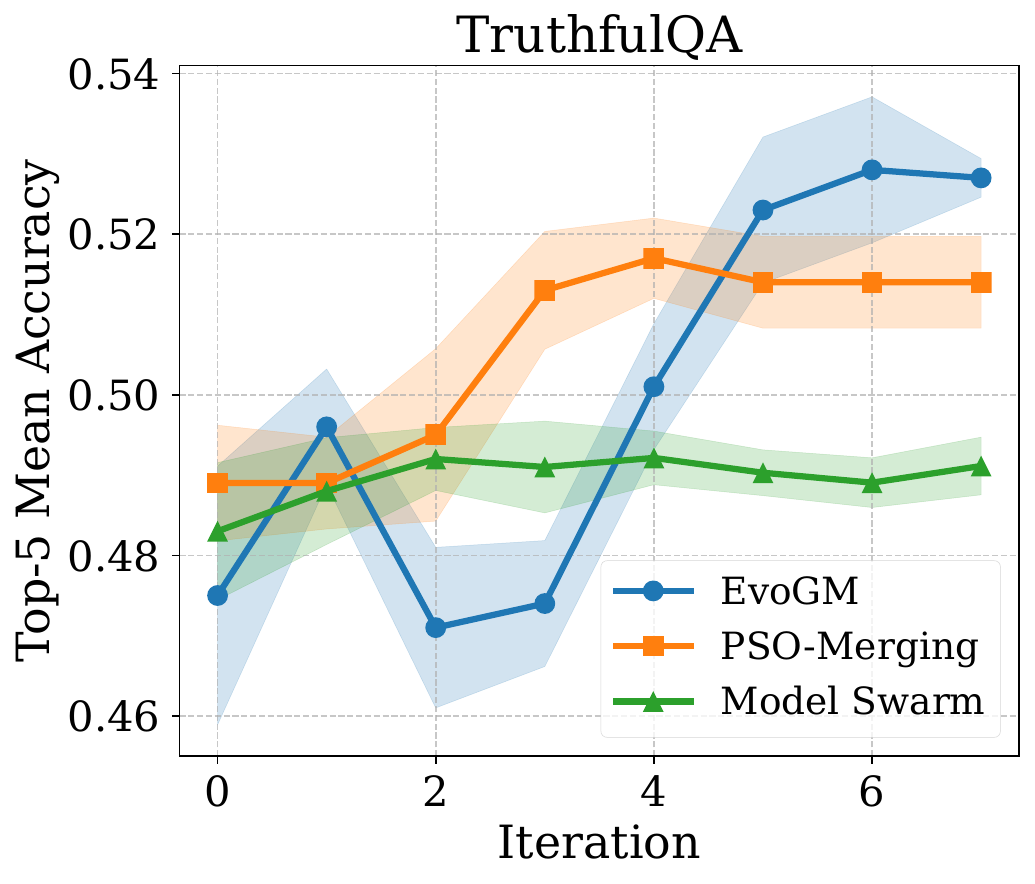}
        \caption{TruthfulQA}
    \end{subfigure}

    \vspace{0.5em}

    \begin{subfigure}{0.48\linewidth}
        \centering
        \includegraphics[width=\linewidth]{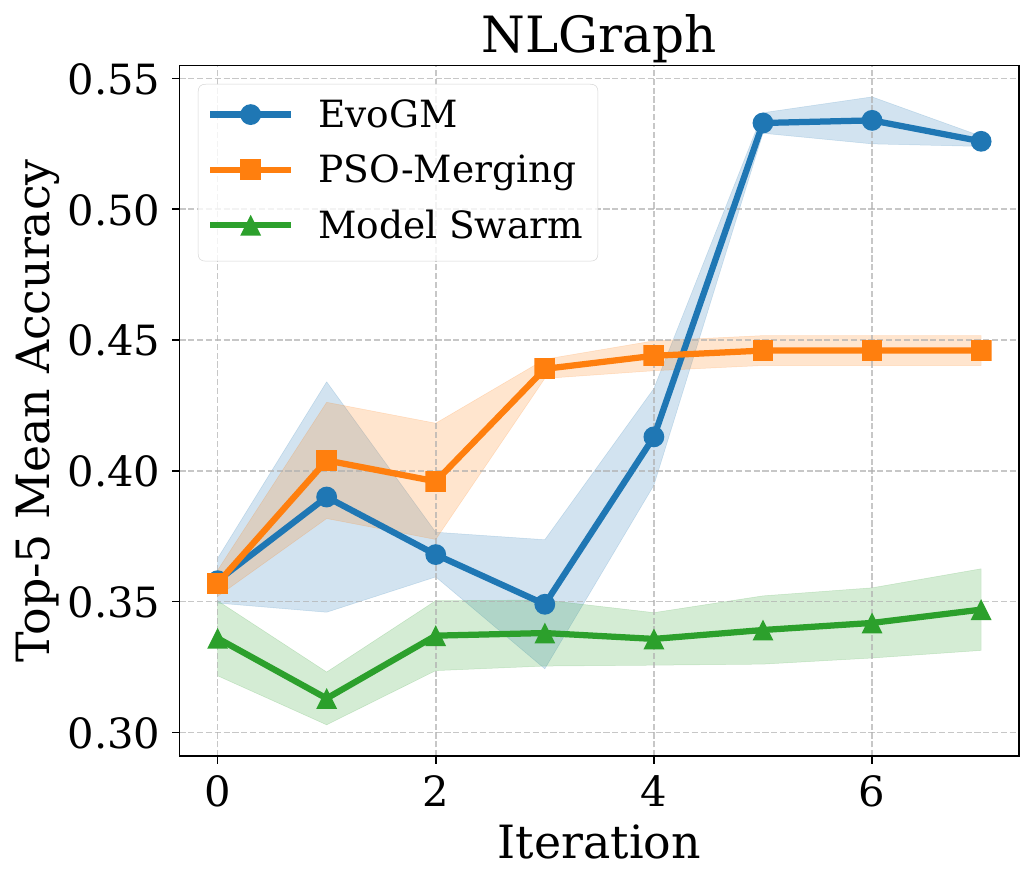}
        \caption{NLGraph}
    \end{subfigure}
    \hfill
    \begin{subfigure}{0.48\linewidth}
        \centering
        \includegraphics[width=\linewidth]{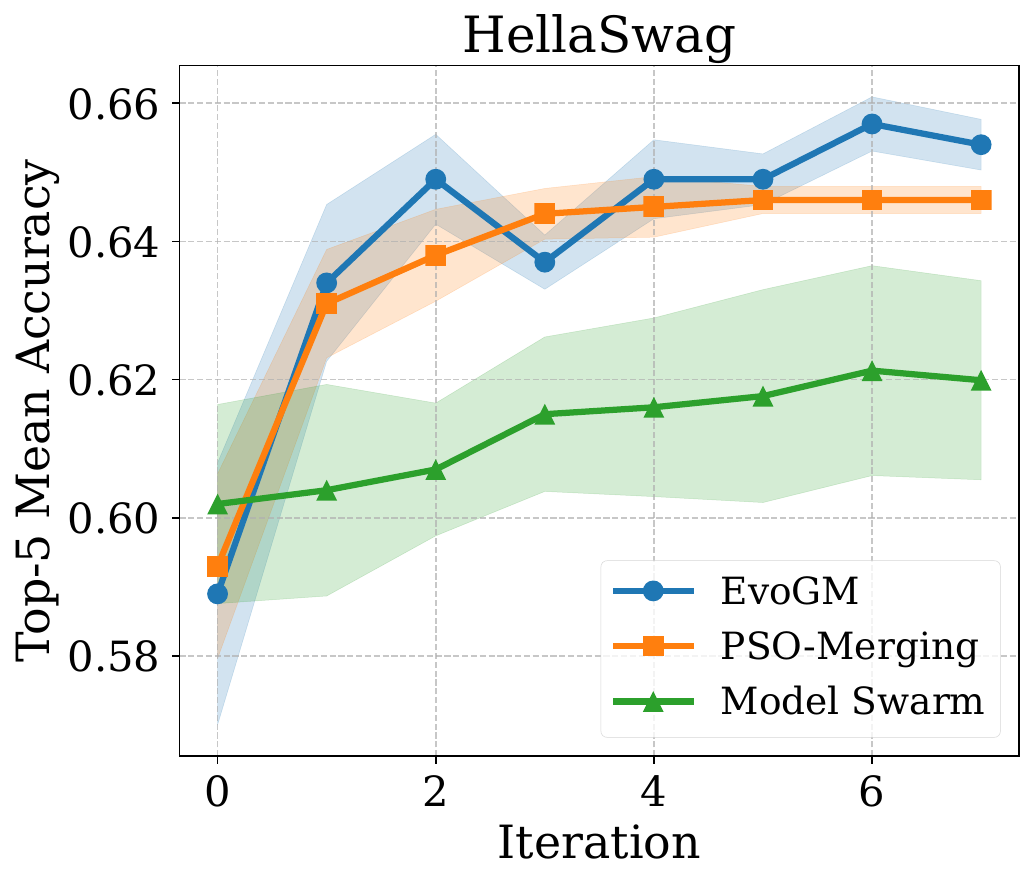}
        \caption{HellaSwag}
    \end{subfigure}

    \caption{Fitness evolution of EvoGM and SOTA methods in multi-task scenarios. The curves represent the mean performance of the top five individuals in the population, with shaded areas indicating the confidence intervals computed from these individuals.}
    \label{fig:convergence}
\end{figure}
}

\newcommand{\FigureMergingTrends}{
\begin{figure}[!htbp]
    \centering
    \includegraphics[width=0.8\linewidth]{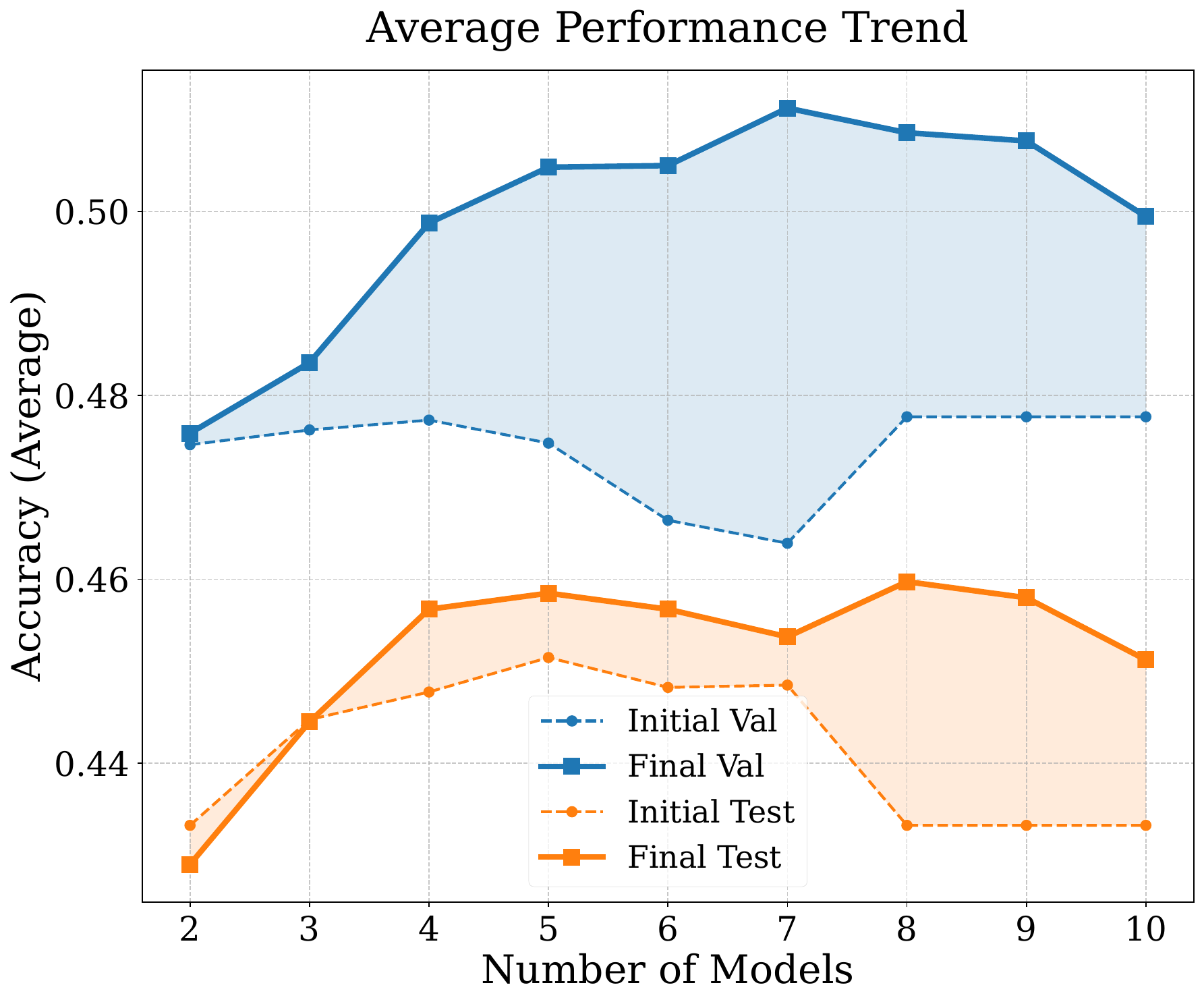}
    \caption{Impact of the number of merged models on performance. We evaluate how the merging quality scales when integrating different quantities of models.}
    \label{fig:merging_trends}
\end{figure}
}

\newcommand{\FigureMethod}{
\begin{figure*}[t]
    \centering
    \includegraphics[width=\linewidth]{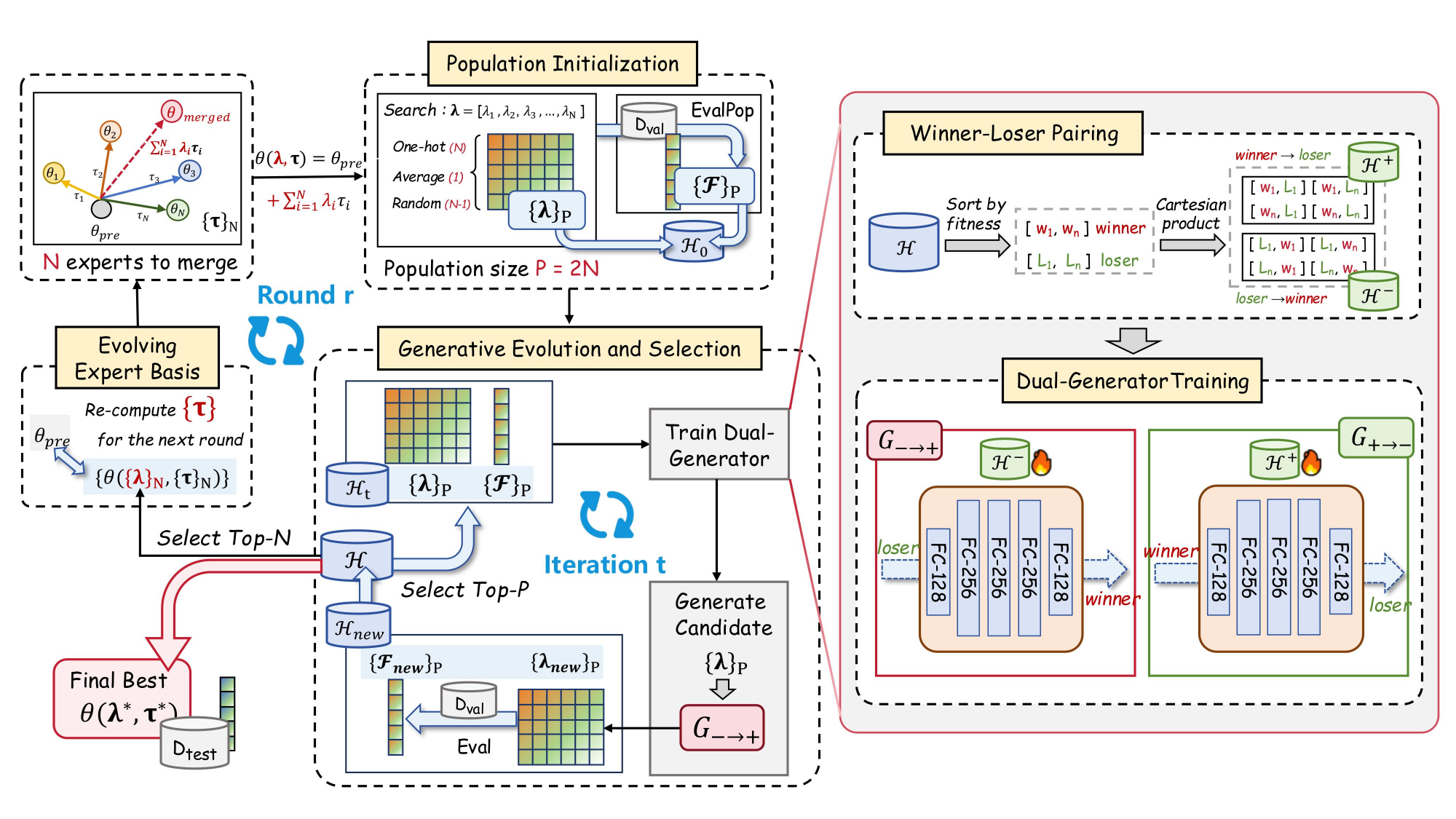}
    \caption{\textbf{Overview of EvoGM.} We optimize merging coefficients $\boldsymbol{\lambda}\in\mathbb{R}^N$ for task-vector merging. (1) \emph{Population Initialization:} initialize a diverse population of $\boldsymbol{\lambda}$ (average merge, one-hot, random) and evaluate on $\mathcal{D}_{\mathrm{val}}$ to build the history set $\mathcal{H}=\{(\boldsymbol{\lambda},f(\boldsymbol{\lambda}))\}$. (2) \emph{Winner--Loser Pairing:} split $\mathcal{H}$ into winners $\mathcal{H}^+$ and losers $\mathcal{H}^-$. (3) \emph{Dual-Generator Training:} train $(G_{-\to +},G_{+\to -})$ with cycle-consistency and optimization-guided losses. (4) \emph{Generative Evolution \& Selection:} apply $G_{-\to +}$ to generate new candidates and select improved $\boldsymbol{\lambda}$ by evaluation. (5) \emph{Evolving Expert Basis:} periodically refresh the expert pool via elite merges and recompute $\{\boldsymbol{\tau}_i\}_{i=1}^N$, returning the best $\boldsymbol{\lambda}^*$ and merged model $\theta(\boldsymbol{\lambda}^*)$.}
    \label{fig:method1}
\end{figure*}
}

\newcommand{\FigureAbstart}{
\begin{figure}[!htp]
    \centering
    \includegraphics[width=0.8\linewidth]{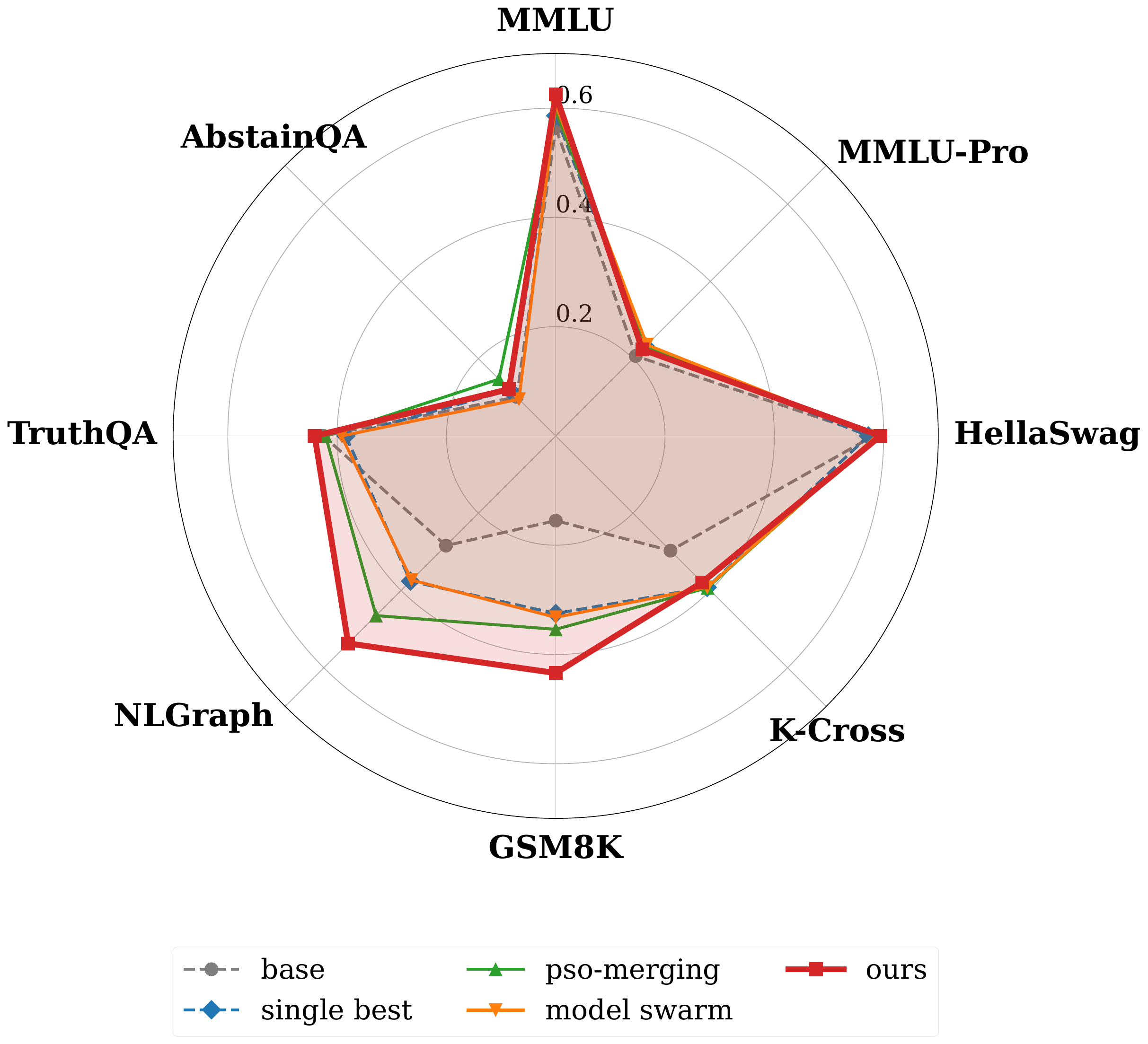}
    \caption{Single-task performance comparison of 10 fine-tuned Qwen2.5-1.5B models. The proposed method outperforms baseline models on almost all targeted tasks.}
    \label{fig:performance_single}
\end{figure}
}

\newcommand{\FigurePerformanceAll}{
\begin{figure*}[!htp]
    \centering
    \includegraphics[width=\linewidth]{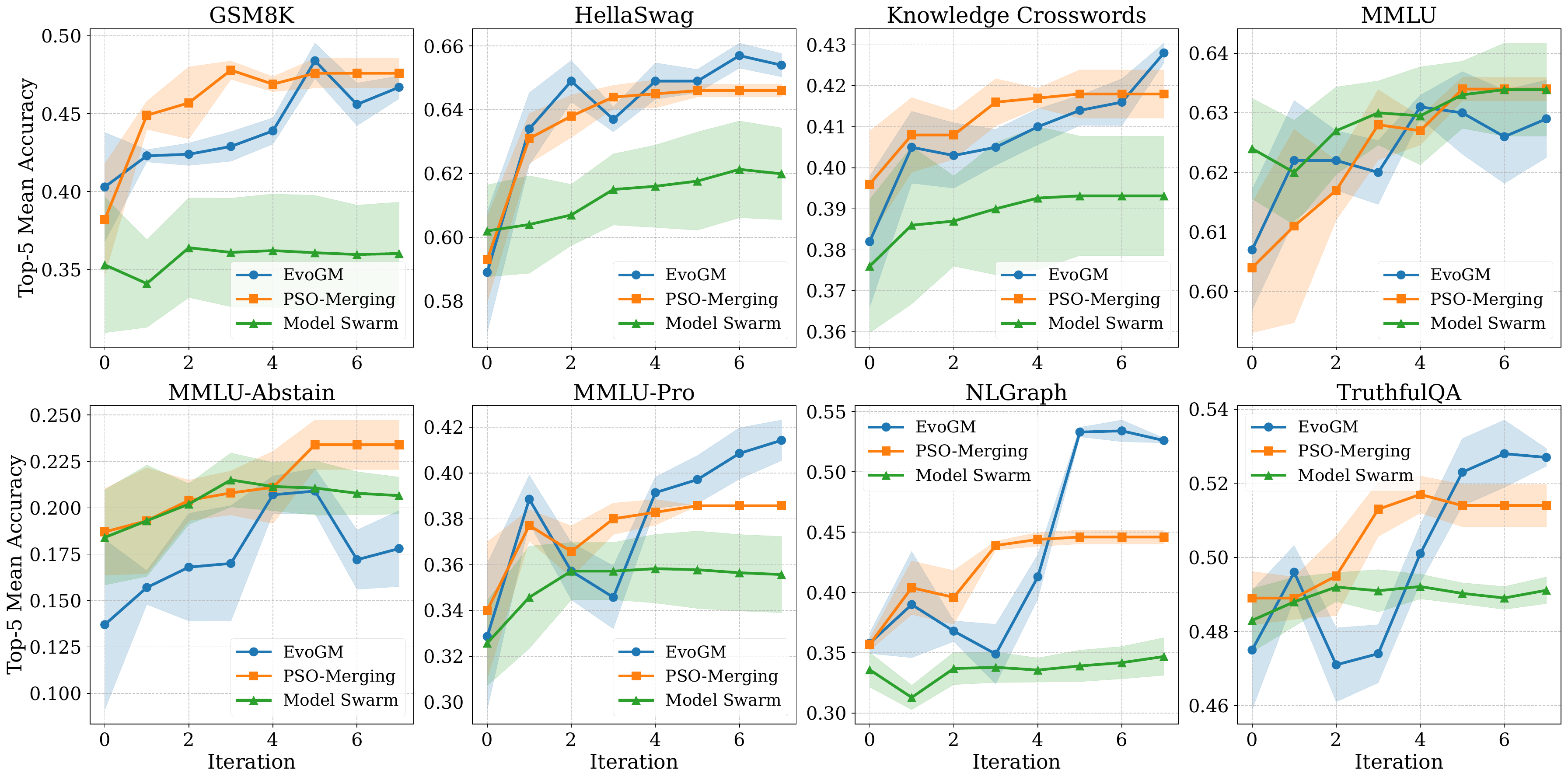}
    \caption{Fitness evolution of EvoGM and SOTA methods in multi-task scenarios. The curves represent the mean performance of the top five individuals in the population, with shaded areas indicating the confidence intervals computed from these individuals.}
    \label{fig:performance_all}
\end{figure*}
}

\newcommand{\FigureAblationAll}{
\begin{figure*}[!htp]
    \centering
    \includegraphics[width=\linewidth]{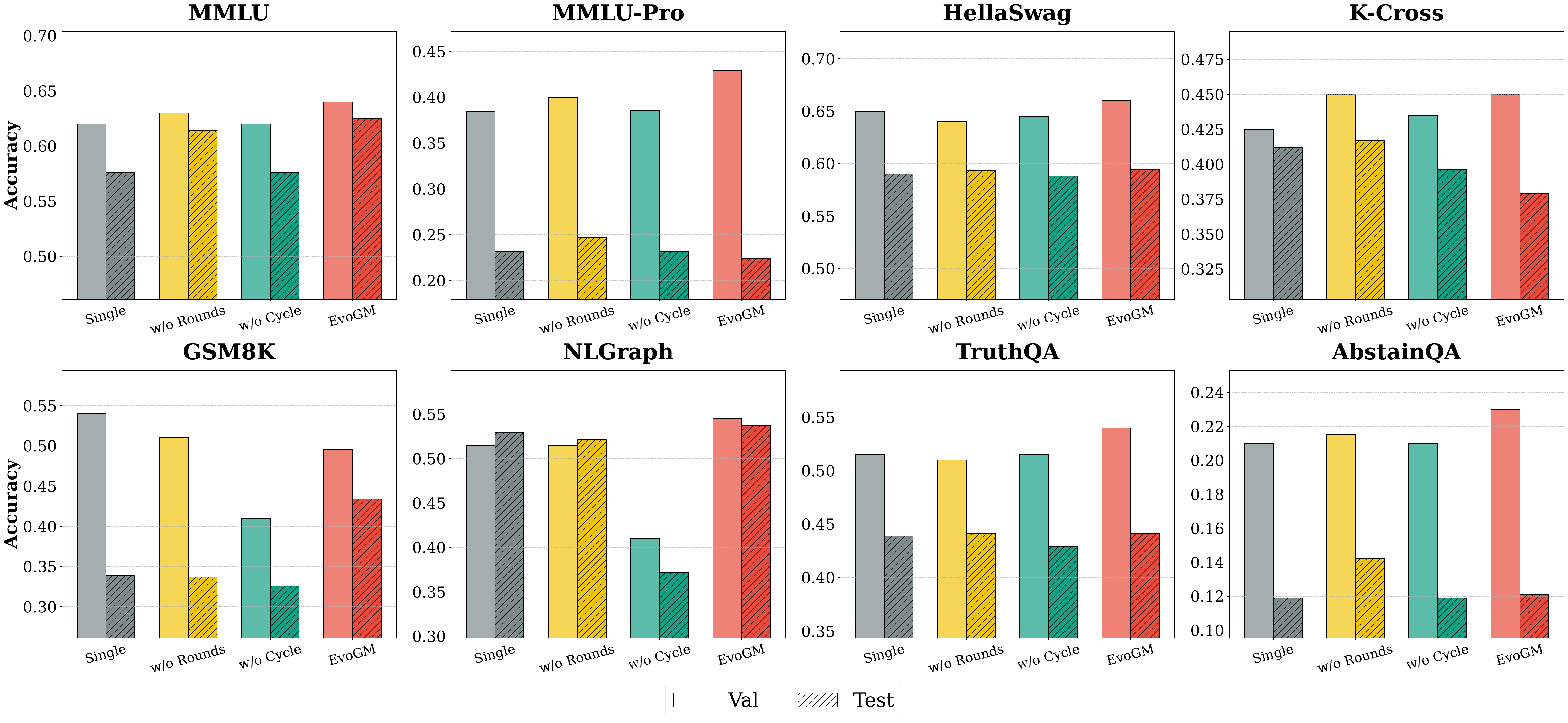}
    \caption{PAblation study on different components. (1) Single-Generator represents the single generative model variant; (2) w/o Rounds represents the replacement of multi-round updates with five continuous iterations; and (3) w/o Cycle Loss represents the removal of the cycle-consistency constraint.}
\end{figure*}
}

\newcommand{\FigureNumberRadarAll}{
\begin{figure*}[!htp]
    \centering
    \includegraphics[width=\linewidth]{figs/number/merging_radar_2x5_grid.pdf}
    \caption{Performance Comparison across Multiple Benchmarks. The radar plots illustrate the performance of different model merging methods on Validation (top row) and Test (bottom row) sets. The benchmarks include HellaSwag, Knowledge Crosswords, MMLU Pro, and MMLU, with the rightmost column showing the overall average performance across all tasks. }
\end{figure*}
}

\begin{document}

\twocolumn[
  \icmltitle{EvoGM: Learning to Merge LLMs via Evolutionary Generative Optimization}



  \icmlsetsymbol{corresponding}{\textdagger}

  \begin{icmlauthorlist}
    \icmlauthor{Tao Jiang}{sustechritas,pcl}
    \icmlauthor{Xinmeng Yu}{sustechritas,pcl}
    \icmlauthor{Chenhao Yi}{ucas,pcl}
    \icmlauthor{Yiling Wu}{pcl}
    \icmlauthor{Yan Li}{pcl}
    \icmlauthor{Ran Cheng}{polyu,polyuszri,polydbti}
    \icmlauthor{Dongmei Jiang}{pcl,corresponding}
    \icmlauthor{Jianguo Zhang}{sustechritas,pcl,brain,corresponding}
  \end{icmlauthorlist}


  \icmlaffiliation{sustechritas}{Research Institute of Trustworthy Autonomous Systems and Department of Computer Science and Engineering, Southern University of Science and Technology}
  \icmlaffiliation{pcl}{Pengcheng Laboratory}
  \icmlaffiliation{ucas}{University of Chinese Academy of Sciences}
  \icmlaffiliation{polyu}{Department of Data Science and Artificial Intelligence, The Hong Kong Polytechnic University}
  \icmlaffiliation{polyuszri}{Hong Kong Polytechnic University Shenzhen Research Institute}
  \icmlaffiliation{polydbti}{Hong Kong Polytechnic University-Daya Bay Technology and Innovation Research Institute}
  \icmlaffiliation{brain}{Guangdong Provincial Key Laboratory of Brain-inspired Intelligent Computation}

  \icmlcorrespondingauthor{Dongmei Jiang}{jiangdm@pcl.ac.cn}
  \icmlcorrespondingauthor{Jianguo Zhang}{zhangjg@sustech.edu.cn}


  \vskip 0.3in
]



\printAffiliationsAndNotice{\textsuperscript{\textdagger}Corresponding authors. }

\begin{abstract}
  Evolutionary model merging provides a powerful framework for the automated, training-free composition of LLMs through parameter-space search. However, existing methods predominantly rely on stochastic, hand-crafted operators that overlook the underlying performance landscape of the coefficient space. We propose Evolutionary Generative Merging (EvoGM), a framework that transcends manual heuristics by employing learnable generative modeling to optimize merging coefficients. Specifically, EvoGM features a dual-generator architecture with cycle-consistent learning to adaptively sample and refine promising merging candidates. By constructing winner-loser pairs from historical search trajectories, our framework effectively captures high-performance parameter distributions and maximizes data efficiency. This generative process is seamlessly integrated into a multi-round evolutionary pipeline, where elite merged models iteratively serve as new expert foundations. Extensive experiments across diverse benchmarks demonstrate that EvoGM significantly outperforms state-of-the-art baselines, exhibiting robust performance on both seen and unseen tasks. Code and data are available at \url{https://github.com/JiangTao97/evogm}.

\end{abstract}

\FigureAbstart

\section{Introduction}
The prevailing paradigm of large language models (LLMs) relies on large-scale pretraining followed by task-specific adaptation, yet the rapid growth in model size makes full-parameter fine-tuning increasingly impractical under realistic computational and data constraints~\cite{Li2021, Hu2022, Lv2024}. In this context, a key challenge is how to accumulate and compose model capabilities without repeatedly training new models. Model merging addresses this challenge by directly integrating multiple expert models into a single one, enabling efficient knowledge transfer and capability composition at no additional training cost, and has therefore emerged as a promising mechanism for scalable and sustainable model improvement~\cite{Yang2026,Lu2024a, Zheng2025}.

\FigureMethod

Most existing model merging methods rely on manually designed heuristics, including linear averaging~\cite{Wortsman2022, Ilharco2023, Tang2023, Yu2024, Jin2025}, spherical interpolation~\cite{Jiang2025}, or the use of scaling and sparsifying schemes to mitigate parameter interference~\cite{Yang2025,Ma2025,Sun2025b}. These approaches are largely task-agnostic and lack the ability to adapt their merging strategies according to validation objectives. To address this limitation, more recent studies have explored optimization- or search-based formulations~\cite{Wang2020,Liu2025,Li2025b}. Furthermore, evolutionary model merging has emerged as a practical alternative, enabling gradient-free population-based search and demonstrating competitive performance across a wide range of tasks~\cite{Akiba2025,Mencattini2025,Inoshita2026}. 

Despite their empirical success, existing evolutionary model merging methods largely operate over an unstructured search space and depend on stochastic perturbations, resulting in low search efficiency under realistic deployment constraints with small validation sets and limited evaluation budgets. 
In optimization problems characterized by similarly sparse and expensive feedback, generative optimization approaches have been shown to substantially reduce search randomness by learning structured proposal distributions from limited evaluative signals~\cite{Gao2024,Wang2025,jiang2026evolutionary}. Motivated by these advances, we argue that model merging should be formulated as a data-driven and learnable search process rather than a purely stochastic one. By treating validation performance as a reward signal, a generative model can capture the distribution of high-performing merging configurations, enabling structured exploration of the merging space and improved generalization.

Building on this insight, we propose Evolutionary Generative Merging (EvoGM), a unified framework that integrates generative modeling into evolutionary model merging. EvoGM formulates the search over merging coefficients as adaptive sampling from a learned probability distribution, replacing static, hand-crafted mutation operators. The framework employs a dual-generator architecture with cycle-consistent constraints to capture the geometric structure of high-performance regions while preserving diversity. To effectively leverage past search experience, we introduce a winner-loser preference mechanism that constructs comparative signals from historical trajectories, enabling the generative model to distinguish promising directions from ineffective ones. Embedded within a multi-round evolutionary pipeline, elite merged models iteratively serve as new expert foundations, allowing the search strategy and model capabilities to co-evolve. Extensive experiments demonstrate that EvoGM achieves significantly improved efficiency and robustness, consistently outperforming prior approaches on both seen and unseen tasks.

Our main contributions are summarized as follows: \begin{itemize} \item We reformulate evolutionary model merging as a learnable generative task, enabling adaptive characterization of high-performance regions. \item We introduce a dual-generator architecture with cycle-consistency and a winner-loser preference strategy to maximize data utilization from sparse feedback, enabling efficient model synthesis for unseen tasks. \item Extensive experiments demonstrate that EvoGM significantly outperforms state-of-the-art baselines across diverse benchmarks and model families. \end{itemize}

\section{Methodology}

\subsection{Problem Formulation}
We formulate model merging as an optimization task that seeks the optimal coefficients for combining task vectors. Let $\boldsymbol{\theta}_{pre} \in \mathbb{R}^d$ denote the parameters of a pre-trained base model, and $\{\boldsymbol{\theta}_i\}_{i=1}^N$ be a set of $N$ expert models fine-tuned from $\boldsymbol{\theta}_{pre}$ for specific tasks. Following the principles of task arithmetic, we define the task vector for each expert as $\boldsymbol{\tau}_i = \boldsymbol{\theta}_i - \boldsymbol{\theta}_{pre}$. The merged model is parameterized by a coefficient vector $\boldsymbol{\lambda} = [\lambda_1, \dots, \lambda_N]^\top \in \mathbb{R}^N$, where the dimensionality of the optimization space corresponds directly to the number of expert models $N$. This results in parameters $\boldsymbol{\theta}(\boldsymbol{\lambda}) = \boldsymbol{\theta}_{pre} + \sum_{i=1}^N \lambda_i \boldsymbol{\tau}_i$. Given a validation set $\mathcal{D}_{val}$ and an evaluation metric $f(\cdot)$, our objective is to find the optimal coefficients $\boldsymbol{\lambda}^* = \arg\max_{\boldsymbol{\lambda}} f(\boldsymbol{\theta}(\boldsymbol{\lambda}); \mathcal{D}_{val})$. In this framework, the optimization is restricted to the low-dimensional vector $\boldsymbol{\lambda} \in \mathbb{R}^N$.

\paragraph{Limitations} Finding $\boldsymbol{\lambda}^*$  is non-trivial. Grid search is computationally prohibitive due to the exponential growth of the search space with $N$, while existing evolutionary methods rely on stochastic perturbations and manual heuristics. These approaches often fail to account for the underlying performance distribution, frequently leading to sub-optimal results in complex merging scenarios.

\begin{tcolorbox}[colback=gray!5, colframe=black, arc=5pt, outer arc=5pt]
\centering
\itshape Can we transcend manual search heuristics by learning to locate optimal merging configurations?
\end{tcolorbox}

\subsection{Evolutionary Generative Merging}

To address the challenge, we propose to transform the search process into a learnable generative task. Instead of directly perturbing $\boldsymbol{\lambda}$ through random mutations, EvoGM characterizes the distribution of optimal merging coefficients by modeling the relationship between historical search trajectories and their corresponding performance on $\mathcal{D}_{val}$. The EvoGM framework is illustrated in Figure~\ref{fig:method1}, with the full procedure detailed in Algorithm~\ref{alg:evogm}.

\algEvoGM

\paragraph{Population Initialization}
The search process begins by initializing a population $\mathcal{P} = \{\boldsymbol{\lambda}^{(p)}\}_{p=1}^P$ of candidate configurations. Throughout this work, we set the population size to $P = 2N$. To ensure diverse coverage of the coefficient space, we employ a hybrid initialization strategy comprising: (i) average merging, where $\lambda_i = 1/N$ for all $i$; (ii) one-hot vectors, representing individual experts; and (iii) random sampling from a uniform distribution. Each candidate $\boldsymbol{\lambda}^{(p)}$ is evaluated on $\mathcal{D}_{val}$ to compute its performance score $f(\boldsymbol{\theta}(\boldsymbol{\lambda}^{(p)}), \mathcal{D}_{val})$. These initial evaluations establish the baseline performance and provide the essential seed trajectories for generative learning.

\paragraph{Winner-Loser Pairing}
In each iteration, we maintain a search history archive
$\mathcal{H}=\{(\boldsymbol{\lambda}, f(\boldsymbol{\lambda}))\}$ that stores all
previously evaluated coefficient configurations and their fitness scores.
We split $\mathcal{H}$ into a winner set $\mathcal{H}^+$ and a loser set
$\mathcal{H}^-$ using a ratio $\rho$ (default $0.3$), where $\mathcal{H}^+$
contains the top-$\rho$ fraction of configurations.
For clarity, we denote samples drawn from the loser and winner sets as
$\boldsymbol{\lambda}^{-}\in\mathcal{H}^{-}$ and
$\boldsymbol{\lambda}^{+}\in\mathcal{H}^{+}$, respectively.

We construct a paired training set via the Cartesian product:
\begin{equation*}
\mathcal{H}^- \times \mathcal{H}^+
=
\{(\boldsymbol{\lambda}^{-}, \boldsymbol{\lambda}^{+})
\mid \boldsymbol{\lambda}^{-}\in\mathcal{H}^{-},
\boldsymbol{\lambda}^{+}\in\mathcal{H}^{+}\}.
\end{equation*}
We form mini-batches by independently sampling
$\boldsymbol{\lambda}^{-}\sim\mathcal{H}^{-}$ and
$\boldsymbol{\lambda}^{+}\sim\mathcal{H}^{+}$, which is equivalent to sampling from
the product distribution.

\paragraph{Train Dual-Generator}
To model transitions between different performance levels in the coefficient
space $\mathbb{R}^N$, we employ a dual-generator architecture consisting of
a forward generator $G_{-\to +}$ (mapping losers to winners) and a backward
generator $G_{+\to -}$ (mapping winners to losers). Both generators are
implemented as five-layer MLPs, with a \texttt{tanh} activation in the output
layer to constrain the generated coefficients $\boldsymbol{\lambda}_{new}$
within the range $[-1,1]$.

\begin{itemize}

\item \textbf{Training Objective.}
The generators are jointly optimized by balancing cycle-consistency and
optimization guidance:
\begin{equation*}
\mathcal{L}_{total}
=
\alpha_{c}\,\mathcal{L}_{cyc}
+
\alpha_{o}\,\mathcal{L}_{opt},
\label{eq:total_loss}
\end{equation*}
where $\alpha_{c}$ and $\alpha_{o}$ control the contributions of
cycle-consistency regularization and optimization guidance, respectively.

\item \textbf{Cycle-Consistency Loss.}
To ensure that the learned mapping is reversible and preserves the structural
distribution of the coefficient population, we impose the following
cycle-consistency constraint:
\begin{align*}
\mathcal{L}_{cyc}
=&\;
\mathbb{E}_{\boldsymbol{\lambda}^{-}\in\mathcal{H}^{-}}\!
\Big[
\big\|
G_{+\to -}(G_{-\to +}(\boldsymbol{\lambda}^{-}))
-
\boldsymbol{\lambda}^{-}
\big\|_2^2
\Big]
\nonumber\\
&+
\mathbb{E}_{\boldsymbol{\lambda}^{+}\in\mathcal{H}^{+}}\!
\Big[
\big\|
G_{-\to +}(G_{+\to -}(\boldsymbol{\lambda}^{+}))
-
\boldsymbol{\lambda}^{+}
\big\|_2^2
\Big].
\label{eq:cycle_loss}
\end{align*}
Minimizing $\mathcal{L}_{cyc}$ regularizes $G_{-\to +}$ to capture the
geometric relationship between $\mathcal{H}^-$ and $\mathcal{H}^+$ and
prevents trivial mode collapse.

\item \textbf{Optimization-Guided Loss.}
Beyond consistency, we explicitly guide the generative search toward
high-performance regions. Let the centroid of the winner set be
\begin{equation*}
\boldsymbol{\mu}^+
=
\frac{1}{|\mathcal{H}^+|}
\sum_{\boldsymbol{\lambda}^{+}\in\mathcal{H}^{+}}
\boldsymbol{\lambda}^{+}.
\label{eq:winner_centroid}
\end{equation*}
The optimization loss is defined as
\begin{equation*}
\mathcal{L}_{opt}
=
\mathbb{E}_{\boldsymbol{\lambda}^{-}\in\mathcal{H}^{-}}
\Big[
\big\|
G_{-\to +}(\boldsymbol{\lambda}^{-})
-
\boldsymbol{\mu}^+
\big\|_2^2
\Big].
\label{eq:opt_loss}
\end{equation*}
This objective drives $G_{-\to +}$ to transform sub-optimal configurations
toward the statistical average of elite solutions, thereby accelerating the
discovery of the global optimum $\boldsymbol{\lambda}^*$.

\end{itemize}

\paragraph{Generative Evolution and Selection}
With the dual-generators optimized, the forward generator $G_{-\to +}$ serves as a learned evolutionary operator. In each iteration $t$, we generate a set of new candidates by applying the learned mapping to the current population $\mathcal{P}$:
$
\boldsymbol{\lambda}_{new} = G_{-\to +}(\boldsymbol{\lambda}), \quad \forall \boldsymbol{\lambda} \in \mathcal{P}.
$
By transforming the entire population through $G_{-\to +}$, the framework shifts the candidate distribution toward regions with higher performance potential. These new candidates are subsequently evaluated on $\mathcal{D}_{val}$ to obtain their fitness $\mathcal{F}_{new}$. We then update the search history $\mathcal{H} \leftarrow \mathcal{H} \cup \{(\boldsymbol{\lambda}_{new}, \mathcal{F}_{new})\}$ and refine the population $\mathcal{P}$ by selecting the top-$P$ individuals based on their fitness scores. This process ensures that the population continuously migrates toward higher-performance regions within the coefficient space.

\paragraph{Evolving Expert Basis}
A distinctive feature of EvoGM is that the expert pool itself evolves across outer rounds, which we refer to as a \emph{basis shift}. While the inner loop optimizes the merging coefficients $\boldsymbol{\lambda}$ over a fixed set of task vectors, the outer loop explicitly redefines the underlying expert basis by updating the task vectors $\{\boldsymbol{\tau}_i\}$.
At the end of each round ($r < R$), we select the top-$N$ coefficient vectors
$\{\boldsymbol{\lambda}^{(i)}\}_{i=1}^N$ from the accumulated history archive
$\mathcal{H}$. These elite configurations are then used to synthesize a new
expert pool for the next round:
$\boldsymbol{\theta}^{(r+1)}_i = \boldsymbol{\theta}\!\left(\boldsymbol{\lambda}^{(i)}\right),
\quad i = 1,\dots,N,$
where $\boldsymbol{\theta}(\boldsymbol{\lambda}^{(i)})$ denotes the merged model
constructed by applying $\boldsymbol{\lambda}^{(i)}$ to the experts of round
$r$. We subsequently recompute task vectors with respect to the pretrained base
model:
$\boldsymbol{\tau}^{(r+1)}_i = \boldsymbol{\theta}^{(r+1)}_i - \boldsymbol{\theta}_{pre}.$
This basis shift enables EvoGM to progressively refine the expert directions
available for merging, rather than searching over a static expert pool.

\section{Experimental Setup}
We study model merging under two settings: merging models trained on seen tasks and merging models evaluated on unseen tasks. 
Here, seen tasks refer to tasks on which the expert models are explicitly fine-tuned, while unseen tasks denote tasks that are not used during expert fine-tuning.
Experiments span two model architectures, FLAN-T5~\cite{Chung2024} and Qwen~\cite{Bai2023}, different merging scales (8 and 10 models), and both single-task and multi-task configurations.

\TableSingletask

\paragraph{Model Configuration}
For the seen tasks setting, we use FLAN-T5-base\footnote{\scriptsize\url{https://huggingface.co/google/flan-t5-base}} together with 8 expert models fine-tuned on single corresponding tasks.
For the unseen tasks setting, we construct a set of ten domain-specific experts\footnote{\scriptsize\url{https://huggingface.co/TaoJiangCN/qwen2.5-1.5b-tulu-v2-lora-experts}} by independently fine-tuning Qwen2.5-1.5B \footnote{\scriptsize\url{https://huggingface.co/Qwen/Qwen2.5-1.5B}} on each of the ten supervised fine-tuning (SFT) domains in Tulu-v2\footnote{\scriptsize\url{https://huggingface.co/collections/allenai/tulu-v2-suite}}~\cite{Ivison2023}.
All experts are adapted using LoRA-based parameter-efficient fine-tuning~\cite{Hu2022}. 
Each model is trained for five epochs with an initial learning rate of $2\times10^{-4}$ and an effective batch size of 32.

\paragraph{Baselines}
For the seen tasks setting, we compare against a broad range of representative merging approaches, including TA~\cite{Ilharco2023}, DARE~\cite{Yu2024a}, TIES~\cite{Yadav2023a}, DARE-TIES~\cite{Yadav2023a, Yu2024a}, DELLA-Merging~\cite{Deep2024}, RankMean~\cite{Perin2024}, CMA~\cite{Akiba2025}, AdaMerging~\cite{Yang2024a}, Fisher-Merging~\cite{Matena2022}, and RegMean~\cite{Jin2023}.
For the unseen tasks setting, we consider both training-based and merging-based baselines. 
These include the base pretrained model (Base), multi-task learning (MTL), the best single expert (Single Best), and Model Soup~\cite{Wortsman2022}, as well as merging methods such as TA, TIES, DARE, CMA, PSO-Merging~\cite{Zhang2025}, and Model Swarm~\cite{feng2025model}.

\paragraph{Benchmarks and Tasks}
For the seen tasks setting, we evaluate model merging on a suite of eight text-to-text generation tasks drawn from the GLUE benchmark~\cite{Wang2018}, covering grammaticality judgment, natural language inference, paraphrase identification, sentiment analysis, and semantic similarity.
The tasks include CoLA, MNLI, MRPC, QNLI, QQP, RTE, SST-2, and STS-B
\footnote{\scriptsize\url{https://huggingface.co/collections/tanganke/flan-t5-base-models-fine-tuned-on-glue-benchmark}}.

For the unseen tasks setting, we assess generalization on eight datasets spanning three categories: knowledge, reasoning, and safety. 
The knowledge category includes MMLU~\cite{Hendrycks2021}, MMLU-Pro~\cite{Wang2024}, and HellaSwag~\cite{Zellers2019}; reasoning tasks cover GSM8K~\cite{Cobbe2021}, Knowledge Crosswords~\cite{Ding2024}, and NLGraph~\cite{Wang2023,Zhang2024}; and safety-related evaluation is conducted on TruthfulQA~\cite{Lin2021}, and AbstainQA~\cite{Gehman2020}. 
Unless otherwise specified, we randomly sample 200 examples for validation and 1,000 examples for testing on each dataset.

\subsection{Seen Tasks}

We evaluate the performance of merging 8 FLAN-T5 models on their corresponding tasks. Baseline results are taken from PSO-Merging for comparison.
The multi-task performance on the GLUE benchmark is summarized in \cref{tab:seen-task}. Our proposed method consistently achieves the highest average accuracy among all tested approaches, providing a relative improvement of over 1.4\% compared to the previous state-of-the-art, PSO-Merging. In semantic similarity (STS-B), our approach yields a relative performance gain of more than 12\% over PSO-Merging and approximately 6\% over the best-performing traditional merging baseline. 

\TableSeenTask

\TableMultitask

\subsection{Unseen Tasks}
Merging on unseen tasks effectively benchmarks generalization under data scarcity. By optimizing weights on small validation sets, evolutionary methods support both single-task specialization (tailoring weights to specific task) and multi-task merging (identifying a unified model for all tasks simultaneously).

\paragraph{Single-task}
We evaluate the merging of 10 Qwen2.5-1.5B experts across 8 unseen tasks by optimizing dedicated weight configurations for each task independently.
As shown in \cref{tab:singletask} and Figure~\ref{fig:performance_single}, EvoGM achieves the highest test accuracy on 5 of 8 benchmarks. For instance, a 15\% relative performance gain is observed in NLGraph compared to PSO-Merging. Notably, multi-task learning and traditional merging methods like TIES or DARE often fail to outperform the single best expert. This indicates that static merging schemes or joint training struggle to transfer specialized knowledge to new domains. In contrast, evolutionary methods such as CMA and PSO-Merging achieve better results by searching for optimal weights on validation sets. EvoGM further advances these baselines by incorporating learnable generative modeling, achieving superior and more robust performance across complex reasoning and knowledge tasks.


\paragraph{Multi-task} We evaluate the generalization of a single merged model across 8 unseen tasks simultaneously. As shown in \cref{tab:multitask}, EvoGM achieves a leading average score of 0.380 among all merging approaches. Notably, standard multi-task fine-tuning yields a score of 0.330, which falls below the base model performance of 0.352. This performance drop underscores the severe inter-task interference inherent in joint fine-tuning across diverse domains. While evolutionary baselines such as CMA and PSO-Merging improve results through weight search, their efficacy is often constrained by the stochasticity of their operators. Conversely, EvoGM navigates the high-performance parameter space more effectively through its generative architecture. This advantage is especially pronounced in AbstainQA, where EvoGM exhibits a 66\% relative improvement over the strongest evolutionary baseline. These results demonstrate that by leveraging historical search trajectories, our framework discovers balanced weight combinations that successfully mitigate task conflicts in few-shot scenarios.

\FloatBarrier

\section{Analysis}\label{subsec:analysis}

\paragraph{Convergence and Optimization Efficiency}
We analyze the convergence of EvoGM by comparing it with Model Swarm and PSO-Merging. All methods are evaluated using a population size of 20. To visualize the optimization process, we plot the average performance of the top 5 solutions from each iteration. 

As shown in Figure~\ref{fig:convergence}, EvoGM consistently outperforms the baselines across all tasks. The baseline algorithms typically reach a performance plateau around iteration 5, showing little improvement thereafter. In contrast, EvoGM follows a multi-round optimization strategy, configured here with 2 rounds and 3 iterations per round. A noticeable performance shift occurs at iteration 3, which marks the transition between rounds. At this stage, the framework performs a basis shift by generating new parameter candidates based on the elite models from the previous round. While this re-initialization causes a temporary fluctuation, it effectively prevents the search from getting trapped in local optima. This phase allows EvoGM to explore more promising regions of the parameter space, leading to substantial performance gains in the subsequent round.

\FigureConvergence

\paragraph{Ablation Study}\label{subsec:ablation}
We conduct an ablation study to evaluate the contribution of each core component in EvoGM. This experiment involves merging 4 expert models with a population size of 8. The full framework is configured with 2 rounds and 2 iterations per round. We compare the full version against 3 variants: (1) Single-Generator, which uses only a single generative model; (2) w/o Rounds, which replaces the multi-round expert update with 5 continuous iterations on the initial experts; and (3) w/o Cycle Loss, which removes the cycle-consistency constraint. Results are summarized in Figure~\ref{fig:ablation} and \cref{tab:ablation} in Appendix.

Each component is shown to be critical for achieving optimal performance across diverse benchmarks. The removal of the multi-round mechanism (w/o Rounds) leads to a noticeable decline in complex tasks such as NLGraph and MMLU. This confirms that iteratively updating the expert foundations allows the model to escape local optima and integrate knowledge more deeply. Furthermore, the variant without cycle loss (w/o Cycle Loss) suffers a significant performance drop, particularly in reasoning tasks like NLGraph. This suggests that the cycle-consistent constraint is vital for accurately capturing the high-performance regions of the parameter space. Finally, the dual-generator architecture consistently outperforms the single-generator variant, proving that bidirectional learning of winner-loser pairs more effectively identifies optimal merging coefficients.

\FigureAblation

\paragraph{Hyperparameter Sensitivity}
We evaluate the sensitivity of EvoGM to key hyperparameters by merging 3 models and reporting the average performance across 2 tasks. As summarized in \cref{tab:parameter-tuning} of the Appendix, the population size exhibits a significant influence on search quality. Increasing the population from 10 to 30 leads to a 0.0305 gain in test accuracy. This suggests that a larger number of candidates allows for a more thorough exploration of the high-dimensional parameter manifold. In contrast, the number of iterations per round has a relatively stable effect, with a slight performance increase observed as iterations scale from 3 to 8.

Regarding the training of the generative model, we find that 200 to 400 epochs are sufficient. Increasing the epochs to 600 causes a minor performance drop of 0.0035, likely due to the generator over-fitting on specific historical search trajectories. For the learning rate, a smaller value of 0.0001 yields the best results, providing a 0.0220 improvement over the baseline. Finally, increasing the evolution rounds from 1 to 5 consistently enhances performance. This confirms that the iterative multi-round process is essential for refining expert foundations and achieving higher generalization.

\paragraph{Impact of Model Scale} We investigate the scalability of EvoGM by varying the number of merged experts from 2 to 10 in a multi-task setting. Results are summarized in Figure~\ref{fig:merging_trends} and \cref{tab:ensemble_results} in Appendix, where we compare the initial performance against the final optimized results.

\FigureMergingTrends

Our framework consistently improves average accuracy across all model counts. When merging 8 models, the average test score increases from 0.433 to 0.460. This trend indicates that EvoGM is highly robust to the initial configuration and the complexity of the weight space. Even as the number of experts increases, the generative evolutionary process successfully identifies superior merging coefficients that mitigate task interference. These results demonstrate that our method scales effectively, maintaining consistent performance gains regardless of whether a small ensemble or a larger set of 10 experts is used. This stability proves that EvoGM effectively navigates the high-dimensional parameter space across various scales of model composition.

\paragraph{Extended Scaling and Robustness Experiments}
We further examine EvoGM under larger model and expert-space settings, with full results reported in Appendix~\ref{app:rebuttal_results}. To test model-scale transfer, we merge ten Tulu-finetuned experts built on Qwen3-8B; EvoGM achieves the best average score among all compared methods (0.603) while maintaining competitive wall-clock search cost. To test scalability with respect to the number of experts, we additionally evaluate a 20-expert ViT-B-16 merging setting under the same validation/test split and search budget as CMA-ES, where EvoGM obtains a higher average test score (0.6363 vs. 0.6268). Finally, multi-seed experiments on Qwen2.5-1.5B show statistically significant average improvements over PSO-Merging ($p=0.0100$) and Model Swarm ($p=0.0006$), indicating that the performance gains are not due to random variation.

\section{Related Work}
\paragraph{\textbf{Model Merging}}

Model merging facilitates the integration of diverse expert models into a unified framework without additional training costs~\cite{Yang2026}. Early research demonstrates that models fine-tuned from the same initialization can be linearly combined when they reside in compatible loss basins~\cite{Wortsman2022,Ilharco2023}. However, direct weight averaging often suffers from parameter interference. To address this, subsequent studies propose conflict-aware mechanisms, such as sign-based trimming (TIES)~\cite{Yadav2023a}, importance-aware pruning and rescaling~\cite{Guodong2024}, and the selective removal of redundant or conflicting components~\cite{Sun2025}. More recently, these techniques have been extended to Parameter-Efficient Fine-Tuning (PEFT) and multimodal settings~\cite{li2025multimodal}, including LoRA-style merging~\cite{Panariello2025} and dynamic routing frameworks for heterogeneous architectures~\cite{Lu2024,Chen2025a}.

Beyond direct parameter manipulation, another research thrust focuses on improving model mergeability through representation-level synchronization. This includes resolving neuron permutation symmetries~\cite{Crisostomi2024}, enforcing dual-space consistency~\cite{Xu2024}, and performing subspace alignment or activation-level corrections~\cite{Horoi2024,Sun2025a,Yao2025,shao2026pivotmerge}. To further enhance robustness across heterogeneous task distributions, iterative merging strategies have been proposed to refine weights through multi-stage compositions~\cite{Tang2025,Yuan2025}. 

While early methods rely on hand-crafted heuristics, recent empirical evidence suggests that performance is often governed by the precision of merging coefficients rather than the specific operators used~\cite{Lan2025}. This shift in perspective has motivated the use of formal optimization to determine optimal blending weights. Current approaches employ entropy-based objectives~\cite{Yang2024a}, Bayesian inference~\cite{Li2025}, and Pareto-aware multi-objective designs~\cite{Li2024,Zhou2025,Chen2025} to explore the weight space. However, these methods often rely on stochastic search or static probabilistic priors, leaving a gap for more adaptive, generative search mechanisms that can effectively capture high-performance parameter configurations.

\paragraph{\textbf{Evolutionary Model Merging}}

A burgeoning line of research reframes model merging as a gradient-free optimization challenge, moving away from manual heuristic design. Evolutionary Algorithms (EAs) have become a preferred paradigm due to their flexibility in handling non-differentiable objectives. One category focuses on optimizing merging hyperparameters, such as layer-wise coefficients and weight sparsity ratios~\cite{Akiba2025, Baba2024}. Alternatively, population-centric methods evolve the models themselves, employing mutation and crossover operators to generate novel architectural candidates~\cite{Du2024}. More recent swarm-intelligence frameworks harness collaborative search trajectories to identify optimal weight configurations in the parameter space~\cite{Zhang2025, feng2025model}.


Despite their potential, evolutionary model merging methods are often bottlenecked by the prohibitive computational cost of evaluating large-scale models. Scalable evolutionary-computation platforms such as EvoX ~\cite{huang2024evox} have improved the practicality and programmability of evolutionary search, and efficiency-oriented merging frameworks further mitigate evaluation costs through surrogate models or sparse validation subsets~\cite{Mencattini2025, Akizuki2025}. Yet these advances mainly concern the execution or evaluation of search, leaving the generation of candidate merging configurations largely governed by generic stochastic operators. These operators often lack the structural awareness needed to navigate high-dimensional merging landscapes, leading to poor sample efficiency under limited evaluation budgets.


\section{Conclusion and Limitations}
\label{app:discussion_limitations}

\textbf{Conclusion.}
This paper presents EvoGM, a framework that reformulates model merging as a learnable generative task. By integrating dual-generator learning with a hierarchical basis shift mechanism, EvoGM effectively navigates the complex performance landscape of large language models without relying on manual heuristics. Our experiments demonstrate that by learning from historical search trajectories, EvoGM achieves superior data efficiency and discovers high-synergy model configurations that outperform current state-of-the-art baselines across various benchmarks.

\textbf{Limitations.}
The current study focuses on homogeneous task-vector merging, where all expert models are obtained from the same pretrained base model. This setting makes the linear coefficient space well-defined, but it also limits the direct use of EvoGM in more general merging scenarios. For models with different architectures, initializations, or parameter structures, additional alignment is needed before coefficient-based search can be applied. Moreover, our multi-task setting optimizes a single scalar validation objective and returns one deployable merged model. This is suitable when a fixed task balance is given, but it does not directly provide a set of Pareto-optimal models for users with different preferences. Extending EvoGM to heterogeneous model merging and preference-aware multi-objective merging is therefore an important direction for future work.

\section*{Acknowledgements}
This work is supported in part by National Natural Science Foundation of China (Grant No. 62276121), in part by the TianYuan funds for Mathematics of the National Science Foundation of China (Grant No. 12326604), in part by Innovation Team and Talents Cultivation Program of National Administration of Traditional Chinese Medicine (No: ZYYCXTD-D-202403),  in part by Guangdong Basic and Applied Basic Research Foundation (No. 2024B1515020019), in part by National Natural Science Foundation of China (Grant No. 62502246).

\section*{Impact Statement}

This work studies training-free model merging for composing multiple fine-tuned models. By learning to propose high-performing merge coefficients from historical search trajectories, EvoGM can reduce the need to retrain full models or maintain many separate expert models. This makes model adaptation more practical when compute, validation data, or engineering resources are limited. More broadly, efficient model merging can encourage the reuse of existing task-specialized checkpoints, improve the accessibility of customized AI systems, and support more sustainable model development by amortizing prior fine-tuning costs across multiple downstream tasks.

The societal effects of EvoGM are therefore expected to be largely aligned with those of efficient model reuse and composition. As with other model-merging methods, a merged model may inherit biases, privacy concerns, or unsafe behaviors from its source models, so appropriate source-model selection and task-relevant evaluation remain important. 


\bibliography{evoGM}

@InProceedings{Wortsman2022,
  author    = {Mitchell Wortsman and Gabriel Ilharco and Samir Yitzhak Gadre and Rebecca Roelofs and Raphael Gontijo Lopes and Ari S. Morcos and Hongseok Namkoong and Ali Farhadi and Yair Carmon and Simon Kornblith and Ludwig Schmidt},
  booktitle = {International conference on machine learning},
  title     = {Model soups: averaging weights of multiple fine-tuned models improves accuracy without increasing inference time},
  year      = {2022},
  pages     = {23965--23998},
  cdate     = {1640995200000},
  comment   = {The conventional recipe for maximizing model accuracy is to (1) train multiple models with various hyperparameters and (2) pick the individual model which performs best on a held-out validation set, discarding the remainder. In this paper, we revisit the second step of this procedure in the context of fine-tuning large pre-trained models, where fine-tuned models often appear to lie in a single low error basin. We show that averaging the weights of multiple models fine-tuned with different hyperparameter configurations often improves accuracy and robustness. Unlike a conventional ensemble, we may average many models without incurring any additional inference or memory costs -- we call the results "model soups." When fine-tuning large pre-trained models such as CLIP, ALIGN, and a ViT-G pre-trained on JFT, our soup recipe provides significant improvements over the best model in a hyperparameter sweep on ImageNet. The resulting ViT-G model, which attains 90.94% top-1 accuracy on ImageNet, achieved a new state of the art. Furthermore, we show that the model soup approach extends to multiple image classification and natural language processing tasks, improves out-of-distribution performance, and improves zero-shot performance on new downstream tasks. Finally, we analytically relate the performance similarity of weight-averaging and logit-ensembling to flatness of the loss and confidence of the predictions, and validate this relation empirically. Code is available at [this https URL](https://github.com/mlfoundations/model-soups).},
  url       = {https://proceedings.mlr.press/v162/wortsman22a.html},
}

@InProceedings{Ma2025,
  author    = {Qianli Ma and Dongrui Liu and Qian Chen and Linfeng Zhang and Jing Shao},
  booktitle = {Proceedings of the 63rd Annual Meeting of the Association for Computational Linguistics (Volume 1: Long Papers)},
  title     = {{LED-Merging}: Mitigating Safety-Utility Conflicts in Model Merging with Location-Election-Disjoint},
  year      = {2025},
  pages     = {21749-21767},
  cdate     = {1735689600000},
  url       = {https://aclanthology.org/2025.acl-long.1055/},
}

@article{huang2024evox,
  title={EvoX: A distributed GPU-accelerated framework for scalable evolutionary computation},
  author={Huang, Beichen and Cheng, Ran and Li, Zhuozhao and Jin, Yaochu and Tan, Kay Chen},
  journal={IEEE Transactions on Evolutionary Computation},
  year={2024},
  publisher={IEEE}
}

@inproceedings{feng2025model,
  title={Model Swarms: Collaborative Search to Adapt LLM Experts via Swarm Intelligence},
  author={Feng, Shangbin and Wang, Zifeng and Wang, Yike and Ebrahimi, Sayna and Palangi, Hamid and Miculicich, Lesly and Kulshrestha, Achin and Rauschmayr, Nathalie and Choi, Yejin and Tsvetkov, Yulia and others},
  booktitle={International Conference on Machine Learning},
  pages={16904--16930},
  year={2025},
  organization={PMLR}
}

@InProceedings{Horoi2024,
  author    = {Stefan Horoi and Albert Manuel Orozco Camacho and Eugene Belilovsky and Guy Wolf},
  booktitle = {Forty-first International Conference on Machine Learning},
  title     = {Harmony in Diversity: Merging Neural Networks with Canonical Correlation Analysis},
  year      = {2024},
  comment   = {Combining the predictions of multiple trained models through ensembling is generally a good way to improve accuracy by leveraging the different learned features of the models, however it comes with high computational and storage costs. Model fusion, the act of merging multiple models into one by combining their parameters reduces these costs but doesn't work as well in practice. Indeed, neural network loss landscapes are high-dimensional and non-convex and the minima found through learning are typically separated by high loss barriers. Numerous recent works have been focused on finding permutations matching one network features to the features of a second one, lowering the loss barrier on the linear path between them in parameter space. However, permutations are restrictive since they assume a one-to-one mapping between the different models' neurons exists. We propose a new model merging algorithm, CCA Merge, which is based on Canonical Correlation Analysis and aims to maximize the correlations between linear combinations of the model features. We show that our alignment method leads to better performances than past methods when averaging models trained on the same, or differing data splits. We also extend this analysis into the harder setting where more than 2 models are merged, and we find that CCA Merge works significantly better than past methods. Our code is publicly available at [this https URL](https://github.com/shoroi/align-n-merge)},
  url       = {https://openreview.net/forum?id=hLuNVjRnY3},
}

@inproceedings{Mencattini2025,
  title     = {MERGE$^3$: Efficient Evolutionary Merging on Consumer-grade GPUs},
  author    = {Mencattini, Tommaso and Minut, Adrian Robert and Crisostomi, Donato and Santilli, Andrea and Rodolà, Emanuele},
  year      = 2025,
  booktitle = {Proceedings of the 42nd International Conference on Machine Learning (ICML)},
  url       = {https://openreview.net/pdf?id=qFXDv0X4yc},
}

@InProceedings{Zhou2025,
  author    = {Yu Zhou and Xingyu Wu and Jibin Wu and Liang Feng and KC Tan},
  booktitle = {The Thirty-ninth Annual Conference on Neural Information Processing Systems},
  title     = {{HM}3: Hierarchical Multi-Objective Model Merging for Pretrained Models},
  year      = {2025},
  comment   = {Model merging is a technique that combines multiple large pretrained models into a single model with enhanced performance and broader task adaptability. It has gained popularity in large pretrained model development due to its ability to bypass the need for original training data and further training processes. However, most existing model merging approaches focus solely on exploring the parameter space, merging models with identical architectures. Merging within the architecture space, despite its potential, remains in its early stages due to the vast search space and the challenges of layer compatibility. This paper marks a significant advance toward more flexible and comprehensive model merging techniques by modeling the architecture-space merging process as a reinforcement learning task. We train policy and value networks using offline sampling of weight vectors, which are then employed for the online optimization of merging strategies. Moreover, a multi-objective optimization paradigm is introduced to accommodate users' diverse task preferences, learning the Pareto front of optimal models to offer customized merging suggestions. Experimental results across multiple tasks, including text translation, mathematical reasoning, and code generation, validate the effectiveness and superiority of the proposed framework in model merging. The code will be made publicly available after the review process.},
  url       = {https://openreview.net/forum?id=JeP0lpusYw},
}

@Article{Akiba2025,
  author  = {Akiba, Takuya and Shing, Makoto and Tang, Yujin and Sun, Qi and Ha, David},
  journal = {Nature Machine Intelligence},
  title   = {Evolutionary optimization of model merging recipes},
  year    = {2025},
  issn    = {2522-5839},
  month   = {feb},
  number  = {2},
  pages   = {195--204},
  volume  = {7},
  doi     = {10.1038/s42256-024-00975-8},
  url     = {https://doi.org/10.1038/s42256-024-00975-8},
}

@article{li2025multimodal,
  title={Multimodal PEAR chain-of-thought reasoning for multimodal sentiment analysis},
  author={Li, Yan and Lan, Xiangyuan and Chen, Haifeng and Lu, Ke and Jiang, Dongmei},
  journal={ACM Transactions on Multimedia Computing, Communications and Applications},
  volume={20},
  number={9},
  pages={1--23},
  year={2025},
  publisher={ACM New York, NY}
}

@InProceedings{Du2024,
  author    = {Guodong Du and Jing Li and Hanting Liu and Runhua Jiang and Shuyang Yu and Yifei Guo and Sim Kuan Goh and Ho-Kin Tang},
  booktitle = {Findings of the Association for Computational Linguistics: ACL 2024},
  title     = {Knowledge Fusion By Evolving Weights of Language Models},
  year      = {2024},
  volume    = {abs/2406.12208},
  cdate     = {1704067200000},
  journal   = {Findings of the Association for Computational Linguistics: ACL 2024},
  publtype  = {informal},
  url       = {https://doi.org/10.48550/arXiv.2406.12208},
}

@InProceedings{Yao2025,
  author       = {Yuxuan Yao and Shuqi LIU and Zehua Liu and Qintong Li and Mingyang LIU and Xiongwei Han and Zhijiang Guo and Han Wu and Linqi Song},
  booktitle    = {The Thirty-ninth Annual Conference on Neural Information Processing Systems},
  title        = {Activation-Guided Consensus Merging for Large Language Models},
  year         = {2025},
  comment      = {Recent research has increasingly focused on reconciling the reasoning capabilities of System 2 with the efficiency of System 1. While existing training-based and prompt-based approaches face significant challenges in terms of efficiency and stability, model merging emerges as a promising strategy to integrate the diverse capabilities of different Large Language Models (LLMs) into a unified model. However, conventional model merging methods often assume uniform importance across layers, overlooking the functional heterogeneity inherent in neural components. To address this limitation, we propose \\textbf{A}ctivation-Guided \\textbf{C}onsensus \\textbf{M}erging (\\textbf{ACM}), a plug-and-play merging framework that determines layer-specific merging coefficients based on mutual information between activations of pre-trained and fine-tuned models. ACM effectively preserves task-specific capabilities without requiring gradient computations or additional training. Extensive experiments on Long-to-Short (L2S) and general merging tasks demonstrate that ACM consistently outperforms all baseline methods. For instance, in the case of Qwen-7B models, TIES-Merging equipped with ACM achieves a \\textbf{55.3\\%} reduction in response length while simultaneously improving reasoning accuracy by \\textbf{1.3} points.},
  comment-dell = {近期的研究愈发注重将系统 2 的推理能力与系统 1 的高效性相协调。尽管现有的基于训练和基于提示的方法在效率和稳定性方面面临诸多挑战，但模型合并作为一种整合不同大型语言模型（LLMs）多样化能力的有前景的策略应运而生，它能将这些模型的多种能力整合到一个统一的模型中。然而，传统的模型合并方法往往假定各层的重要性是统一的，而忽略了神经组件中固有的功能异质性。为解决这一局限性，我们提出了“激活引导共识合并”（ACM）这一可插拔的合并框架，该框架根据预训练模型和微调模型的激活之间的互信息来确定层特定的合并系数。ACM 有效地保留了任务特定的能力，无需进行梯度计算或额外的训练。在长到短（L2S）和一般合并任务上的大量实验表明，ACM 持续优于所有基准方法。例如，对于 Qwen-7B 模型而言，配备 ACM 的 TIES-Merging 技术在缩短响应长度方面实现了约 55.3% 的降幅，同时推理准确率也提高了 1.3 个百分点。},
  url          = {https://openreview.net/forum?id=ayzWTxb9ZD},
}

@InProceedings{Chen2025,
  author    = {Weiyu Chen and James Kwok},
  booktitle = {Forty-second International Conference on Machine Learning},
  title     = {{Pareto Merging}: Multi-Objective Optimization for Preference-Aware Model Merging},
  year      = {2025},
  comment   = {Model merging, which combines multiple models into a single model, has gained popularity in recent years. By efficiently integrating the capabilities of various models, this significantly reduces the parameter count and memory usage. However, current methods can only produce one single merged model. This necessitates a performance trade-off due to conflicts among the various models, and the resultant one-size-fits-all model may not align with the preferences of different users who may prioritize certain models over others. To address this issue, we propose preference-aware model merging, and formulate this as a multi-objective optimization problem in which the performance of the merged model on each base model's task is treated as an objective. In a single merging process, the proposed parameter-efficient structure generates a Pareto set of merged models, with each representing a Pareto-optimal solution for a preference. Users can then select merged models tailored to their preferences from this learned Pareto set. Experimental results demonstrate that the proposed Pareto Merging produces diverse trade-off models and achieves higher test accuracy compared to state-of-the-art merging baselines.},
  url       = {https://openreview.net/forum?id=D7qRwx6BOS},
}

@InProceedings{Lu2024,
  author       = {Zhenyi Lu and Chenghao Fan and Wei Wei and Xiaoye Qu and Dangyang Chen and Yu Cheng},
  booktitle    = {The Thirty-eighth Annual Conference on Neural Information Processing Systems},
  title        = {{Twin-Merging}: Dynamic Integration of Modular Expertise in Model Merging},
  year         = {2024},
  comment      = {In the era of large language models, model merging is a promising way to combine multiple task-specific models into a single multitask model without extra training. However, two challenges remain: (a) interference between different models and (b) heterogeneous data during testing. Traditional model merging methods often show significant performance gaps compared to fine-tuned models due to these issues. Additionally, a one-size-fits-all model lacks flexibility for diverse test data, leading to performance degradation. We show that both shared and exclusive task-specific knowledge are crucial for merging performance, but directly merging exclusive knowledge hinders overall performance. In view of this, we propose Twin-Merging, a method that encompasses two principal stages: (1) modularizing knowledge into shared and exclusive components, with compression to reduce redundancy and enhance efficiency; (2) dynamically merging shared and task-specific knowledge based on the input. This approach narrows the performance gap between merged and fine-tuned models and improves adaptability to heterogeneous data. Extensive experiments on 20 datasets for both language and vision tasks demonstrate the effectiveness of our method, showing an average improvement of 28.34\\% in absolute normalized score for discriminative tasks and even surpassing the fine-tuned upper bound on the generative tasks. Our implementation is available in \\url{[this https URL](https://github.com/LZY-the-boys/Twin-Merging)}},
  comment-dell = {在大型语言模型的时代，模型合并是一种将多个针对特定任务的模型整合为一个多任务模型的有前景的方法，无需额外的训练。然而，仍存在两个挑战：（a）不同模型之间的干扰以及（b）测试期间的异构数据。传统的模型合并方法由于这些问题往往与经过微调的模型相比存在显著的性能差距。此外，一刀切的模型缺乏对不同测试数据的灵活性，导致性能下降。我们表明，共享和专属的任务特定知识对于合并性能至关重要，但直接合并专属知识会阻碍整体性能。鉴于此，我们提出了 Twin-Merging 方法，该方法包含两个主要阶段：（1）将知识模块化为共享和专属组件，并进行压缩以减少冗余并提高效率；（2）根据输入动态合并共享和任务特定的知识。这种方法缩小了合并模型和微调模型之间的性能差距，并提高了对异构数据的适应性。在 20 个语言和视觉任务的数据集上进行的大量实验表明，我们的方法是有效的。在判别性任务中，其平均绝对标准化得分提高了 28.34%；在生成性任务中，甚至超过了经过微调的上限值。我们的实现代码可在 \url{[此 https 网址](https://github.com/LZY-the-boys/Twin-Merging)} 上获取。},
  url          = {https://openreview.net/forum?id=81YIt63TTn},
}

@InProceedings{Sun2025,
  author       = {Wenju Sun and Qingyong Li and Yangliao Geng and Boyang Li},
  booktitle    = {Forty-second International Conference on Machine Learning},
  title        = {{CAT} Merging: A Training-Free Approach for Resolving Conflicts in Model Merging},
  year         = {2025},
  comment      = {Multi-task model merging offers a promising paradigm for integrating multiple expert models into a unified model without additional training. Existing state-of-the-art techniques, such as Task Arithmetic and its variants, merge models by accumulating task vectors -- the parameter differences between pretrained and finetuned models. However, task vector accumulation is often hindered by knowledge conflicts, leading to performance degradation. To address this challenge, we propose Conflict-Aware Task Merging (CAT Merging), a novel training-free framework that selectively trims conflict-prone components from the task vectors. CAT Merging introduces several parameter-specific strategies, including projection for linear weights and masking for scaling and shifting parameters in normalization layers. Extensive experiments on vision, language, and vision-language tasks demonstrate that CAT Merging effectively suppresses knowledge conflicts, achieving average accuracy improvements of up to 2.5% (ViT-B/32) and 2.0% (ViT-L/14) over state-of-the-art methods.  
{#arxiv-doi-link}

|-----------|------------------------------------------------------------------------------------------|
| Subjects: | Artificial Intelligence (cs.AI); Machine Learning (cs.LG)                                |
| Cite as:  | [arXiv:2505.06977](https://arxiv.org/abs/2505.06977) \[cs.AI\]                           |
|           | (or [arXiv:2505.06977v2](https://arxiv.org/abs/2505.06977v2) \[cs.AI\] for this version) |
|           | <https://doi.org/10.48550/arXiv.2505.06977> Focus to learn more                          |},
  comment-dell = {多任务模型合并提供了一种将多个专家模型整合为一个统一模型的有前景的方法，无需额外训练。现有的前沿技术，如任务算术及其变体，通过累积任务向量（预训练模型和微调模型之间的参数差异）来合并模型。然而，任务向量的累积常常受到知识冲突的阻碍，导致性能下降。为了解决这一挑战，我们提出了冲突感知任务合并（CAT 合并）这一新颖的无需训练的框架，它有选择地从任务向量中修剪出容易产生冲突的组件。CAT 合并引入了几个针对参数的特定策略，包括对线性权重的投影以及对归一化层中的缩放和移动参数的掩码。在视觉、语言和视觉-语言任务上的大量实验表明，CAT 合并有效地抑制了知识冲突，与现有最先进的方法相比，平均准确率提高了 2.5%（ViT-B/32）和 2.0%（ViT-L/14）。
|-----------|------------------------------------------------------------------------------------------|
| 学科： | 人工智能（cs.AI）；机器学习（cs.LG）                                |
| 引用方式：  | [arXiv:2505.06977](https://arxiv.org/abs/2505.06977) \[cs.AI\]                           |
|           | （此版本对应的引用为 [arXiv:2505.06977v2](https://arxiv.org/abs/2505.06977v2) \[cs.AI\]） |
|           | <https://doi.org/10.48550/arXiv.2505.06977>  更多内容请关注学习  |},
  url          = {https://openreview.net/forum?id=zy7Jw91tdh},
}

@InProceedings{Jin2023,
  author       = {Xisen Jin and Xiang Ren and Daniel Preotiuc-Pietro and Pengxiang Cheng},
  booktitle    = {The Eleventh International Conference on Learning Representations},
  title        = {Dataless Knowledge Fusion by Merging Weights of Language Models},
  year         = {2023},
  comment      = {Fine-tuning pre-trained language models has become the prevalent paradigm for building downstream NLP models. Oftentimes fine-tuned models are readily available but their training data is not, due to data privacy or intellectual property concerns. This creates a barrier to fusing knowledge across individual models to yield a better single model. In this paper, we study the problem of merging individual models built on different training data sets to obtain a single model that performs well both across all data set domains and can generalize on out-of-domain data. We propose a dataless knowledge fusion method that merges models in their parameter space, guided by weights that minimize prediction differences between the merged model and the individual models. Over a battery of evaluation settings, we show that the proposed method significantly outperforms baselines such as Fisher-weighted averaging or model ensembling. Further, we find that our method is a promising alternative to multi-task learning that can preserve or sometimes improve over the individual models without access to the training data. Finally, model merging is more efficient than training a multi-task model, thus making it applicable to a wider set of scenarios.},
  comment-dell = {对预训练语言模型进行微调已成为构建下游自然语言处理模型的主流方法。通常情况下，经过微调的模型很容易获取，但其训练数据却难以获得，这是因为存在数据隐私或知识产权方面的顾虑。这给将不同模型的知识融合以生成性能更优的单一模型造成了障碍。在本文中，我们研究了将基于不同训练数据集构建的单个模型合并为一个整体模型的问题，以获得一个在所有数据集领域都能表现良好，并且能够在非领域数据上进行泛化的单一模型。我们提出了一种无需数据的知识融合方法，该方法在参数空间中对模型进行合并，并由最小化合并模型与单个模型之间预测差异的权重来引导。在一系列评估设置中，我们表明所提出的方法显著优于诸如费希尔加权平均或模型组合等基准方法。此外，我们发现我们的方法是多任务学习的一种有前景的替代方案，无需访问训练数据即可保留或有时改进单个模型的表现。最后，模型合并比训练一个多任务模型更为高效，因此它适用于更广泛的场景。},
  url          = {https://openreview.net/forum?id=FCnohuR6AnM},
}

@InProceedings{Li2025,
  author    = {Mei Li and Yuxiang Lu and Qinyan Dai and Suizhi Huang and Yue Ding and Hongtao Lu},
  booktitle = {Forty-second International Conference on Machine Learning},
  title     = {{BECAME}: Bayesian Continual Learning with Adaptive Model Merging},
  year      = {2025},
  comment   = {Continual Learning (CL) strives to learn incrementally across tasks while mitigating catastrophic forgetting. A key challenge in CL is balancing stability (retaining prior knowledge) and plasticity (learning new tasks). While representative gradient projection methods ensure stability, they often limit plasticity. Model merging techniques offer promising solutions, but prior methods typically rely on empirical assumptions and carefully selected hyperparameters. In this paper, we explore the potential of model merging to enhance the stability-plasticity trade-off, providing theoretical insights that underscore its benefits. Specifically, we reformulate the merging mechanism using Bayesian continual learning principles and derive a closed-form solution for the optimal merging coefficient that adapts to the diverse characteristics of tasks. To validate our approach, we introduce a two-stage framework named BECAME, which synergizes the expertise of gradient projection and adaptive merging. Extensive experiments show that our approach outperforms state-of-the-art CL methods and existing merging strategies.},
  url       = {https://openreview.net/forum?id=gU0MwTihsn},
}

@InProceedings{Guodong2024,
  author       = {Guodong DU and Junlin Lee and Jing Li and Runhua Jiang and Yifei Guo and Shuyang Yu and Hanting Liu and Sim Kuan Goh and Ho-Kin Tang and Daojing He and Min Zhang},
  booktitle    = {The Thirty-eighth Annual Conference on Neural Information Processing Systems},
  title        = {Parameter Competition Balancing for Model Merging},
  year         = {2024},
  comment      = {While fine-tuning pretrained models has become common practice, these models often underperform outside their specific domains. Recently developed model merging techniques enable the direct integration of multiple models, each fine-tuned for distinct tasks, into a single model. This strategy promotes multitasking capabilities without requiring retraining on the original datasets. However, existing methods fall short in addressing potential conflicts and complex correlations between tasks, especially in parameter-level adjustments, posing a challenge in effectively balancing parameter competition across various tasks. This paper introduces an innovative technique named PCB-Merging (Parameter Competition Balancing), a lightweight and training-free technique that adjusts the coefficients of each parameter for effective model merging. PCB-Merging employs intra-balancing to gauge parameter significance within individual tasks and inter-balancing to assess parameter similarities across different tasks. Parameters with low importance scores are dropped, and the remaining ones are rescaled to form the final merged model. We assessed our approach in diverse merging scenarios, including cross-task, cross-domain, and cross-training configurations, as well as out-of-domain generalization. The experimental results reveal that our approach achieves substantial performance enhancements across multiple modalities, domains, model sizes, number of tasks, fine-tuning forms, and large language models, outperforming existing model merging methods. The code is publicly available at: \\url{[this https URL](https://github.com/duguodong7/pcb-merging)}.},
  comment-dell = {虽然对预训练模型进行微调已成为常见的做法，但这些模型在超出其特定领域时往往表现不佳。最近发展起来的模型合并技术能够直接将多个针对不同任务进行微调的模型整合到一个单一模型中。这种策略促进了多任务处理能力，且无需基于原始数据集进行重新训练。然而，现有的方法在解决任务之间潜在的冲突和复杂的相关性方面存在不足，尤其是在参数级别的调整方面，这给有效平衡各种任务中的参数竞争带来了挑战。本文介绍了一种创新技术，名为 PCB-Merging（参数竞争平衡），这是一种轻量级且无需训练的技术，用于调整每个参数的系数以实现有效的模型合并。PCB-Merging 采用内平衡来衡量每个任务中的参数重要性，采用外平衡来评估不同任务中的参数相似性。重要性得分低的参数会被删除，剩余的参数会被重新缩放以形成最终的合并模型。我们在各种合并场景中评估了我们的方法，包括跨任务、跨领域和跨训练配置，以及跨域泛化。实验结果表明，我们的方法在多种模态、领域、模型大小、任务数量、微调形式和大型语言模型等方面都实现了显著的性能提升，优于现有的模型合并方法。代码可在以下链接获取：\\url{[此 https URL](https://github.com/duguodong7/pcb-merging)} 。},
  url          = {https://openreview.net/forum?id=l5SbrtvSRS},
}

@InProceedings{Tang2025,
  author    = {Anke Tang and Enneng Yang and Li Shen and Yong Luo and Han Hu and Lefei Zhang and Bo Du and Dacheng Tao},
  booktitle = {The Thirty-ninth Annual Conference on Neural Information Processing Systems},
  title     = {Merging on the Fly Without Retraining: A Sequential Approach to Scalable Continual Model Merging},
  year      = {2025},
  comment   = {Deep model merging represents an emerging research direction that combines multiple fine-tuned models to harness their specialized capabilities across different tasks and domains. Current model merging techniques focus on merging all available models simultaneously, with weight interpolation-based methods being the predominant approach. However, these conventional approaches are not well-suited for scenarios where models become available sequentially, and they often suffer from high memory requirements and potential interference between tasks. In this study, we propose a training-free projection-based continual merging method that processes models sequentially through orthogonal projections of weight matrices and adaptive scaling mechanisms. Our method operates by projecting new parameter updates onto subspaces orthogonal to existing merged parameter updates while using an adaptive scaling mechanism to maintain stable parameter distances, enabling efficient sequential integration of task-specific knowledge. Our approach maintains constant memory complexity to the number of models, minimizes interference between tasks through orthogonal projections, and retains the performance of previously merged models through adaptive task vector scaling. Extensive experiments on CLIP-ViT models demonstrate that our method achieves a 5-8% average accuracy improvement while maintaining robust performance in different task orderings. Code is publicly available at https://github.com/tanganke/opcm .},
  url       = {https://openreview.net/forum?id=rdGMyTPhui},
}

@InProceedings{Chen2025a,
  author       = {Shiqi Chen and Jinghan Zhang and Tongyao Zhu and Wei Liu and Siyang Gao and Miao Xiong and Manling Li and Junxian He},
  booktitle    = {Forty-second International Conference on Machine Learning},
  title        = {Bring Reason to Vision: Understanding Perception and Reasoning through Model Merging},
  year         = {2025},
  comment      = {Vision-Language Models (VLMs) combine visual perception with the general capabilities, such as reasoning, of Large Language Models (LLMs). However, the mechanisms by which these two abilities can be combined and contribute remain poorly understood. In this work, we explore to compose perception and reasoning through model merging that connects parameters of different models. Unlike previous works that often focus on merging models of the same kind, we propose merging models across modalities, enabling the incorporation of the reasoning capabilities of LLMs into VLMs. Through extensive experiments, we demonstrate that model merging offers a successful pathway to transfer reasoning abilities from LLMs to VLMs in a training-free manner. Moreover, we utilize the merged models to understand the internal mechanism of perception and reasoning and how merging affects it. We find that perception capabilities are predominantly encoded in the early layers of the model, whereas reasoning is largely facilitated by the middle-to-late layers. After merging, we observe that all layers begin to contribute to reasoning, whereas the distribution of perception abilities across layers remains largely unchanged. These observations shed light on the potential of model merging as a tool for multimodal integration and interpretation.},
  comment-dell = {视觉语言模型（VLMs）将视觉感知与大型语言模型（LLMs）的通用能力（如推理能力）相结合。然而，这两种能力如何能够相互融合并发挥作用的机制仍不为人所熟知。在这项工作中，我们通过模型合并的方式探索将感知和推理结合起来，并连接不同模型的参数。与以往那些往往专注于相同类型模型合并的研究不同，我们提议跨模态合并模型，从而能够将 LLM 的推理能力融入 VLM 中。通过大量的实验，我们证明了模型合并提供了一种无需训练即可将 LLM 的推理能力转移到 VLM 中的成功途径。此外，我们利用合并后的模型来理解感知和推理的内部机制以及合并对其产生的影响。我们发现，感知能力主要编码在模型的早期层中，而推理则主要由中间到后期的层来促进。在合并之后，我们发现所有层都开始参与到推理过程中，而不同层之间的感知能力分布则基本保持不变。这些观察结果揭示了模型合并作为多模态整合与解释工具的潜在价值。},
  url          = {https://openreview.net/forum?id=ntCAP6tMoX},
}

@InProceedings{Yadav2023a,
  author    = {Yadav, Prateek and Tam, Derek and Choshen, Leshem and Raffel, Colin A and Bansal, Mohit},
  booktitle = {Advances in Neural Information Processing Systems},
  title     = {{TIES}-Merging: Resolving Interference When Merging Models},
  year      = {2023},
  editor    = {A. Oh and T. Naumann and A. Globerson and K. Saenko and M. Hardt and S. Levine},
  pages     = {7093--7115},
  publisher = {Curran Associates, Inc.},
  volume    = {36},
  comment   = {Transfer learning - i.e., further fine-tuning a pre-trained model on a downstream task - can confer significant advantages, including improved downstream performance, faster convergence, and better sample efficiency. These advantages have led to a proliferation of task-specific fine-tuned models, which typically can only perform a single task and do not benefit from one another. Recently, model merging techniques have emerged as a solution to combine multiple task-specific models into a single multitask model without performing additional training. However, existing merging methods often ignore the interference between parameters of different models, resulting in large performance drops when merging multiple models. In this paper, we demonstrate that prior merging techniques inadvertently lose valuable information due to two major sources of interference: (a) interference due to redundant parameter values and (b) disagreement on the sign of a given parameter's values across models. To address this, we propose our method, TRIM, ELECT SIGN \& MERGE (TIES-Merging), which introduces three novel steps when merging models: (1) resetting parameters that only changed a small amount during fine-tuning, (2) resolving sign conflicts, and (3) merging only the parameters that are in alignment with the final agreed-upon sign. We find that TIES-Merging outperforms several existing methods in diverse settings covering a range of modalities, domains, number of tasks, model sizes, architectures, and fine-tuning settings. We further analyze the impact of different types of interference on model parameters, and highlight the importance of resolving sign interference. Our code is available at [this https URL](https://github.com/prateeky2806/ties-merging)},
  
}

@InProceedings{Panariello2025,
  author    = {Aniello Panariello and Daniel Marczak and Simone Magistri and Angelo Porrello and Bart{\l}omiej Twardowski and Andrew D. Bagdanov and Simone Calderara and Joost van de Weijer},
  booktitle = {The Thirty-ninth Annual Conference on Neural Information Processing Systems},
  title     = {Accurate and Efficient Low-Rank Model Merging in Core Space},
  year      = {2025},
  comment   = {In this paper, we address the challenges associated with merging low-rank adaptations of large neural networks. With the rise of parameter-efficient adaptation techniques, such as Low-Rank Adaptation (LoRA), model fine-tuning has become more accessible. While fine-tuning models with LoRA is highly efficient, existing merging methods often sacrifice this efficiency by merging fully-sized weight matrices. We propose the Core Space merging framework, which enables the merging of LoRA-adapted models within a common alignment basis, thereby preserving the efficiency of low-rank adaptation while substantially improving accuracy across tasks. We further provide a formal proof that projection into Core Space ensures no loss of information and provide a complexity analysis showing the efficiency gains. Extensive empirical results demonstrate that Core Space significantly improves existing merging techniques and achieves state-of-the-art results on both vision and language tasks while utilizing a fraction of the computational resources. Codebase is available at [this https URL](https://github.com/apanariello4/core-space-merging).},
  url       = {https://openreview.net/forum?id=y1z7SAS8q8},
}

@InProceedings{Sun2025a,
  author       = {Wenju Sun and Qingyong Li and Wen Wang and Yang Liu and Yangliao Geng and Boyang Li},
  booktitle    = {The Thirty-ninth Annual Conference on Neural Information Processing Systems},
  title        = {Towards Minimizing Feature Drift in Model Merging: Layer-wise Task Vector Fusion for Adaptive Knowledge Integration},
  year         = {2025},
  comment      = {Multi-task model merging aims to consolidate knowledge from multiple fine-tuned task-specific experts into a unified model while minimizing performance degradation. Existing methods primarily approach this by minimizing differences between task-specific experts and the unified model, either from a parameter-level or a task-loss perspective. However, parameter-level methods exhibit a significant performance gap compared to the upper bound, while task-loss approaches entail costly secondary training procedures. In contrast, we observe that performance degradation closely correlates with feature drift, i.e., differences in feature representations of the same sample caused by model merging. Motivated by this observation, we propose Layer-wise Optimal Task Vector Merging (LOT Merging), a technique that explicitly minimizes feature drift between task-specific experts and the unified model in a layer-by-layer manner. LOT Merging can be formulated as a convex quadratic optimization problem, enabling us to analytically derive closed-form solutions for the parameters of linear and normalization layers. Consequently, LOT Merging achieves efficient model consolidation through basic matrix operations. Extensive experiments across vision and vision-language benchmarks demonstrate that LOT Merging significantly outperforms baseline methods, achieving improvements of up to 4.4% (ViT-B/32) over state-of-the-art approaches. The source code is available at [this https URL](https://github.com/SunWenJu123/model-merging).},
  comment-dell = {多任务模型合并旨在将来自多个经过精细调整的特定任务专家的知识整合到一个统一的模型中，同时尽量减少性能下降。现有的方法主要通过从参数层面或任务损失角度来最小化任务特定专家与统一模型之间的差异来实现这一目标。然而，参数层面的方法与上限值相比存在显著的性能差距，而任务损失方法则需要进行昂贵的二次训练过程。相比之下，我们观察到性能下降与特征漂移密切相关，即同一样本的特征表示因模型合并而产生的差异。基于这一观察，我们提出了层级最优任务向量合并（LOT 合并）技术，该技术以逐层的方式明确最小化任务特定专家与统一模型之间的特征漂移。LOT 合并可以表述为一个凸二次优化问题，使我们能够通过解析方法推导出线性和归一化层的参数的闭式解。因此，LOT 合并通过基本的矩阵运算实现了高效的模型整合。在视觉和视觉语言基准测试中的大量实验表明，LOT 合并明显优于基线方法，相较于最先进的方法，其性能提升了高达 4.4%（ViT-B/32）。源代码可在 [此 https URL](https://github.com/SunWenJu123/model-merging) 查看。},
  url          = {https://openreview.net/forum?id=0KOfAUiHua},
}

@InProceedings{Crisostomi2024,
  author    = {Donato Crisostomi and Marco Fumero and Daniele Baieri and Florian Bernard and Emanuele Rodol{\`a}},
  booktitle = {The Thirty-eighth Annual Conference on Neural Information Processing Systems},
  title     = {{C}$^2${M}$^3$: Cycle-Consistent Multi-Model Merging},
  year      = {2024},
  comment   = {In this paper, we present a novel data-free method for merging neural networks in weight space. Differently from most existing works, our method optimizes for the permutations of network neurons globally across all layers. This allows us to enforce cycle consistency of the permutations when merging N \\geq 3 models, allowing circular compositions of permutations to be computed without accumulating error along the path. We qualitatively and quantitatively motivate the need for such a constraint, showing its benefits when merging sets of models in scenarios spanning varying architectures and datasets. We finally show that, when coupled with activation renormalization, our approach yields the best results in the task.},
  url       = {https://openreview.net/forum?id=iD18l6prA7},
}

@InProceedings{Li2024,
  author    = {Lu Li and Tianyu Zhang and Zhiqi Bu and Suyuchen Wang and Huan He and Jie Fu and Yonghui Wu and Jiang Bian and Yong Chen and Yoshua Bengio},
  booktitle = {International Workshop on Federated Foundation Models in Conjunction with NeurIPS 2024},
  title     = {{MAP}: Model Merging with Amortized Pareto Front Using Limited Computation},
  year      = {2024},
  comment   = {Model merging has emerged as an effective approach to combine multiple single-task models into a multitask model. However, existing methods focus on enhancing average task accuracy, often neglecting the trade-offs between different tasks. We introduce Model Merging with Amortized Pareto Front (MAP), a novel low-compute algorithm that efficiently identifies a Pareto set of scaling coefficients for merging multiple models. MAP uses a quadratic approximation surrogate model to estimate task metrics, enabling amortized inference. Our approach is particularly valuable in federated learning scenarios, where it can balance performance across diverse client datasets while respecting privacy constraints and minimizing communication overhead. Experimental results on vision and natural language processing tasks demonstrate MAP's ability to accurately identify the Pareto front, offering practitioners a range of solutions with various trade-offs. This makes MAP a promising approach for optimizing multitask performance in both centralized and distributed learning environments, addressing the challenges of task conflicts and privacy preservation in model merging.},
  url       = {https://openreview.net/forum?id=KfOdVp4pfm},
}

@Article{Yang2026,
  author    = {Yang, Enneng and Shen, Li and Guo, Guibing and Wang, Xingwei and Cao, Xiaochun and Zhang, Jie and Tao, Dacheng},
  journal   = {ACM Computing Surveys},
  title     = {Model Merging in {LLMs}, {MLLMs}, and Beyond: Methods, Theories, Applications, and Opportunities},
  year      = {2026},
  issn      = {0360-0300},
  month     = jan,
  note      = {Just Accepted},
  address   = {New York, NY, USA},
  comment   = {Model merging is an efficient empowerment technique in the machine learning community that does not require the collection of raw training data and does not require expensive computation. As model merging becomes increasingly prevalent across various fields, it is crucial to understand the available model merging techniques comprehensively. However, there is a significant gap in the literature regarding a systematic and thorough review of these techniques. This survey provides a comprehensive overview of model merging methods and theories, their applications in various domains and settings, and future research directions. Specifically, we first propose a new taxonomic approach that exhaustively discusses existing model merging methods. Secondly, we discuss the application of model merging techniques in large language models, multimodal large language models, and more than ten machine learning subfields, including continual learning, multi-task learning, few-shot learning, etc. Finally, we highlight the remaining challenges of model merging and discuss future research directions. A comprehensive list of papers about model merging is available at [this https URL](https://github.com/EnnengYang/Awesome-Model-Merging-Methods-Theories-Applications).},
  doi       = {10.1145/3787849},
  publisher = {Association for Computing Machinery},
  url       = {https://doi.org/10.1145/3787849},
}

@InProceedings{Akizuki2025,
  author    = {Rio Akizuki and Yuya Kudo and Nozomu Yoshinari and Yoichi Hirose and Toshiyuki Nishimoto and Kento Uchida and Shinichi Shirakawa},
  booktitle = {AutoML 2025 Non-Archival Content Track},
  title     = {Surrogate Benchmarks for Model Merging Optimization},
  year      = {2025},
  url       = {https://openreview.net/forum?id=Yv3tRT8olz},
}

@InProceedings{Baba2024,
  author    = {Kaito Baba and Ryota Yagi and Junichiro Takahashi and Risa Kishikawa and Satoshi Kodera},
  booktitle = {Advancements In Medical Foundation Models: Explainability, Robustness, Security, and Beyond},
  title     = {{JR}adi{Evo}: A {Japanese} Radiology Report Generation Model Enhanced by Evolutionary Optimization of Model Merging},
  year      = {2024},
  url       = {https://openreview.net/forum?id=H1osvc7tMP},
}

@Article{Yuan2025,
  author   = {Xiangchi Yuan and Chunhui Zhang and Zheyuan Liu and Dachuan Shi and Soroush Vosoughi and Wenke Lee},
  journal  = {Proceedings of the 2025 Conference on Empirical Methods in Natural Language Processing},
  title    = {Superficial Self-Improved Reasoners Benefit from Model Merging},
  year     = {2025},
  month    = {March},
  pages    = {5912--5932},
  volume   = {abs/2503.02103},
  cdate    = {1740787200000},
  comment  = {As scaled language models (LMs) approach human-level reasoning capabilities, self-improvement emerges as a solution to synthesizing high-quality data corpus. While previous research has identified model collapse as a risk in self-improvement, where model outputs become increasingly deterministic, we discover a more fundamental challenge: the superficial self-improved reasoners phenomenon. In particular, our analysis reveals that even when LMs show improved in-domain (ID) reasoning accuracy, they actually compromise their generalized reasoning capabilities on out-of-domain (OOD) tasks due to memorization rather than genuine. Through a systematic investigation of LM architecture, we discover that during self-improvement, LM weight updates are concentrated in less reasoning-critical layers, leading to superficial learning. To address this, we propose Iterative Model Merging (IMM), a method that strategically combines weights from original and self-improved models to preserve generalization while incorporating genuine reasoning improvements. Our approach effectively mitigates both LM collapse and superficial learning, moving towards more stable self-improving systems.},
  publtype = {informal},
  url      = {https://doi.org/10.48550/arXiv.2503.02103},
}

@Article{Li2025b,
  author   = {Li, Bingdong and Di, Zixiang and Yang, Yanting and Qian, Hong and Yang, Peng and Hao, Hao and Tang, Ke and Zhou, Aimin},
  journal  = {IEEE Transactions on Evolutionary Computation},
  title    = {It’s Morphing Time: Unleashing the Potential of Multiple {LLMs} via Multi-Objective Optimization},
  year     = {2025},
  doi      = {10.1109/TEVC.2025.3613937},
}

@InProceedings{Xu2024,
  author    = {Xu, Zhengqi and Yuan, Ke and Wang, Huiqiong and Wang, Yong and Song, Mingli and Song, Jie},
  booktitle = {Proceedings of the IEEE/CVF Conference on Computer Vision and Pattern Recognition (CVPR)},
  title     = {Training-Free Pretrained Model Merging},
  year      = {2024},
  month     = {June},
  pages     = {5915-5925},
  comment   = {Recently, model merging techniques have surfaced as a solution to combine multiple single-talent models into a single multi-talent model. However, previous endeavors in this field have either necessitated additional training or fine-tuning processes, or require that the models possess the same pre-trained initialization. In this work, we identify a common drawback in prior works w.r.t. the inconsistency of unit similarity in the weight space and the activation space. To address this inconsistency, we propose an innovative model merging framework, coined as merging under dual-space constraints (MuDSC). Specifically, instead of solely maximizing the objective of a single space, we advocate for the exploration of permutation matrices situated in a region with a unified high similarity in the dual space, achieved through the linear combination of activation and weight similarity matrices. In order to enhance usability, we have also incorporated adaptations for group structure, including Multi-Head Attention and Group Normalization. Comprehensive experimental comparisons demonstrate that MuDSC can significantly boost the performance of merged models with various task combinations and architectures. Furthermore, the visualization of the merged model within the multi-task loss landscape reveals that MuDSC enables the merged model to reside in the overlapping segment, featuring a unified lower loss for each task. Our code is publicly available at [this https URL](https://github.com/zju-vipa/training_free_model_merging).},
}

@Article{Lan2025,
  author   = {Xiaochong Lan and Yu Zheng and Shiteng Cao and Yong Li},
  journal  = {arXiv preprint arXiv:2509.22034},
  title    = {The Thinking Spectrum: An Empirical Study of Tunable Reasoning in {LLMs} through Model Merging},
  year     = {2025},
  month    = {September},
  volume   = {abs/2509.22034},
  cdate    = {1756684800000},
  comment  = {The growing demand for large language models (LLMs) with tunable reasoning capabilities in many real-world applications highlights a critical need for methods that can efficiently produce a spectrum of models balancing reasoning depth and computational cost. Model merging has emerged as a promising, training-free technique to address this challenge by arithmetically combining the weights of a general-purpose model with a specialized reasoning model. While various merging techniques exist, their potential to create a spectrum of models with fine-grained control over reasoning abilities remains largely unexplored. This work presents a large-scale empirical study evaluating a range of model merging techniques across multiple reasoning benchmarks. We systematically vary merging strengths to construct accuracy-efficiency curves, providing the first comprehensive view of the tunable performance landscape. Our findings reveal that model merging offers an effective and controllable method for calibrating the trade-off between reasoning accuracy and token efficiency, even when parent models have highly divergent weight spaces. Crucially, we identify instances of Pareto Improvement, where a merged model achieves both higher accuracy and lower token consumption than one of its parents. Our study provides the first comprehensive analysis of this tunable space, offering practical guidelines for creating LLMs with specific reasoning profiles to meet diverse application demands.},
  publtype = {informal},
  url      = {https://doi.org/10.48550/arXiv.2509.22034},
}

@InProceedings{Yang2024a,
  author    = {Enneng Yang and Zhenyi Wang and Li Shen and Shiwei Liu and Guibing Guo and Xingwei Wang and Dacheng Tao},
  booktitle = {The Twelfth International Conference on Learning Representations},
  title     = {{AdaMerging}: Adaptive Model Merging for Multi-Task Learning},
  year      = {2024},
  comment   = {Multi-task learning (MTL) aims to empower a model to tackle multiple tasks simultaneously. A recent development known as task arithmetic has revealed that several models, each fine-tuned for distinct tasks, can be directly merged into a single model to execute MTL without necessitating a retraining process using the initial training data. Nevertheless, this direct addition of models often leads to a significant deterioration in the overall performance of the merged model. This decline occurs due to potential conflicts and intricate correlations among the multiple tasks. Consequently, the challenge emerges of how to merge pre-trained models more effectively without using their original training data. This paper introduces an innovative technique called Adaptive Model Merging (AdaMerging). This approach aims to autonomously learn the coefficients for model merging, either in a task-wise or layer-wise manner, without relying on the original training data. Specifically, our AdaMerging method operates as an automatic, unsupervised task arithmetic scheme. It leverages entropy minimization on unlabeled test samples from the multi-task setup as a surrogate objective function to iteratively refine the merging coefficients of the multiple models. Our experimental findings across eight tasks demonstrate the efficacy of the AdaMerging scheme we put forth. Compared to the current state-of-the-art task arithmetic merging scheme, AdaMerging showcases a remarkable 11\\% improvement in performance. Notably, AdaMerging also exhibits superior generalization capabilities when applied to unseen downstream tasks. Furthermore, it displays a significantly enhanced robustness to data distribution shifts that may occur during the testing phase.},
  url       = {https://openreview.net/forum?id=nZP6NgD3QY},
}

@InProceedings{Ilharco2023,
  author    = {Gabriel Ilharco and Marco Tulio Ribeiro and Mitchell Wortsman and Ludwig Schmidt and Hannaneh Hajishirzi and Ali Farhadi},
  booktitle = {The Eleventh International Conference on Learning Representations},
  title     = {Editing models with task arithmetic},
  year      = {2023},
  comment   = {Changing how pre-trained models behave -- e.g., improving their performance on a downstream task or mitigating biases learned during pre-training -- is a common practice when developing machine learning systems. In this work, we propose a new paradigm for steering the behavior of neural networks, centered around \\textit{task vectors}. A task vector specifies a direction in the weight space of a pre-trained model, such that movement in that direction improves performance on the task. We build task vectors by subtracting the weights of a pre-trained model from the weights of the same model after fine-tuning on a task. We show that these task vectors can be modified and combined together through arithmetic operations such as negation and addition, and the behavior of the resulting model is steered accordingly. Negating a task vector decreases performance on the target task, with little change in model behavior on control tasks. Moreover, adding task vectors together can improve performance on multiple tasks at once. Finally, when tasks are linked by an analogy relationship of the form \`\`A is to B as C is to D", combining task vectors from three of the tasks can improve performance on the fourth, even when no data from the fourth task is used for training. Overall, our experiments with several models, modalities and tasks show that task arithmetic is a simple, efficient and effective way of editing models.},
  url       = {https://openreview.net/forum?id=6t0Kwf8-jrj},
}

@InProceedings{Hu2022,
  author    = {Edward J Hu and yelong shen and Phillip Wallis and Zeyuan Allen-Zhu and Yuanzhi Li and Shean Wang and Lu Wang and Weizhu Chen},
  booktitle = {International Conference on Learning Representations},
  title     = {Lo{RA}: Low-Rank Adaptation of Large Language Models},
  year      = {2022},
  url       = {https://openreview.net/forum?id=nZeVKeeFYf9},
}

@Article{Li2021,
  author  = {Xiang Lisa Li and Percy Liang},
  journal = {Proceedings of the 59th Annual Meeting of the Association for Computational Linguistics and the 11th International Joint Conference on Natural Language Processing (Volume 1: Long Papers)},
  title   = {{Prefix-Tuning}: Optimizing Continuous Prompts for Generation},
  year    = {2021},
  pages   = {4582-4597},
  url     = {https://api.semanticscholar.org/CorpusID:230433941},
}

@Article{Jiang2025,
  author  = {Jiang, Haiyin and Wang, Ruilin and Liang, Weijie and Sun, Qi and Zhang, Xiang and Liu, Yanan},
  journal = {The Journal of Supercomputing},
  title   = {{Slerp-Opt}: merging large language models via adaptive strategies},
  year    = {2025},
  issn    = {1573-0484},
  month   = {08},
  number  = {12},
  pages   = {1223},
  volume  = {81},
  day     = {09},
  doi     = {10.1007/s11227-025-07727-4},
  url     = {https://doi.org/10.1007/s11227-025-07727-4},
}

@Article{Yu2024,
  author        = {Le Yu and Bowen Yu and Haiyang Yu and Fei Huang and Yongbin Li},
  journal       = {arXiv preprint arXiv:2408.03092},
  title         = {Extend Model Merging from Fine-Tuned to Pre-Trained Large Language Models via Weight Disentanglement},
  year          = {2024},
  archiveprefix = {arXiv},
  eprint        = {2408.03092},
  primaryclass  = {cs.CL},
  url           = {https://arxiv.org/abs/2408.03092},
}

@Article{Lu2024a,
  author        = {Jinliang Lu and Ziliang Pang and Min Xiao and Yaochen Zhu and Rui Xia and Jiajun Zhang},
  journal       = {arXiv preprint arXiv:2407.06089},
  title         = {Merge, Ensemble, and Cooperate! A Survey on Collaborative Strategies in the Era of Large Language Models},
  year          = {2024},
  archiveprefix = {arXiv},
  eprint        = {2407.06089},
  primaryclass  = {cs.CL},
  url           = {https://arxiv.org/abs/2407.06089},
}

@Article{Zheng2025,
  author  = {Zheng, Hongling and Shen, Li and Tang, Anke and Luo, Yong and Hu, Han and Du, Bo and Wen, Yonggang and Tao, Dacheng},
  journal = {Nature Machine Intelligence},
  title   = {Learning from models beyond fine-tuning},
  year    = {2025},
  issn    = {2522-5839},
  month   = {01},
  number  = {1},
  pages   = {6--17},
  volume  = {7},
  day     = {01},
  doi     = {10.1038/s42256-024-00961-0},
  url     = {https://doi.org/10.1038/s42256-024-00961-0},
}

@InProceedings{Lv2024,
  author    = {Lv, Kai and Yang, Yuqing and Liu, Tengxiao and Guo, Qipeng and Qiu, Xipeng},
  booktitle = {Proceedings of the 62nd Annual Meeting of the Association for Computational Linguistics (Volume 1: Long Papers)},
  title     = {Full Parameter Fine-tuning for Large Language Models with Limited Resources},
  year      = {2024},
  address   = {Bangkok, Thailand},
  editor    = {Ku, Lun-Wei and Martins, Andre and Srikumar, Vivek},
  month     = aug,
  pages     = {8187--8198},
  publisher = {Association for Computational Linguistics},
  doi       = {10.18653/v1/2024.acl-long.445},
  url       = {https://aclanthology.org/2024.acl-long.445/},
}

@InProceedings{Jin2025,
  author    = {Ruochen Jin and Bojian Hou and Jiancong Xiao and Weijie J Su and Li Shen},
  booktitle = {The Thirteenth International Conference on Learning Representations},
  title     = {Fine-Tuning Attention Modules Only: Enhancing Weight Disentanglement in Task Arithmetic},
  year      = {2025},
  url       = {https://openreview.net/forum?id=dj0TktJcVI},
}

@Article{Tang2023,
  author   = {Anke Tang and Li Shen and Yong Luo and Yibing Zhan and Han Hu and Bo Du and Yixin Chen and Dacheng Tao},
  journal  = {The Twelfth International Conference on Learning Representations},
  title    = {Parameter-Efficient Multi-Task Model Fusion with Partial Linearization},
  year     = {2024},
  volume   = {abs/2310.04742},
  cdate    = {1672531200000},
  publtype = {informal},
  url      = {https://doi.org/10.48550/arXiv.2310.04742},
}

@InProceedings{Sun2025b,
  author    = {Sun, Wenju and Li, Qingyong and Wang, Wen and Geng, Yangliao and Li, Boyang},
  booktitle = {Proceedings of the 33rd ACM International Conference on Multimedia},
  title     = {Task Arithmetic in Trust Region: A Training-Free Model Merging Approach to Navigate Knowledge Conflicts},
  year      = {2025},
  address   = {New York, NY, USA},
  pages     = {5178–5187},
  publisher = {Association for Computing Machinery},
  series    = {MM '25},
  doi       = {10.1145/3746027.3755789},
  isbn      = {9798400720352},
  location  = {Dublin, Ireland},
  numpages  = {10},
  url       = {https://doi.org/10.1145/3746027.3755789},
}

@InProceedings{Wang2020,
  author    = {Wang, Jianyu and Liu, Qinghua and Liang, Hao and Joshi, Gauri and Poor, H. Vincent},
  booktitle = {Proceedings of the 34th International Conference on Neural Information Processing Systems},
  title     = {Tackling the objective inconsistency problem in heterogeneous federated optimization},
  year      = {2020},
  address   = {Red Hook, NY, USA},
  publisher = {Curran Associates Inc.},
  series    = {NIPS '20},
  articleno = {638},
  isbn      = {9781713829546},
  location  = {Vancouver, BC, Canada},
  numpages  = {13},
}

@InProceedings{Liu2025,
  author    = {Deyuan Liu and Zecheng Wang and Bingning Wang and Weipeng Chen and Chunshan Li and Zhiying Tu and Dianhui Chu and Dianbo Sui},
  booktitle = {Forty-second International Conference on Machine Learning},
  title     = {Maximizing Intermediate Checkpoint Value in {LLM} Pretraining with Bayesian Optimization},
  year      = {2025},
  url       = {https://openreview.net/forum?id=UvwWrUV1JV},
}

@article{jiang2026evolutionary,
  title={Evolutionary generative optimization: Towards fully data-driven evolutionary optimization via generative learning},
  author={Jiang, Tao and Sun, Kebin and Liang, Zhenyu and Cheng, Ran and Jin, Yaochu and Tan, Kay Chen},
  journal={IEEE Transactions on Evolutionary Computation},
  year={2026},
  publisher={IEEE}
}

@Article{Gao2024,
  author  = {Gao, Han and Kaltenbach, Sebastian and Koumoutsakos, Petros},
  journal = {Nature Communications},
  title   = {Generative learning for forecasting the dynamics of high-dimensional complex systems},
  year    = {2024},
  number  = {1},
  pages   = {8904},
  volume  = {15},
  doi     = {10.1038/s41467-024-53165-w},
}

@Article{Wang2025,
  author   = {Wang, Feng and Xie, Jinsong and Zhou, Aimin and Tang, Ke},
  journal  = {IEEE Transactions on Evolutionary Computation},
  title    = {A New Prediction Strategy for Dynamic Multiobjective Optimization Using Diffusion Model},
  year     = {2025},
  number   = {5},
  pages    = {1575-1589},
  volume   = {29},
  doi      = {10.1109/TEVC.2025.3551323},
}

@Article{Chung2024,
  author  = {Chung, Hyung Won and Hou, Le and Longpre, Shayne and Zoph, Barret and Tay, Yi and Fedus, William and Li, Yunxuan and Wang, Xuezhi and Dehghani, Mostafa and Brahma, Siddhartha and others},
  journal = {Journal of Machine Learning Research},
  title   = {Scaling instruction-finetuned language models},
  year    = {2024},
  number  = {70},
  pages   = {1--53},
  volume  = {25},
}

@Article{Bai2023,
  author  = {Bai, Jinze and Bai, Shuai and Chu, Yunfei and Cui, Zeyu and Dang, Kai and Deng, Xiaodong and Fan, Yang and Ge, Wenbin and Han, Yu and Huang, Fei and others},
  journal = {arXiv preprint arXiv:2309.16609},
  title   = {Qwen technical report},
  year    = {2023},
}

@Article{Ivison2023,
  author  = {Ivison, Hamish and Wang, Yizhong and Pyatkin, Valentina and Lambert, Nathan and Peters, Matthew and Dasigi, Pradeep and Jang, Joel and Wadden, David and Smith, Noah A and Beltagy, Iz and others},
  journal = {arXiv preprint arXiv:2311.10702},
  title   = {Camels in a changing climate: Enhancing {LM} adaptation with {TULU} 2},
  year    = {2023},
}

@Article{Yang2025,
  author  = {Yang, Zongzhen and Qi, Binhang and Sun, Hailong and Long, Wenrui and Zhao, Ruobing and Gao, Xiang},
  journal = {arXiv preprint arXiv:2503.01874},
  title   = {{CABS}: Conflict-Aware and Balanced Sparsification for Enhancing Model Merging},
  year    = {2025},
}

@InProceedings{Perin2024,
  author    = {Perin, Gabriel and Chen, Xuxi and Liu, Shusen and Kailkhura, Bhavya and Wang, Zhangyang and Gallagher, Brian},
  booktitle = {Findings of the Association for Computational Linguistics: ACL 2024},
  title     = {{R}ank{M}ean: Module-Level Importance Score for Merging Fine-tuned {LLM} Models},
  year      = {2024},
  address   = {Bangkok, Thailand},
  editor    = {Ku, Lun-Wei and Martins, Andre and Srikumar, Vivek},
  month     = aug,
  pages     = {1776--1782},
  publisher = {Association for Computational Linguistics},
  doi       = {10.18653/v1/2024.findings-acl.104},
  url       = {https://aclanthology.org/2024.findings-acl.104/},
}

@Article{Deep2024,
  author   = {Pala Tej Deep and Rishabh Bhardwaj and Soujanya Poria},
  journal  = {arXiv preprint arXiv:2406.11617},
  title    = {{DELLA-Merging}: Reducing Interference in Model Merging through Magnitude-Based Sampling},
  year     = {2024},
  volume   = {abs/2406.11617},
  cdate    = {1704067200000},
  publtype = {informal},
  url      = {https://doi.org/10.48550/arXiv.2406.11617},
}

@InProceedings{Yu2024a,
  author    = {Yu, Le and Yu, Bowen and Yu, Haiyang and Huang, Fei and Li, Yongbin},
  booktitle = {Forty-first International Conference on Machine Learning},
  title     = {Language models are super mario: Absorbing abilities from homologous models as a free lunch},
  year      = {2024},
}

@Article{Matena2022,
  author  = {Matena, Michael S and Raffel, Colin A},
  journal = {Advances in Neural Information Processing Systems},
  title   = {Merging models with {Fisher-weighted} averaging},
  year    = {2022},
  pages   = {17703--17716},
  volume  = {35},
}

@Article{Zhang2025,
  author        = {Kehao Zhang and Shaolei Zhang and Yang Feng},
  journal       = {arXiv preprint arXiv:2508.19839},
  title         = {{PSO-Merging}: Merging Models Based on Particle Swarm Optimization},
  year          = {2025},
  archiveprefix = {arXiv},
  eprint        = {2508.19839},
  primaryclass  = {cs.LG},
  url           = {https://arxiv.org/abs/2508.19839},
}

@InProceedings{Wang2018,
  author    = {Wang, Alex and Singh, Amanpreet and Michael, Julian and Hill, Felix and Levy, Omer and Bowman, Samuel},
  booktitle = {Proceedings of the 2018 EMNLP workshop BlackboxNLP: Analyzing and interpreting neural networks for NLP},
  title     = {{GLUE}: A multi-task benchmark and analysis platform for natural language understanding},
  year      = {2018},
  pages     = {353--355},
}

@InProceedings{Hendrycks2021,
  author    = {Dan Hendrycks and Collin Burns and Steven Basart and Andy Zou and Mantas Mazeika and Dawn Song and Jacob Steinhardt},
  booktitle = {International Conference on Learning Representations},
  title     = {Measuring Massive Multitask Language Understanding},
  year      = {2021},
  url       = {https://openreview.net/forum?id=d7KBjmI3GmQ},
}

@Article{Wang2024,
  author  = {Wang, Yubo and Ma, Xueguang and Zhang, Ge and Ni, Yuansheng and Chandra, Abhranil and Guo, Shiguang and Ren, Weiming and Arulraj, Aaran and He, Xuan and Jiang, Ziyan and others},
  journal = {Advances in Neural Information Processing Systems},
  title   = {{MMLU-Pro}: A more robust and challenging multi-task language understanding benchmark},
  year    = {2024},
  pages   = {95266--95290},
  volume  = {37},
}

@InProceedings{Zellers2019,
  author    = {Rowan Zellers and Ari Holtzman and Yonatan Bisk and Ali Farhadi and Yejin Choi},
  booktitle = {Annual Meeting of the Association for Computational Linguistics},
  title     = {{HellaSwag}: Can a Machine Really Finish Your Sentence?},
  year      = {2019},
  pages     = {4791-4800},
  cdate     = {1546300800000},
  url       = {https://doi.org/10.18653/v1/p19-1472},
}

@Article{Cobbe2021,
  author   = {Karl Cobbe and Vineet Kosaraju and Mohammad Bavarian and Mark Chen and Heewoo Jun and Lukasz Kaiser and Matthias Plappert and Jerry Tworek and Jacob Hilton and Reiichiro Nakano and Christopher Hesse and John Schulman},
  journal  = {arXiv preprint arXiv:2110.14168},
  title    = {Training Verifiers to Solve Math Word Problems},
  year     = {2021},
  volume   = {abs/2110.14168},
  cdate    = {1609459200000},
  publtype = {informal},
  url      = {https://arxiv.org/abs/2110.14168},
}

@InProceedings{Ding2024,
  author    = {Wenxuan Ding and Shangbin Feng and Yuhan Liu and Zhaoxuan Tan and Vidhisha Balachandran and Tianxing He and Yulia Tsvetkov},
  booktitle = {Findings of the Association for Computational Linguistics: ACL 2024},
  title     = {Knowledge Crosswords: Geometric Knowledge Reasoning with Large Language Models},
  year      = {2024},
  pages     = {2609-2636},
  cdate     = {1704067200000},
  url       = {https://doi.org/10.18653/v1/2024.findings-acl.154},
}

@Article{Wang2023,
  author  = {Wang, Heng and Feng, Shangbin and He, Tianxing and Tan, Zhaoxuan and Han, Xiaochuang and Tsvetkov, Yulia},
  journal = {Advances in Neural Information Processing Systems},
  title   = {Can language models solve graph problems in natural language?},
  year    = {2023},
  pages   = {30840--30861},
  volume  = {36},
}

@InProceedings{Zhang2024,
  author    = {Yizhuo Zhang and Heng Wang and Shangbin Feng and Zhaoxuan Tan and Xiaochuang Han and Tianxing He and Yulia Tsvetkov},
  booktitle = {Findings of the Conference on Empirical Methods in Natural Language Processing (EMNLP 2024)},
  title     = {Can {LLM} Graph Reasoning Generalize beyond Pattern Memorization?},
  year      = {2024},
  pages     = {2289-2305},
  cdate     = {1704067200000},
  url       = {https://aclanthology.org/2024.findings-emnlp.127},
}

@Article{Lin2021,
  author  = {Lin, Stephanie and Hilton, Jacob and Evans, Owain},
  journal = {URL https://arxiv. org/abs/2109.07958},
  title   = {Truthfulqa: Measuring how models mimic human falsehoods, 2022},
  year    = {2021},
  volume  = {1},
}

@InProceedings{Gehman2020,
  author    = {Gehman, Samuel and Gururangan, Suchin and Sap, Maarten and Choi, Yejin and Smith, Noah A.},
  booktitle = {Findings of the Association for Computational Linguistics: EMNLP 2020},
  title     = {{R}eal{T}oxicity{P}rompts: Evaluating Neural Toxic Degeneration in Language Models},
  year      = {2020},
  address   = {Online},
  editor    = {Cohn, Trevor and He, Yulan and Liu, Yang},
  month     = nov,
  pages     = {3356--3369},
  publisher = {Association for Computational Linguistics},
  doi       = {10.18653/v1/2020.findings-emnlp.301},
  url       = {https://aclanthology.org/2020.findings-emnlp.301/},
}

@Article{Warstadt2019,
  author    = {Warstadt, Alex and Singh, Amanpreet and Bowman, Samuel R},
  journal   = {Transactions of the Association for Computational Linguistics},
  title     = {Neural network acceptability judgments},
  year      = {2019},
  pages     = {625--641},
  volume    = {7},
  publisher = {MIT Press One Rogers Street, Cambridge, MA 02142-1209, USA journals-info~…},
}

@InProceedings{Socher2013,
  author    = {Socher, Richard and Perelygin, Alex and Wu, Jean and Chuang, Jason and Manning, Christopher D and Ng, Andrew Y and Potts, Christopher},
  booktitle = {Proceedings of the 2013 conference on empirical methods in natural language processing},
  title     = {Recursive deep models for semantic compositionality over a sentiment treebank},
  year      = {2013},
  pages     = {1631--1642},
}

@InProceedings{Cer2017,
  author    = {Cer, Daniel and Diab, Mona and Agirre, Eneko and Lopez-Gazpio, I{\~n}igo and Specia, Lucia},
  booktitle = {Proceedings of the 11th International Workshop on Semantic Evaluation ({S}em{E}val-2017)},
  title     = {{S}em{E}val-2017 Task 1: Semantic Textual Similarity Multilingual and Crosslingual Focused Evaluation},
  year      = {2017},
  address   = {Vancouver, Canada},
  editor    = {Bethard, Steven and Carpuat, Marine and Apidianaki, Marianna and Mohammad, Saif M. and Cer, Daniel and Jurgens, David},
  month     = aug,
  pages     = {1--14},
  publisher = {Association for Computational Linguistics},
  doi       = {10.18653/v1/S17-2001},
  url       = {https://aclanthology.org/S17-2001/},
}

@InProceedings{Dolan2005,
  author    = {Dolan, William B. and Brockett, Chris},
  booktitle = {Proceedings of the Third International Workshop on Paraphrasing ({IWP}2005)},
  title     = {Automatically Constructing a Corpus of Sentential Paraphrases},
  year      = {2005},
  url       = {https://aclanthology.org/I05-5002/},
}

@InProceedings{Williams2018,
  author    = {Williams, Adina and Nangia, Nikita and Bowman, Samuel},
  booktitle = {Proceedings of the 2018 Conference of the North {A}merican Chapter of the Association for Computational Linguistics: Human Language Technologies, Volume 1 (Long Papers)},
  title     = {A Broad-Coverage Challenge Corpus for Sentence Understanding through Inference},
  year      = {2018},
  address   = {New Orleans, Louisiana},
  editor    = {Walker, Marilyn and Ji, Heng and Stent, Amanda},
  month     = jun,
  pages     = {1112--1122},
  publisher = {Association for Computational Linguistics},
  doi       = {10.18653/v1/N18-1101},
  url       = {https://aclanthology.org/N18-1101/},
}

@InProceedings{Giampiccolo2007,
  author    = {Giampiccolo, Danilo and Magnini, Bernardo and Dagan, Ido and Dolan, Bill},
  booktitle = {Proceedings of the {ACL}-{PASCAL} Workshop on Textual Entailment and Paraphrasing},
  title     = {The Third {PASCAL} Recognizing Textual Entailment Challenge},
  year      = {2007},
  address   = {Prague},
  editor    = {Sekine, Satoshi and Inui, Kentaro and Dagan, Ido and Dolan, Bill and Giampiccolo, Danilo and Magnini, Bernardo},
  month     = jun,
  pages     = {1--9},
  publisher = {Association for Computational Linguistics},
  url       = {https://aclanthology.org/W07-1401/},
}

@InProceedings{BarHaim2006,
  author    = {Roy Bar-Haim and Ido Dagan and Bill Dolan and Lisa Ferro and Danilo Giampiccolo and Bernardo Magnini and Idan Szpektor},
  booktitle = {Proceedings of the Second PASCAL Challenges Workshop on Recognising Textual Entailment},
  title     = {The Second {PASCAL} Recognising Textual Entailment Challenge},
  year      = {2006},
  url       = {https://api.semanticscholar.org/CorpusID:13385138},
}

@InProceedings{Dagan2006,
  author    = {Dagan, Ido and Glickman, Oren and Magnini, Bernardo},
  booktitle = {Machine Learning Challenges. Evaluating Predictive Uncertainty, Visual Object Classification, and Recognising Tectual Entailment},
  title     = {The {PASCAL} Recognising Textual Entailment Challenge},
  year      = {2006},
  address   = {Berlin, Heidelberg},
  editor    = {Qui{\~{n}}onero-Candela, Joaquin and Dagan, Ido and Magnini, Bernardo and d'Alch{\'e}-Buc, Florence},
  pages     = {177--190},
  publisher = {Springer Berlin Heidelberg},
  isbn      = {978-3-540-33428-6},
}

@article{bentivogli2009fifth,
  title={The Fifth PASCAL Recognizing Textual Entailment Challenge.},
  author={Bentivogli, Luisa and Clark, Peter and Dagan, Ido and Giampiccolo, Danilo},
  journal={TAC},
  volume={7},
  number={8},
  pages={1},
  year={2009}
}

@InProceedings{Levesque2011,
  author    = {Hector J. Levesque and Ernest Davis and L. Morgenstern},
  booktitle = {AAAI Spring Symposium: Logical Formalizations of Commonsense Reasoning},
  title     = {The Winograd Schema Challenge},
  year      = {2011},
  url       = {https://api.semanticscholar.org/CorpusID:15710851},
}

@Article{Koepf2023,
  author  = {K{\"o}pf, Andreas and Kilcher, Yannic and Von R{\"u}tte, Dimitri and Anagnostidis, Sotiris and Tam, Zhi Rui and Stevens, Keith and Barhoum, Abdullah and Nguyen, Duc and Stanley, Oliver and Nagyfi, Rich{\'a}rd and others},
  journal = {Advances in neural information processing systems},
  title   = {Openassistant conversations-democratizing large language model alignment},
  year    = {2023},
  pages   = {47669--47681},
  volume  = {36},
}

@Misc{Chaudhary2023,
  author       = {Sahil Chaudhary},
  howpublished = {\url{https://github.com/sahil280114/codealpaca}},
  title        = {{Code Alpaca}: An Instruction-following {LLaMA} model for code generation},
  year         = {2023},
  journal      = {GitHub repository},
  publisher    = {GitHub},
}

@misc{OpenOrca,
  title = {OpenOrca: An Open Dataset of GPT Augmented FLAN Reasoning Traces},
  author = {Wing Lian and Bleys Goodson and Eugene Pentland and Austin Cook and Chanvichet Vong and "Teknium"},
  year = {2023},
  publisher = {HuggingFace},
  journal = {HuggingFace repository},
  howpublished = {\url{https://https://huggingface.co/datasets/Open-Orca/OpenOrca}},
}

@article{shao2026pivotmerge,
  title={PivotMerge: Bridging Heterogeneous Multimodal Pre-training via Post-Alignment Model Merging},
  author={Shao, Zibo and Xiong, Baochen and Yang, Xiaoshan and Song, Yaguang and Zhang, Qimeng and Chen, Haifeng and Xu, Changsheng},
  journal={arXiv preprint arXiv:2604.22823},
  year={2026}
}

@InProceedings{Zhou2023,
  author    = {Zhou, Chunting and Liu, Pengfei and Xu, Puxin and Iyer, Srini and Sun, Jiao and Mao, Yuning and Ma, Xuezhe and Efrat, Avia and Yu, Ping and Yu, Lili and Zhang, Susan and Ghosh, Gargi and Lewis, Mike and Zettlemoyer, Luke and Levy, Omer},
  booktitle = {Proceedings of the 37th International Conference on Neural Information Processing Systems},
  title     = {{LIMA}: less is more for alignment},
  year      = {2023},
  address   = {Red Hook, NY, USA},
  publisher = {Curran Associates Inc.},
  series    = {NIPS '23},
  articleno = {2400},
  location  = {New Orleans, LA, USA},
  numpages  = {16},
}

@InProceedings{Xu2024a,
  author    = {Can Xu and Qingfeng Sun and Kai Zheng and Xiubo Geng and Pu Zhao and Jiazhan Feng and Chongyang Tao and Qingwei Lin and Daxin Jiang},
  booktitle = {The Twelfth International Conference on Learning Representations},
  title     = {Wizard{LM}: Empowering Large Pre-Trained Language Models to Follow Complex Instructions},
  year      = {2024},
  url       = {https://openreview.net/forum?id=CfXh93NDgH},
}

@InProceedings{Wei2022,
  author    = {Wei, Jason and Wang, Xuezhi and Schuurmans, Dale and Bosma, Maarten and Ichter, Brian and Xia, Fei and Chi, Ed H. and Le, Quoc V. and Zhou, Denny},
  booktitle = {Proceedings of the 36th International Conference on Neural Information Processing Systems},
  title     = {Chain-of-thought prompting elicits reasoning in large language models},
  year      = {2022},
  address   = {Red Hook, NY, USA},
  publisher = {Curran Associates Inc.},
  series    = {NIPS '22},
  articleno = {1800},
  isbn      = {9781713871088},
  location  = {New Orleans, LA, USA},
  numpages  = {14},
}

@Article{Inoshita2026,
  author  = {Keito Inoshita and Xiaokang Zhou and Akira Kawai},
  journal = {2024 IEEE International Conference on Data Science and Systems (DSS)},
  title   = {Multi-Stage Evolutionary Model Merging with Meta Data Driven Curriculum Learning for Sentiment-Specialized Large Language Modeling},
  year    = {2024},
  pages   = {58-65},
  url     = {https://api.semanticscholar.org/CorpusID:284648786},
}
\bibliographystyle{icml2026}

\newpage
\appendix
\onecolumn

\section{Complexity Analysis}
To evaluate the efficiency of EvoGM, we analyze its computational overhead in terms of both time and space complexity. Let $N$ denote the number of expert models, $D$ the number of parameters per model (or LoRA rank), $R$ the number of outer rounds, $T$ the inner evolutionary iterations, $P$ the population size, and $E$ the cost of a single evaluation on the validation set. 

\paragraph{Time Complexity.} The time complexity of EvoGM is primarily dominated by the iterative evaluation and expert refinement phases. Initially, computing task vectors requires $O(ND)$ operations in the parameter space. During the main loop, the generative optimization occurs in the low-dimensional coefficient space with a negligible cost of $O(RTP \cdot \text{poly}(N))$, as $N \ll D$. The most significant bottleneck is the fitness evaluation of candidate models, which sequentially takes $O(RTPE)$. However, since evaluations of candidate coefficients are embarrassingly parallel, the total wall-clock time can be significantly reduced by employing $M$ parallel workers. In such a parallelized setting, the time complexity is optimized to $O(ND + RT \lceil P/M \rceil E + RKND)$, where the final term accounts for the $K$ model merging operations per round during expert refinement. This linear scalability with respect to $M$ allows EvoGM to handle large populations and intensive search rounds efficiently.

\paragraph{Space Complexity.} In terms of space complexity, EvoGM is designed to be highly memory-efficient, especially within parameter-efficient fine-tuning (PEFT) frameworks. The primary memory footprint consists of the frozen backbone model and $N$ task vectors, resulting in a total requirement of $O(D_{base} + ND_{task})$. Throughout the evolutionary process, the algorithm only needs to store a history of population coefficients and their corresponding scores, which occupies $O(RTPN)$ space. Given that $N$ is typically small, this storage cost is marginal compared to the model parameters. During the evaluation phase, EvoGM avoids the need for multiple full-model replicas by dynamically merging task vectors in-place, maintaining a consistent space complexity of $O(D + ND)$. This efficiency ensures that our approach remains feasible even for large-scale language models with limited hardware resources.


\section{Additional Robustness and Scaling Results}
\label{app:rebuttal_results}

For the experiments reported in this section, we used Ascend 910 NPUs together with HiSilicon AArch64 processors. The other experiments were conducted on NVIDIA A100-SXM4-40GB GPUs with AMD EPYC 7742 64-Core processors. Since these platforms differ in their underlying hardware architectures, minor discrepancies in absolute performance may arise. However, within each experimental setting, our method and all baselines were executed on identical hardware, thereby ensuring the fairness of the comparative evaluation.

\subsection{Statistical Robustness}
\label{app:statistical_robustness}
To assess whether small average gains are driven by random variation, we reran EvoGM, PSO-Merging, and Model Swarm with matched random seeds on the Qwen2.5-1.5B unseen-task benchmark. \cref{tab:statistical_robustness} reports mean and standard deviation over four runs and independent two-sample Welch's t-tests. While not every task-level difference is significant, EvoGM obtains a statistically significant improvement in the average score over PSO-Merging ($p=0.0100$) and Model Swarm ($p=0.0006$).

\begin{table}[htb]
\centering
\scriptsize
\caption{Multi-seed statistical robustness on Qwen2.5-1.5B unseen tasks.}
\label{tab:statistical_robustness}
\resizebox{\textwidth}{!}{%
\begin{tabular}{lccccc}
\toprule
\textbf{Task} & \textbf{EvoGM} & \textbf{PSO-Merging} & \textbf{Model Swarm} & \textbf{EvoGM vs. PSO} & \textbf{EvoGM vs. Swarm} \\
 & mean $\pm$ std & mean $\pm$ std & mean $\pm$ std & $p$-value & $p$-value \\
\midrule
MMLU & 0.5942 $\pm$ 0.0184 & 0.5990 $\pm$ 0.0072 & 0.6015 $\pm$ 0.0163 & 0.6614 & 0.5770 \\
MMLU-Pro & 0.2412 $\pm$ 0.0084 & 0.2298 $\pm$ 0.0072 & 0.2400 $\pm$ 0.0139 & 0.0840 & 0.8839 \\
HellaSwag & 0.5930 $\pm$ 0.0074 & 0.5925 $\pm$ 0.0115 & 0.5900 $\pm$ 0.0262 & 0.9446 & 0.8377 \\
Knowledge Crosswords & 0.4080 $\pm$ 0.0041 & 0.3853 $\pm$ 0.0418 & 0.3948 $\pm$ 0.0068 & 0.3563 & 0.0211 \\
GSM8K & 0.4408 $\pm$ 0.0256 & 0.3675 $\pm$ 0.0454 & 0.3295 $\pm$ 0.0070 & 0.0400 & 0.0021 \\
NLGraph & 0.5262 $\pm$ 0.0184 & 0.5443 $\pm$ 0.0506 & 0.3748 $\pm$ 0.0005 & 0.5420 & 0.0005 \\
TruthfulQA & 0.4344 $\pm$ 0.0159 & 0.4315 $\pm$ 0.0184 & 0.4303 $\pm$ 0.0148 & 0.8234 & 0.7220 \\
MMLU-Abstain & 0.1333 $\pm$ 0.0223 & 0.1137 $\pm$ 0.0172 & 0.1330 $\pm$ 0.0265 & 0.2190 & 0.9890 \\
\textbf{Average} & 0.4214 $\pm$ 0.0073 & 0.4018 $\pm$ 0.0019 & 0.3867 $\pm$ 0.0076 & \textbf{0.0100} & \textbf{0.0006} \\
\bottomrule
\end{tabular}}
\end{table}

\subsection{Fairness of the Basis Shift}
\label{app:basis_shift_fairness}
To test whether EvoGM benefits merely from being granted a multi-round basis shift, we also applied the same two-round, three-iteration-per-round schedule to PSO-Merging. As shown in \cref{tab:multiround_pso}, simply adding the same basis-reset mechanism to PSO-Merging reduces the average score from 0.4018 to 0.3425. This indicates that the basis shift is not a generic advantage for any search method; in EvoGM, it works together with learned winner-loser mappings and cycle consistency to preserve useful search structure across rounds.

\begin{table}[htb]
\centering
\small
\caption{Single-round PSO, multi-round PSO with the same basis-reset schedule, and multi-round EvoGM on Qwen2.5-1.5B.}
\label{tab:multiround_pso}
\begin{tabular}{lccc}
\toprule
\textbf{Task} & \textbf{PSO Single-Round} & \textbf{PSO Multi-Round} & \textbf{EvoGM Multi-Round} \\
\midrule
MMLU & 0.5990 & 0.5660 & 0.5942 \\
MMLU-Pro & 0.2298 & 0.2070 & 0.2412 \\
HellaSwag & 0.5925 & 0.5840 & 0.5930 \\
Knowledge Crosswords & 0.3853 & 0.3790 & 0.4080 \\
GSM8K & 0.3675 & 0.2060 & 0.4408 \\
NLGraph & 0.5443 & 0.2820 & 0.5262 \\
TruthfulQA & 0.4315 & 0.4279 & 0.4344 \\
MMLU-Abstain & 0.1137 & 0.0880 & 0.1333 \\
\textbf{Average} & \textbf{0.4018} & \textbf{0.3425} & \textbf{0.4214} \\
\bottomrule
\end{tabular}
\end{table}

\subsection{Scaling and Search Cost on Qwen3-8B}
\label{app:qwen3_8b_scaling}
We further evaluate EvoGM on Qwen3-8B with ten Tulu-finetuned experts. \cref{tab:qwen3_8b_results} shows that EvoGM achieves the best average score among the tested methods. We also report wall-clock time and forward-pass counts in \cref{tab:qwen3_8b_cost}. Although EvoGM uses more candidate evaluations than Model Swarm in this setup, the total wall-clock time remains competitive because candidate evaluation is parallelizable and the small MLP generators add negligible overhead relative to model inference.

\begin{table}[htb]
\centering
\scriptsize
\caption{Qwen3-8B test performance with ten Tulu-finetuned experts.}
\label{tab:qwen3_8b_results}
\resizebox{\textwidth}{!}{%
\begin{tabular}{lccccccccc}
\toprule
\textbf{Method} & \textbf{MMLU} & \textbf{MMLU-Pro} & \textbf{HellaSwag} & \textbf{K-Cross} & \textbf{GSM8K} & \textbf{NLGraph} & \textbf{TruthQA} & \textbf{AbstainQA} & \textbf{AVG} \\
\midrule
Base & 0.655 & 0.349 & 0.792 & 0.589 & 0.125 & 0.534 & 0.561 & 0.308 & 0.489 \\
MTL & 0.702 & 0.399 & 0.712 & 0.667 & 0.428 & 0.363 & 0.558 & 0.296 & 0.516 \\
Single Best & \textbf{0.732} & \textbf{0.432} & 0.802 & 0.644 & 0.614 & 0.548 & 0.630 & \textbf{0.400} & 0.600 \\
Task Arithmetic & 0.721 & \textbf{0.432} & 0.809 & 0.580 & 0.355 & 0.413 & 0.566 & 0.232 & 0.513 \\
TIES Merging & 0.673 & 0.369 & 0.805 & 0.558 & 0.214 & 0.457 & 0.571 & 0.349 & 0.500 \\
Model Swarm & 0.712 & 0.420 & 0.789 & \textbf{0.669} & 0.588 & 0.551 & \textbf{0.637} & 0.381 & 0.593 \\
\textbf{EvoGM} & 0.718 & 0.429 & \textbf{0.824} & 0.586 & \textbf{0.655} & \textbf{0.607} & 0.614 & 0.394 & \textbf{0.603} \\
\bottomrule
\end{tabular}}
\end{table}

\begin{table}[htb]
\centering
\small
\caption{Search cost on Qwen3-8B using the Ascend 910 NPUs together with HiSilicon AArch64 processors. FPs denotes candidate forward-pass evaluations on validation data.}
\label{tab:qwen3_8b_cost}
\begin{tabular}{lcccc}
\toprule
\textbf{Task} & \textbf{EvoGM Time} & \textbf{EvoGM FPs} & \textbf{Model Swarm Time} & \textbf{Model Swarm FPs} \\
\midrule
MMLU & 43.1 min & 160 & 32.5 min & 100 \\
MMLU-Pro & 40.4 min & 160 & 26.2 min & 100 \\
HellaSwag & 38.4 min & 160 & 31.0 min & 120 \\
K-Cross & 37.7 min & 160 & 40.0 min & 120 \\
GSM8K & 86.5 min & 160 & 98.2 min & 100 \\
NLGraph & 58.5 min & 160 & 68.3 min & 100 \\
TruthQA & 38.7 min & 160 & 23.0 min & 100 \\
AbstainQA & 42.2 min & 160 & 109.7 min & 160 \\
\textbf{Total} & \textbf{385.5 min} & \textbf{1280} & \textbf{428.9 min} & \textbf{900} \\
\bottomrule
\end{tabular}
\end{table}

\subsection{Higher-Dimensional Expert Space}
\label{app:vit20}
To probe a larger coefficient space, we compare EvoGM with CMA-ES on a ViT-B-16 model merging benchmark with 20 experts. Both methods use the same validation/test subsets and the same six-iteration search budget. \cref{tab:vit20_results} shows that EvoGM obtains a higher average test score than CMA-ES (0.6363 vs. 0.6268). It suggests that the generative search mechanism remains useful beyond the 8-10 expert settings used in the main experiments.

\begin{table}[htb]
\centering
\small
\caption{ViT-B-16 merging with 20 experts under an identical six-iteration budget.}
\label{tab:vit20_results}
\begin{tabular}{lcc}
\toprule
\textbf{Task} & \textbf{CMA-ES} & \textbf{EvoGM} \\
\midrule
Cars & 0.5028 & \textbf{0.7006} \\
DTD & \textbf{0.6528} & 0.5111 \\
EuroSAT & 0.4828 & \textbf{0.8106} \\
GTSRB & \textbf{0.7794} & 0.5061 \\
MNIST & \textbf{0.7050} & 0.5894 \\
RESISC45 & 0.6767 & \textbf{0.7367} \\
SVHN & \textbf{0.8956} & 0.4556 \\
SUN397 & 0.6206 & \textbf{0.6783} \\
STL10 & 0.9622 & \textbf{0.9783} \\
OxfordIIITPet & \textbf{0.9189} & 0.8439 \\
Flowers102 & 0.6172 & \textbf{0.6217} \\
CIFAR100 & 0.7300 & \textbf{0.7611} \\
PCAM & \textbf{0.5744} & 0.5706 \\
FER2013 & \textbf{0.4922} & 0.4394 \\
CIFAR10 & 0.8511 & \textbf{0.9433} \\
Food101 & 0.6406 & \textbf{0.8867} \\
FashionMNIST & 0.5128 & \textbf{0.8044} \\
RenderedSST2 & 0.4872 & \textbf{0.6433} \\
EMNIST & \textbf{0.2939} & 0.1294 \\
KMNIST & \textbf{0.1406} & 0.1161 \\
\textbf{Average} & 0.6268 & \textbf{0.6363} \\
\bottomrule
\end{tabular}
\end{table}

\section{Experimental Details}
\subsection{Baselines}
We select a total of 13 representative model merging methods as baselines. The details of these methods are summarized below.

\begin{itemize}

\item \textbf{Model Soup:}
This approach merges multiple fine-tuned models by uniformly averaging
their parameters~\cite{Wortsman2022}. It assumes that independently
trained models lie in a shared low-loss basin, enabling effective
aggregation without additional tuning.

\item \textbf{TA (Task Arithmetic):}
Task Arithmetic~\cite{Ilharco2023} represents task-specific adaptations
as parameter difference vectors relative to a base model and combines
them through linear addition. This formulation enables compositional
transfer across tasks but may suffer from interference under naive
aggregation.

\item \textbf{DARE:}
DARE mitigates parameter conflicts by sparsifying task vectors before
aggregation~\cite{Yu2024a}. By retaining only high-magnitude parameter
updates, it reduces destructive interference during merging.

\item \textbf{TIES:}
Rather than averaging all parameters, this method resolves sign
conflicts across task vectors and selectively trims inconsistent
dimensions~\cite{Yadav2023a}. The resulting merge emphasizes parameters
with consistent directional contributions.

\item \textbf{DARE-TIES:}
Combining sparsification and sign-based conflict resolution, this
variant first applies DARE-style pruning and then performs TIES-based
trimming~\cite{Yadav2023a,Yu2024a}. The hybrid design aims to further
suppress interference while preserving task-relevant updates.

\item \textbf{DELLA:}
This method identifies task-relevant subspaces through low-rank
decomposition and performs merging within these aligned
representations~\cite{Deep2024}. By operating in a shared latent space,
it improves compatibility across merged models.

\item \textbf{RankMean:}
Instead of averaging raw parameter values, RankMean aggregates
parameters based on their relative rankings across
models~\cite{Perin2024}. This rank-based formulation increases
robustness to scale mismatches and outlier updates.

\item \textbf{CMA:}
CMA formulates model merging as a black-box optimization problem and
applies covariance matrix adaptation to search for optimal merging
coefficients~\cite{Akiba2025}. The method iteratively updates a
population of candidates based on validation performance.

\item \textbf{AdaMerging:}
This approach learns adaptive, layer-wise merging coefficients guided
by validation feedback~\cite{Yang2024a}. The coefficients are optimized
to balance task contributions across different network depths.

\item \textbf{Fisher:}
Fisher-weighted merging leverages Fisher information to estimate
parameter importance for each model~\cite{Matena2022}. Parameters
deemed more critical receive higher weights during aggregation,
reducing harmful interference.

\item \textbf{RegMean:}
By incorporating regularization terms during aggregation, this method
penalizes deviations from important parameters of individual
models~\cite{Jin2023}. The regularized formulation stabilizes merging
under heterogeneous tasks.

\item \textbf{PSO-Merging:}
PSO-Merging formulates model merging as a particle swarm optimization
problem in parameter space~\cite{Zhang2025}. It iteratively updates
candidate merged models based on personal and global best solutions
guided by task performance, enabling data-driven yet gradient-free
merging.

\item \textbf{Model Swarm:}
Model Swarm treats each expert model as a particle and performs
collaborative search in the weight space under a task-specific utility
function~\cite{feng2025model}. Instead of static aggregation, it adaptively
explores model combinations and returns the best-performing expert
discovered during optimization.

\end{itemize}

\begin{table}[htb]
  \centering
  \small
  \caption{Details of task datasets used in our experiments.}
  \label{tab:dataset_stats}
  \begin{tabular}{l p{5.5cm} cc c}
  \toprule
  \multirow{2}{*}{\textbf{Dataset}} & \multirow{2}{*}{\textbf{Source}} & \multicolumn{2}{c}{\textbf{Size}} & \multirow{2}{*}{\textbf{Domain}} \\
  \cmidrule(lr){3-4}
   &  & Val & Test &  \\
  \midrule
  CoLA        & ~\cite{Warstadt2019} & 50  & 1043  & Misc. \\
      MNLI        &   ~\cite{Williams2018}      & 50  & 9815  & Misc. \\
  MRPC        &   ~\cite{Dolan2005}      & 50  & 408   & News \\
  QNLI        &    ~\cite{Levesque2011}     & 50  & 5463  & Wikipedia \\
  QQP\footnote{\url{https://www.quora.com/profile/Ricky-Riche-2/First-Quora-Dataset-Release-Question-Pairs}}          &    /    & 50  & 40430 & Social QA questions \\
  RTE         &    ~\cite{Dagan2006,BarHaim2006,Giampiccolo2007,bentivogli2009fifth}     & 50  & 277   & News, Wikipedia \\
  SST-2       &    ~\cite{Socher2013}     & 50  & 872   & Movie reviews \\
  STS-B       &   ~\cite{Cer2017}      & 50  & 1500  & Misc. \\
  \midrule
  MMLU        & ~\cite{Hendrycks2021} & 200 & 1000  & Knowledge \\
  MMLU-Pro    & ~\cite{Wang2024}      & 70  & 1000  & Knowledge \\
  Hellaswag   & ~\cite{Zellers2019}   & 200 & 1000  & Knowledge \\
  K-Cross     & ~\cite{Ding2024}      & 200 & 1000  & Cross-domain reasoning \\
  GSM8k       & ~\cite{Cobbe2021}     & 200 & 1000  &  Reasoning tasks\\
  NLGraph     & ~\cite{Wang2023,Zhang2024} & 200 & 1000  & Reasoning tasks \\
  TruthfulQA  & ~\cite{Lin2021}       & 200 & 617   & Safety-related evaluation \\
  AbstainQA   & ~\cite{Gehman2020}    & 200 & 1000  & Safety-related evaluation \\
  \bottomrule
  \end{tabular}
  \end{table}

\subsection{Datasets}
We use sixteen datasets from established benchmarks,
covering natural language understanding, reasoning, and safety
evaluation. Eight datasets are drawn from the GLUE benchmark~\cite{Wang2018},
including CoLA, MNLI, MRPC, QNLI, QQP, RTE, SST-2, and STS-B, which together
represent a broad spectrum of natural language understanding tasks. In
addition, we include eight widely used datasets for evaluating knowledge,
reasoning, and safety evaluation, which are MMLU~\cite{Hendrycks2021}, MMLU-Pro~\cite{Wang2024},
HellaSwag~\cite{Zellers2019}, GSM8K~\cite{Cobbe2021}, K-Cross
(K-Cross)~\cite{Ding2024}, NLGraph~\cite{Wang2023,Zhang2024},
TruthfulQA~\cite{Lin2021}, and AbstainQA~\cite{Gehman2020}. Detailed
dataset statistics are reported in Table~\ref{tab:dataset_stats}.

\subsection{Expert Models}
Our experiments involve a total of eighteen expert models to be merged.
All experts share the same architecture within each setting and are
independently adapted from a common base model using different training
datasets. An overview of the expert models and their corresponding
training data is provided in Table~\ref{tab:models}.
Eight experts are obtained by fine-tuning FLAN-T5-base\footnote{\url{https://huggingface.co/google/flan-t5-base}} on
individual GLUE datasets~\cite{Wang2018}, while the remaining ten experts are adapted
from Qwen2.5-1.5B\footnote{\url{https://huggingface.co/Qwen/Qwen2.5-1.5B}} using LoRA-based parameter-efficient
fine-tuning on supervised fine-tuning datasets from the
Tulu-v2\footnote{\url{https://huggingface.co/collections/allenai/tulu-v2-suite}}~\cite{Ivison2023} collection. All experts are trained for five epochs with
an initial learning rate of $2\times10^{-4}$ and an effective batch size
of 32.

\begin{table}[htb]
\centering
\small
\caption{Expert models used in our experiments.}
\label{tab:models}
\begin{tabular}{l l p{6cm} c}
\toprule
Base model & Source & Train Dataset & Models \\
\midrule
\multirow{8}{*}{{FLAN-T5\footnote{\url{https://huggingface.co/google/flan-t5-base}}}} 
& \multirow{8}{*}{Open models} 
& CoLA ~\cite{Warstadt2019}& FLAN-T5-cola\footnote{\url{https://huggingface.co/tanganke/flan-t5-base_glue-cola}} \\
& & MNLI  ~\cite{Williams2018} & FLAN-T5-mnli\footnote{\url{https://huggingface.co/tanganke/flan-t5-base_glue-mnli}} \\
& & MRPC  ~\cite{Dolan2005}  & FLAN-T5-mrpc\footnote{\url{https://huggingface.co/tanganke/flan-t5-base_glue-mrpc}} \\
& & QNLI   ~\cite{Levesque2011} &  FLAN-T5-qnli\footnote{\url{https://huggingface.co/tanganke/flan-t5-base_glue-qnli}}\\
& & QQP\footnote{\url{https://huggingface.co/datasets/AlekseyKorshuk/quora-question-pairs}}  &  FLAN-T5-qqp\footnote{\url{https://huggingface.co/tanganke/flan-t5-base_glue-qqp}}   \\
& & RTE  ~\cite{Dagan2006,BarHaim2006,Giampiccolo2007,bentivogli2009fifth}  &  FLAN-T5-rte\footnote{\url{https://huggingface.co/tanganke/flan-t5-base_glue-rte}}      \\
& & SST-2  ~\cite{Socher2013} & FLAN-T5-sst2\footnote{\url{https://huggingface.co/tanganke/flan-t5-base_glue-sst2}}\\
& & STS-B  ~\cite{Cer2017}  &FLAN-T5-stsb\footnote{\url{https://huggingface.co/tanganke/flan-t5-base_glue-stsb}}\\
\midrule
\multirow{10}{*}{{\makecell{QWEN2.5\\-1.5B\footnote{\url{https://huggingface.co/Qwen/Qwen2.5-1.5B}}} }} 
& \multirow{10}{*}{\makecell{LoRA-based \\fine-tuning}} 
& flan  ~\cite{Chung2024}          &        -       \\
& & CoT  ~\cite{Wei2022}          &  -    \\
& & OpenAssistant-1 ~\cite{Koepf2023}       &   -   \\
& & ShareGPT  \footnote{\url{https://sharegpt.com/}}                 & -\\
& & Code Alpaca   ~\cite{Chaudhary2023}         &  -    \\
& & LIMA ~\cite{Zhou2023}                      & - \\
& & WizardLM Evol-Instruct v2 ~\cite{Xu2024a}  & -\\
& & Open-Orca    ~\cite{OpenOrca}               & -\\
& & Science Literature  ~\cite{Ivison2023}       &  -\\
& & Gemini Alpaca (distilled)   &-\\
\bottomrule
\end{tabular}
\end{table}

\subsection{More Experimental Results}

\FigurePerformanceAll

\FigureAblationAll

\TableAblationStudy

\TableParameterTuning

\TableNumberModel

\FigureNumberRadarAll

\end{document}